\documentclass[runningheads]{llncs}

\usepackage{eccv}

\usepackage{eccvabbrv}

\usepackage{bm}
\usepackage{arydshln}
\usepackage{graphicx}
\usepackage{amssymb}
\usepackage{booktabs}
\usepackage{algorithm}
\usepackage{algorithmic}
\usepackage{tabularx}
\usepackage{multirow}
\usepackage{tikz}
\usepackage{pgfplots}
\usetikzlibrary{spy,calc}
\usepackage{arydshln}
\usepackage{anyfontsize}

\newcommand{\fastshap}{{\sc f}ast{\sc shap}\xspace}
\newcommand{\fsod}{\textsc{fsod}\xspace}

\usepackage{tikz}
\usetikzlibrary{calc, fit, positioning, shadows}
\usepackage{dsfont}
\usepackage{pgf}
\usepackage{pgfplots}
\usepackage{xspace}
\pgfplotsset{width=5cm,compat=1.9}
\pgfplotsset{every tick label/.append style={font=\tiny}}

\usepackage{pifont}

\newcommand{\rex}{\textsc{r}\textnormal{e}\textsc{x}\xspace}
\newcommand{\drise}{{\textsc{drise}}\xspace}

\newcommand{\blackcatt}{BlackCAtt\xspace}
\newcommand{\blackcattMOG}{$\text{BlackCAtt}_{\text{MoG}}$\xspace}
\newcommand{\blackcattGreedy}{$\text{BlackCAtt}_{\text{Bl}}$\xspace}
\newcommand{\yolo}{\textsc{yolo}\xspace}

\newcommand{\gradcam}{{\sc g}\textnormal{rad}-{\sc cam}\xspace}

\newcommand{\rise}{{\sc rise}\xspace}
\newcommand{\xai}{XAI\xspace}

\newcommand{\od}{OD\xspace}
\newcommand{\msps}{MSPS\xspace}
\newcommand{\mspss}{MSPSs\xspace}

\newcommand{\commentout}[1]{}

\usepackage[pagebackref,breaklinks,colorlinks,citecolor=eccvblue]{hyperref}

\usepackage{orcidlink}

\begin{document}

\title{Out-of-the-box: Black-box Causal Attacks on Object Detectors}

\titlerunning{Out-of-the-box}

\author{Melane Navaratnarajah\inst{1}\orcidlink{0009-0001-8987-6134} \and
David A. Kelly\inst{1}\orcidlink{0000-0002-5368-6769} \and
Hana Chockler\inst{1}\orcidlink{0000-0003-1219-0713}}

\authorrunning{Navaratnarajah et al.}

\institute{
  King's College, London, UK\\
\email{\{melane.navaratnarajah,david.a.kelly,hana.chockler\}@kcl.ac.uk}
}

\maketitle

\begin{abstract}
Adversarial perturbations are a useful way to expose vulnerabilities in object detectors. Existing perturbation methods are frequently white-box, architecture specific and use a loss function. More importantly, while they are often successful, it is rarely clear \emph{why} they work. Insights into the mechanism of this success would allow developers to understand and analyze these attacks, as well as fine-tune the model to prevent them.
This paper presents \blackcatt{}, a black-box algorithm and tool, which uses minimal, causally sufficient pixel sets to construct explainable, imperceptible, reproducible, architecture-agnostic attacks on object detectors. We evaluate \blackcatt{}
on standard benchmarks and compare it to other black-box adversarial attacks methods.
When \blackcatt{} has access only to the position and label of a bounding box, it produces attacks that are comparable
or better to those produced by other black-box methods. 
When \blackcatt{} has access to the model confidence as well, it can work as a meta-algorithm, improving the ability of 
standard black-box techniques to construct smaller, less perceptible attacks. 
As \blackcatt attacks manipulate causes only, the attacks become fully explainable. 
We compare the performance of \blackcatt{} with other black-box attack methods and show that targeting causal pixels leads to smaller and less perceptible attacks. For example, when using \blackcatt{} with SquareAttack, it reduces the average 
distance ($L_0$ norm) of the attack from the original input from $0.987$ to $0.072$, while maintaining a similar success rate. 
We perform ablation studies on 
the \blackcatt algorithm and analyze the effect of different components on its performance.






\end{abstract}

\section{Introduction}\label{sec:intro}
\begin{figure*}[t]
    \centering
    \begin{subfigure}[t]{0.21\textwidth}
        \centering
        \includegraphics[scale=0.2]{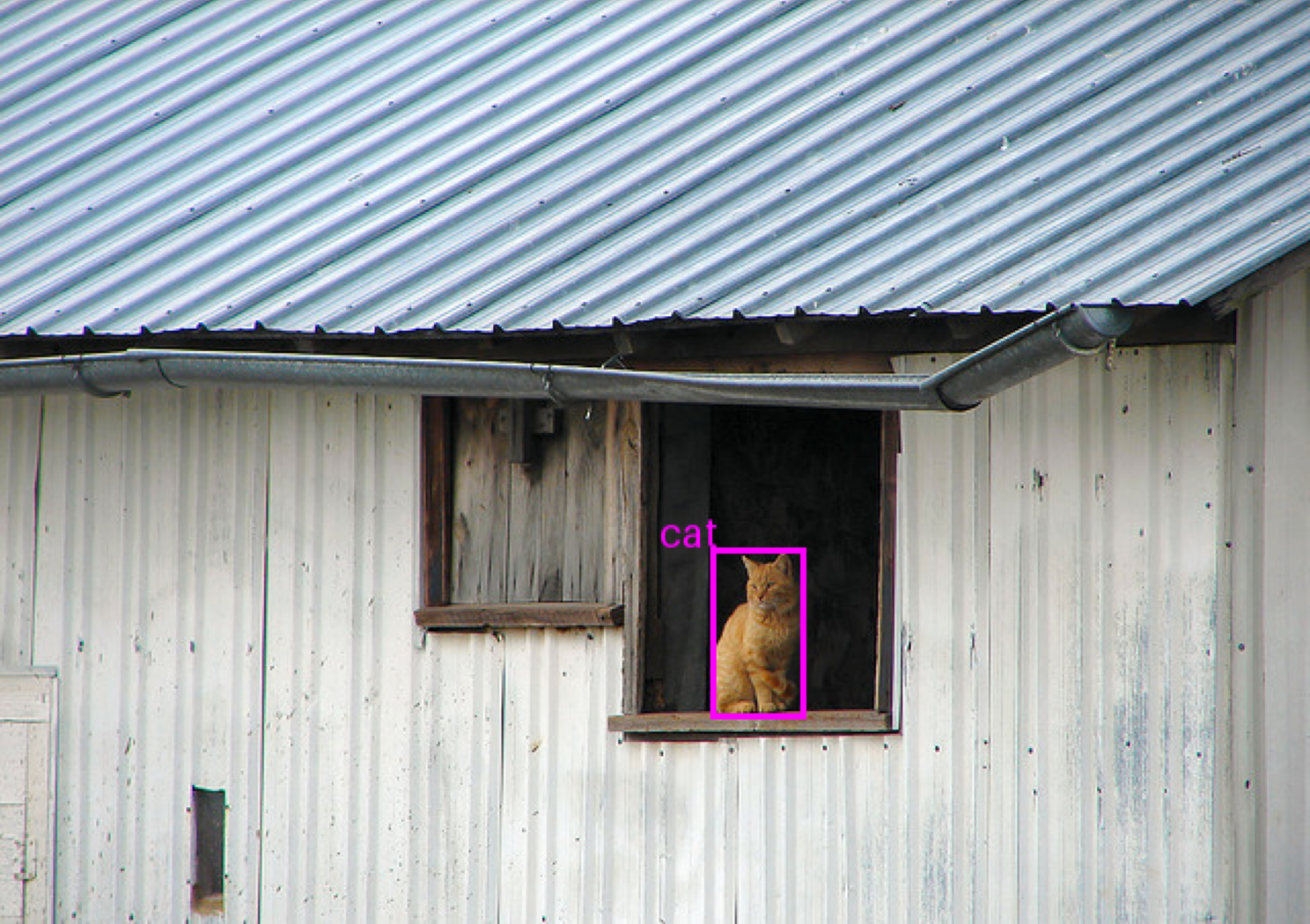}
        \caption{Cat with bounding box}
        \label{fig:cat:main}
    \end{subfigure}
    \hfill
    \centering 
    \begin{subfigure}[t]{0.21\textwidth}
        \centering
        \includegraphics[scale=0.2]{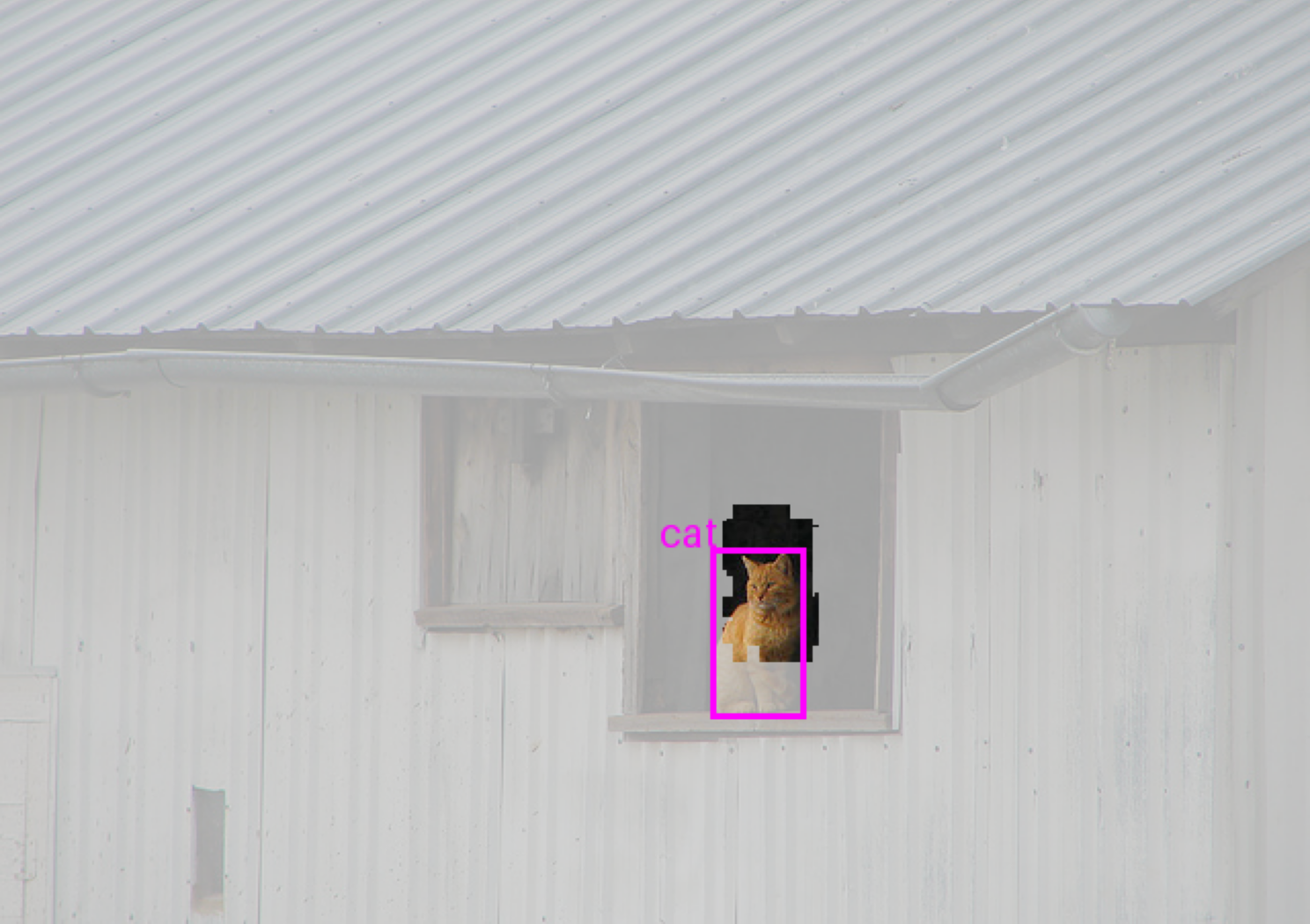}
        \caption{Bbox with \msps}
        \label{fig:cat:msps}
    \end{subfigure}
    \hfill
    \begin{subfigure}[t]{0.21\textwidth}
        \centering
        \includegraphics[scale=0.11]{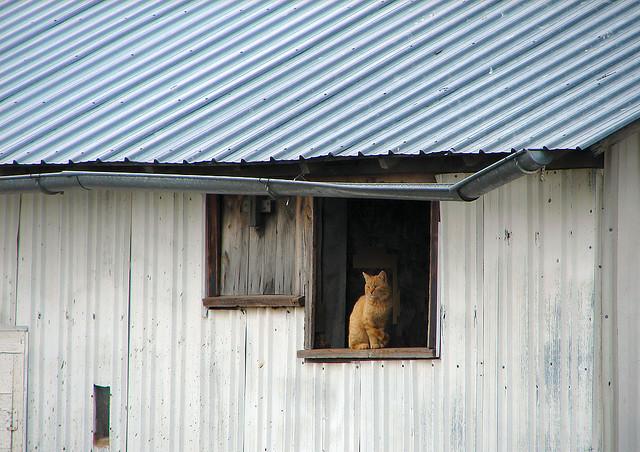}
        \caption{No cat 1: blur}
        \label{fig:cat:blur}
    \end{subfigure}
    \hfill
    \begin{subfigure}[t]{0.21\textwidth}
        \centering
        \includegraphics[scale=0.11]{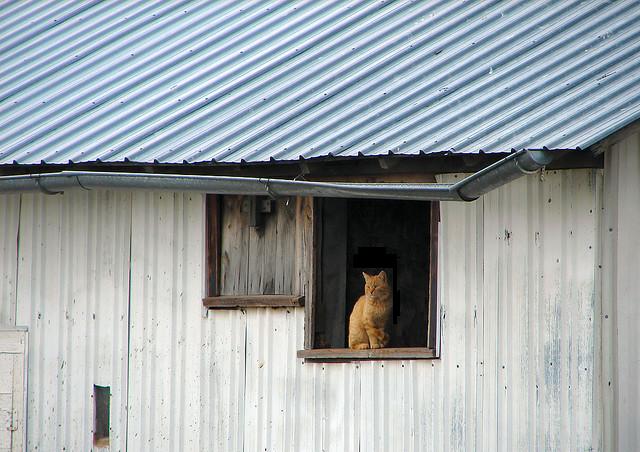}
        \caption{No cat 2: black}
        \label{fig:cat:black}
    \end{subfigure}
     \caption{The \msps for cat (\Cref{fig:cat:msps}) reveals a dependency on the surrounding context. \blackcatt starts with causal pixels \emph{outside} of the bounding box and works inwards in order to maximize imperceptibility. In both~\Cref{fig:cat:blur,fig:cat:black} the cat is still clearly present and complete, but \yolo no longer detects the cat. The attack works because \blackcatt changes part of the \emph{cause} of the detection.}%
    \label{fig:cat}
\end{figure*}

Picture yourself in a self-driving car when you suddenly see a dog in the road in front of you. The car has detected it as well, via its object detector. Then, for no apparent reason, the dog is no longer detected: its bounding box has vanished. The dog is still there, but to the car it has become invisible. You have to intervene and apply the brakes to avoid an accident.
Why did the dog vanish?

Object detectors (\od) are known to be vulnerable to both accidental and adversarial perturbations~\cite{odattacks2025}. In fact, image classification models in general are quite easy to attack~\cite{simba,onepixel2019}. What is harder to understand is \emph{why} the attacks work. Generic attacks, such as global gaussian noise, are reproducible and demonstrate model vulnerability, but do not reveal the causal relationship between pixel-level perturbations and failure modes.

There is growing interest in exploring adversarial attacks on object detectors using e\underline{X}plainable \underline{AI} (\xai) techniques~\cite{Yahn2025AdversarialAP,wang2022adversarial}. These approaches are mostly \emph{white-box} methods: they need access to \od's hidden layers, which is a generous, and unnatural, attack model. Moreover, saliency maps produced from the hidden layers are well known to be noisy, sensitive to input perturbations and not naturally interpretable~\cite{REOW2025,zhang25a}.

In this paper, we present \blackcatt (\underline{Black}-box \underline{C}ausal \underline{Att}acks), a black-box, causal approach to generating adversarial attacks on object detectors. \blackcatt targets \emph{minimal, sufficient pixel sets (\mspss)} for a detected object. These pixels, by themselves, are enough to cause the required detection~\cite{chockler2024causal}. \blackcatt uses these pixels to generate low-distortion attacks that remove, alter, or introduce detections. 

One might expect that all \mspss would be contained within the bounding box. One of the most surprising results in this paper is that \mspss are, in fact, frequently either \emph{fully outside}, or \emph{not fully contained} within the \od bounding box. 
We exploit this phenomenon, showing that perturbing causal pixels outside the box often makes the box disappear (\Cref{sec:results}). 
This happens across different detector architectures (single-stage, two-stage, and transformer-based). We also compare the accuracy and precision of \blackcatt's native \mspss, made using \rex~\cite{chockler2024causal}, by applying our extraction and perturbation techniques to another popular black box saliency tool, \drise, and quantify attack success with a number of measures, including perceptual distortion~\cite{zhang2018perceptual}. 

\blackcatt is model agnostic, architecture agnostic, and produces attacks which are smaller and less perceptible than other black-box methods. \blackcatt works in different information environments. We show that, when given only the bounding box and label, \blackcatt outperforms other similarly restricted attack methods. When \blackcatt also has access to a the model score and loss function, it can work as a meta-algorithm, making other standard attack methods both less perceptible and more explainable.

\Cref{fig:cat} illustrates our approach, showing an image detected by \yolo as a cat, and an \msps (\Cref{fig:cat:msps}) computed by \blackcatt that has causal pixels partially outside of the bounding box. \Cref{fig:cat:blur,fig:cat:black} show two successful adversarial attacks on this image
by \blackcatt which are purely decision-based. 

We note that an \msps does not need to satisfy human intuition: models are not people. A good \xai method should reveal what a model uses, not what a human expects to see~\cite{bhusalface}.
The explainability of our attacks means that we detect the most important causes for the classification. By finding the cause, and attacking only the cause, it is easy to see \emph{why} the attack works. If the cause changes, then it naturally follows that the effect may also change. \blackcatt allows the user to see exactly what information the \od was using and what changes are permissible to that information.

To summarize, the contributions of this paper are as follows.
We introduce \blackcatt, a causally grounded, black-box explainability attack for \od.
We characterize the spatial relationship between \mspss and \od bounding boxes, showing that causal pixels frequently lie \emph{outside} the bounding box.
We propose and evaluate two algorithms, \blackcattGreedy  and \blackcattMOG, that use both pixel ranking and \mspss to synthesize low-distortion perturbations for three different attack goals: removing the detection, changing the classification, and adding a new, spurious detection. These two methods require only the bounding box and label. We show the superiority of both our algorithms compared to two standard techniques. We also use two different black-box tools to generate initial saliency maps, showing the general applicability of our approach.

Moreover, we show that \blackcatt can be used as a \emph{meta-algorithm} to benefit other attack methods 
when more information is available. By allowing \blackcatt access to a loss function, we improve the imperceptibility and explainability of state-of-the-art attack methods.

We provide a reproducible evaluation protocol and a set of quantitative measures---success rate, $L_0$, $L_1$, and $L_2$ norms, 
Learned Perceptual Image Patch Similarity (LPIPS), and others---to evaluate our methods and assess the trade-off between attack effectiveness and perceptual distortion. Due to the lack of space, the full experimental results, additional illustrations and examples, our code, and results are included in the supplementary material.

\section{Background}\label{sec:background}
We provide a brief description of the two \xai tools that we use to generate the initial saliency: \rex and \drise~\cite{drise}. We then discuss the measures
we use to assess the quality of the adversarial attacks and then definitions relevant to the paper.

\textbf{Overview of \rex}\label{sec:rex}:
\rex~\cite{chockler2024causal} is built on the theory of actual causality~\cite{Hal19}. While the full theory of actual causality is complex, an \emph{explanation} for an image reduces to a minimal set of pixels sufficient to have the same top-$1$ classification as the original (\Cref{subsec:defns}). We refer to these as minimal sufficient pixels sets, or \mspss, for the rest of the paper. An \msps may be very small ~\cite{kelly2025big} and an image may have multiple \mspss~\cite{CKK25}, 
\rex is an occlusion-based black-box method and ranks pixels by their approximate \emph{causal responsibility}~\cite{CH04} to produce a \emph{responsibility map}. It uses this ranking to discover one, or more, \mspss. 
\blackcatt uses a bounding-box-aware version of \rex to discover its \mspss. 

\textbf{Overview of \drise}:
\drise~\cite{drise} is a variant of \rise~\cite{petsiuk2018rise}, modified for use with \od models. \drise randomly occludes parts of
the image in order to detect the influence of different pixels on the outcome, and constructs a saliency map reflecting the relative importance of
pixels. \blackcatt uses the saliency map of \drise in the same way it uses the responsibility map of \rex, even though the saliency map of
\drise is not, strictly speaking, causal.

Both tools produce different types of map, indicating the relative importance of pixels. In what follows, we refer to these simply as \emph{maps}.

\textbf{Measures}:
We use a range of different measures to illustrate the differences between the adversarial example and original image. These image similarity measures include pixel-wise distances such as $L_1$ (absolute differences), $L_2$ (squared Euclidean) and $L_0$ (the count of differing elements), while LPIPS~\cite{zhang2018perceptual} measures the perceptual similarity using deep feature representations from a pretrained neural network. We normalize by image height, width and channel for ease of comparison.

\section{Definitions}\label{subsec:defns}
Both \rex and \drise rely on masking parts of the image with a \emph{masking value} that is considered neutral \emph{wrt} the detection.
We take the definition of an \msps from~\cite{chockler2024causal}, where it is called an \emph{explanation} (we use the term \msps here for clarity, as
an \msps does not need to convey any human-compatible intuition).

An \msps is a minimal subset of pixels of a
given input image that is sufficient for the model $\mathcal{N}$ to classify the image, where ``sufficient'' is defined as containing only this subset of pixels from the original image, with all other pixels set to some masking value(s).

For each detection, we use the \msps extraction algorithm from \rex to obtain an \msps.
We then compute the \emph{fraction of the \msps inside the bounding box (bbox)}, denoted FIN, as follows. 
\begin{equation}\label{eq:frac}
\text{FIN} = \frac{|\text{\msps} \cap \text{bbox}|}{|\text{\msps}|}.
\end{equation}
Higher values of FIN indicate that the \msps aligns closely with the detector's localized region. 
The \emph{dice coefficient (DC)}, a standard measure of overlap between sets, is defined in~\cite{Dic45}. 
The DC between the bbox and \msps is computed as follows. 
\begin{equation}\label{eq:dc}
\text{DC}(\text{bbox},\text{\msps}) = \frac{2 * |\text{\msps} \cap \text{bbox}|}{|\text{\msps} + \text{bbox}|}.\\
\end{equation}
In this paper we are only using DC for the overlap between bbox and \msps, hence we omit the parameters of DC.

\section{Preliminary Analysis: Baseline Attacks on \mspss}\label{sec:spatialanalysis}\label{sec:single}
To build intuition and to serve as an ablation study of our techniques, we examine how the \mspss produced by \rex interact with the \od's bounding boxes on the COCO dataset~\cite{lin2014microsoft}. 
For FIN, there is a monotone upwards trend, when computing the spearman rank correlation $\rho$, $ \rho \approx0.3$ across both YOLO~\cite{yolo} and RT-DETR~\cite{lv2023detrs}, showing that higher confidence detections are accompanied by \mspss with larger fractions inside the bounding box. However, while one might expect the \mspss, especially for high confidence detections, 
to fall entirely within the bbox, $62.45\%$ of them ($5437$ images) are not fully contained, indicating that parts of the cause of the detection frequently extend beyond the object boundary annotated by the detector.

\begin{figure}[t]
    \centering
     \resizebox{0.5\linewidth}{!}{\input{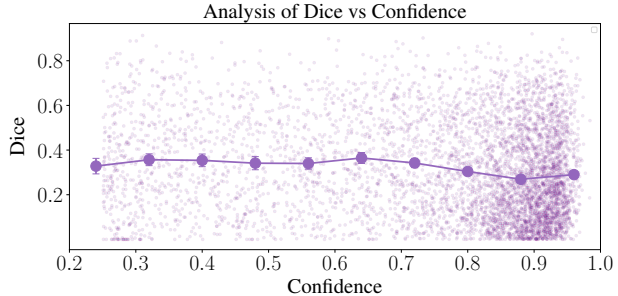}} 
    \caption{The DC between bounding box and \msps stays almost constant on the COCO dataset, regardless of \yolo confidence.} 
    \label{fig:DCvsConf}
\end{figure}

In contrast, the \text{DC} shows a weak negative association, $\rho \approx -0.12$, with detection confidence in FasterRCNN and $\rho \approx -0.18$ for both \yolo (\Cref{fig:DCvsConf}) and RT-DETR, showing that overlap does not improve with confidence and 
that even for confident detections, some causal regions remain outside of the bbox.

We start by evaluating a suite of black-box perturbations applied to pixels of the \msps, split by whether those pixels fall inside or outside the detector's bbox for each model. This is to check that targeting \msps pixels either outside or inside the bbox indeed produces a measurable effect on \od output.

We quantify how detector's predictions change when perturbations are applied to the \msps, thereby testing whether the identified regions are vulnerable. These perturbations are \emph{single-step}, that is, involving one call to the model and no refinement: Gaussian blur (\textbf{blur}), additive Gaussian noise (\textbf{noise}), small spatial shifts (\textbf{shift}) and global pixel offsets (\textbf{pixel value}). For each attack, we report three outcome types: (i) the detection is removed (``no-prediction''), (ii) the predicted class changes, or (iii) a new/spurious detection appears. 

Our experiments show that perturbing the causal pixels, even the pixels that are outside the bounding box, can effectively disrupt 
detections (see supplementary material for the full results). 
\Cref{fig:cake} shows a cake, in which the bounding box (in blue) covers most of the image. 
However, it suffices to perturb the pixels \emph{outside the bounding box} (in red in~\Cref{fig:cake:main}) 
to completely remove the detection (\Cref{fig:cake:blur,fig:cake:noise,fig:cake:pixel,fig:cake:shift}, zoomed in).

There are consistent differences between perturbations inside and outside the detector's bounding box. For YOLO, single-step perturbations of causal pixels inside the box produce a higher no-prediction rate than of those outside the box ($7.8\%$ vs $3.3\%$). However, successful outside perturbations require substantially smaller average area changes (mean area $\approx 948$px outside vs $\approx 3001$px inside). 
For RT-DETR, success rates are much lower (no-prediction: $0.10\%$ inside, $0.07\%$ outside), but the same pattern holds: successful outside perturbations uses a mean area 
$\approx 38$px outside vs $\approx 163$px inside. 

In short, attacking the part of the \msps inside the bounding box is more reliable, while attacking the pixels of the \msps outside 
the bbox works less frequently, but has lower LPIPS and $L_2$. These trends hold across the different attack types and persist when controlling for detection confidence. Different attack goals---no-prediction, changing a prediction, and adding new predictions---work better at different confidence ranges (see the supplementary material). 

These results are for a na\"{i}ve application of attacks to the causal pixels in the \msps. In the next section, \cref{sec:method}, we introduce two novel algorithms, which greatly improve on these initial findings.


\begin{figure*}[t]
    \centering
    \begin{subfigure}[t]{0.28\textwidth}
    \begin{tikzpicture}
        \node[anchor=south west,inner sep=0] at (0,0) 
        {\includegraphics[scale=0.14]{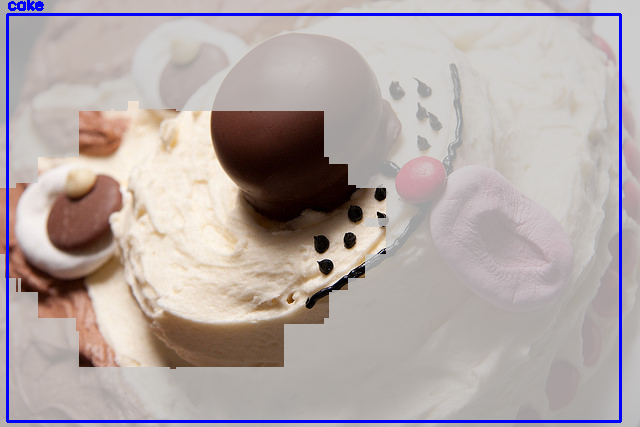}};
        \draw[red,rounded corners] (-0.1, 0.6) rectangle (0.12, 1.4);
        \end{tikzpicture}
        \caption{`Cake' with $0.273$ confidence. \msps outside the bounding box highlighted in red.}
        \label{fig:cake:main}
    \end{subfigure}
    \hfill 
    \begin{subfigure}[t]{0.1\textwidth}
        \includegraphics[width=\linewidth,viewport=0 100 100 270,clip]{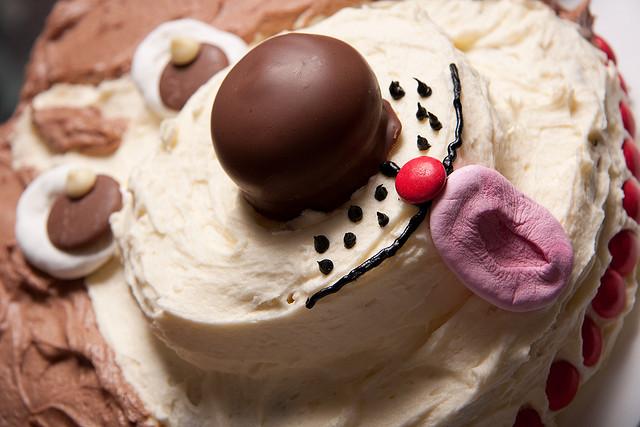}
        \caption{Blur}
        \label{fig:cake:blur}
    \end{subfigure}
    \hfill 
    \begin{subfigure}[t]{0.1\textwidth}
        \includegraphics[width=\linewidth,viewport=0 100 100 270,clip]{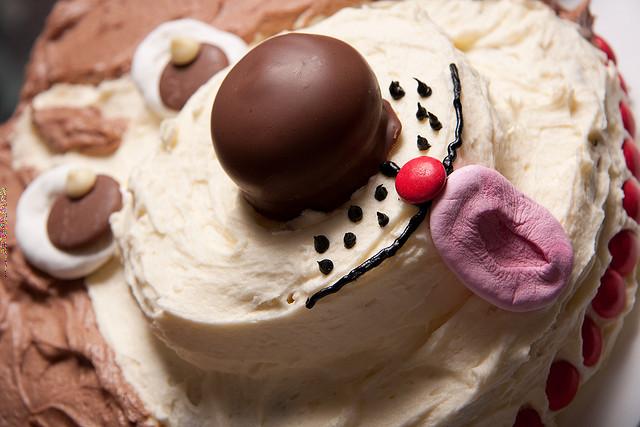}
        \caption{Noise}
        \label{fig:cake:noise}
    \end{subfigure}
    \hfill 
    \begin{subfigure}[t]{0.1\textwidth}
        \includegraphics[width=\linewidth,viewport=0 100 100 270,clip]{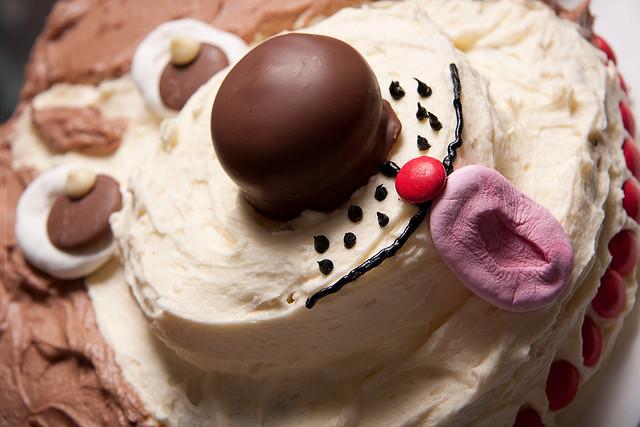}
        \caption{Pixel Value}
        \label{fig:cake:pixel}
    \end{subfigure}
    \hfill 
    \begin{subfigure}[t]{0.1\textwidth}
        \includegraphics[width=\linewidth,viewport=0 100 100 270,clip]{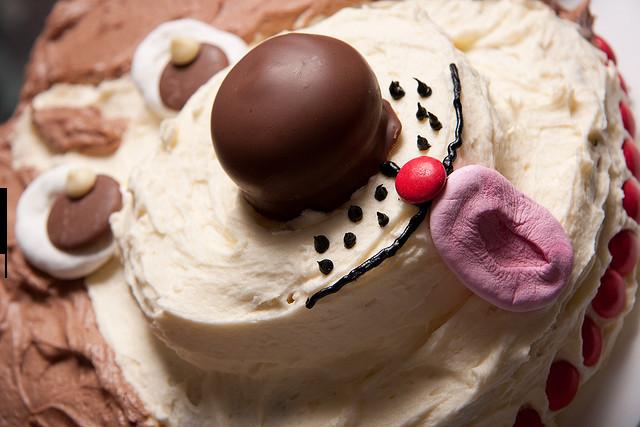}
        \caption{Shift}
        \label{fig:cake:shift}
    \end{subfigure}
     \caption{Causally explainable adversarial attacks on \emph{cake}. Even though the bounding box takes up the majority of the image (\Cref{fig:cake:main}), it is enough to perturb a small number of pixels \emph{outside} the box in order to remove the detection. These pixels are part of the \msps.
    In particular, the gaussian blur attack in \Cref{fig:cake:blur} is imperceptible.}%
    \label{fig:cake}
\end{figure*}

\section{\blackcatt}\label{sec:method}
Based on the initial experiments with the single-step attacks in~\cref{sec:single}, we introduce two complementary black-box attacks that leverage \mspss as spatial priors: \blackcattGreedy and \blackcattMOG. While both work well, their performance differs slightly depending on attack type and model (see~\Cref{sec:results} and supplementary material).
Moreover, while \blackcattGreedy requires two different maps, \blackcattMOG only requires the map of the actual classification, making it a more general approach. Both algorithms rely only on label and bounding box location. We explore the effect of also including the confidence score in~\Cref{sec:results}.

\paragraph*{\textbf{\blackcatt}$_{\bm{Bl}}$} works by blending two different maps. Given an input image $\mathcal{X}$, we obtain its map, $R(\mathcal{X})$, and extract the corresponding \msps. We then calculate a second, complementary, map $R(\bar{\mathcal{X}})$, 
which indicates regions \emph{least associated} with a classification. While both \drise and \rex provide maps for the actual detection, \rex also allows a user to set a ``target'' classification which is different from the actual detection. We use this to compute $R(\bar{\mathcal{X}})$, obtained by setting the target to \textsc{NONE}. Because of this flexibility, we evaluate \blackcattGreedy with \rex maps only (\Cref{sec:results}).

\blackcattGreedy uses the combined map $\mathcal{C}$, computed as
$R(\mathcal{X}) + R(\bar{\mathcal{X}}) = \mathcal{C}$, 
to find a minimally perturbed image, $\mathcal{X}_p$, that causes the label's bounding box, $b$, to be either removed, its label altered, or make a new detection appear, while keeping the distortion below a given parameter $\epsilon$, that is, ensuring that
$\|\mathcal{X} - \mathcal{X}_p\| < \epsilon$.

We restrict our attack search to \emph{modes} within the \msps. These modes are (i) pixels outside the bbox; 
(ii) pixels inside the bbox; and finally (iii) the whole \msps. 
\blackcatt does not consider anything outside the discovered \msps.
\Cref{algo:blended} starts by testing the modes, applying an attack and progressively adding more pixels from the \msps. The order of addition is based on their ranking in the original map.
The $\textit{apply\_refinement}$ function checks if a mode yields a success (\eg bbox removal) at a lower distortion. 
The function \emph{apply\_refinement} tries to reduce the size of the attack by
reducing the perturbation strength and searching over a smaller pixel budget using the ranking obtained from $\mathcal{C}$ (see supplementary material).

\begin{algorithm}[t]
    \caption{\blackcattGreedy \\ $(\mathcal{N}, \mathcal{X}, \text{GAMMAS}, R(\mathcal{X}), R(\bar{\mathcal{X}}), M$) }
    \label{algo:blended}
    \begin{flushleft}
        \textbf{INPUT:}\,\, 
        an object detector $\mathcal{N}$,  an image $\mathcal{X}$,
         a list of perturbation strengths $\gamma \in$ GAMMAS,
        a map, $R(\mathcal{X})$, a no prediction map, $R(\bar{\mathcal{X}})$, 
        the \msps, $M$, extracted from $R(\mathcal{X})$ 
         \\
        \textbf{OUTPUT:}\,\, metrics and masks of the best attacks for each goal(no\_pred, change\_pred and added\_new\_pred)
    \end{flushleft}     
        \begin{algorithmic}[1]
            \STATE $\mathcal{C} \leftarrow R(\mathcal{X}) + R(\bar{\mathcal{X}})$ (Combine Maps)
            \FORALL{$\gamma \in \text{GAMMAS}$}
                \FORALL{$m \in [\text{outside-\msps}, \text{inside-\msps}, M]$}
                    \STATE mask $\leftarrow  m \, \& \, \mathcal{C} $
                    \STATE $\mathcal{X}_p \leftarrow$ apply\_attack(mask, $\mathcal{X})$
                    \STATE preds $\leftarrow \mathcal{N}(\mathcal{X}_p)$
                    \IF{metrics\_improve(preds)}
                        \STATE metrics $\leftarrow$ update\_metrics(preds)
                    \ENDIF
                \ENDFOR
            \ENDFOR
            \RETURN apply\_refinement($\mathcal{X}, \mathcal{N}$, mask, metrics)
        \end{algorithmic}
\end{algorithm}



\begin{figure}[t]
    \centering
    \includegraphics[width=0.48\linewidth]{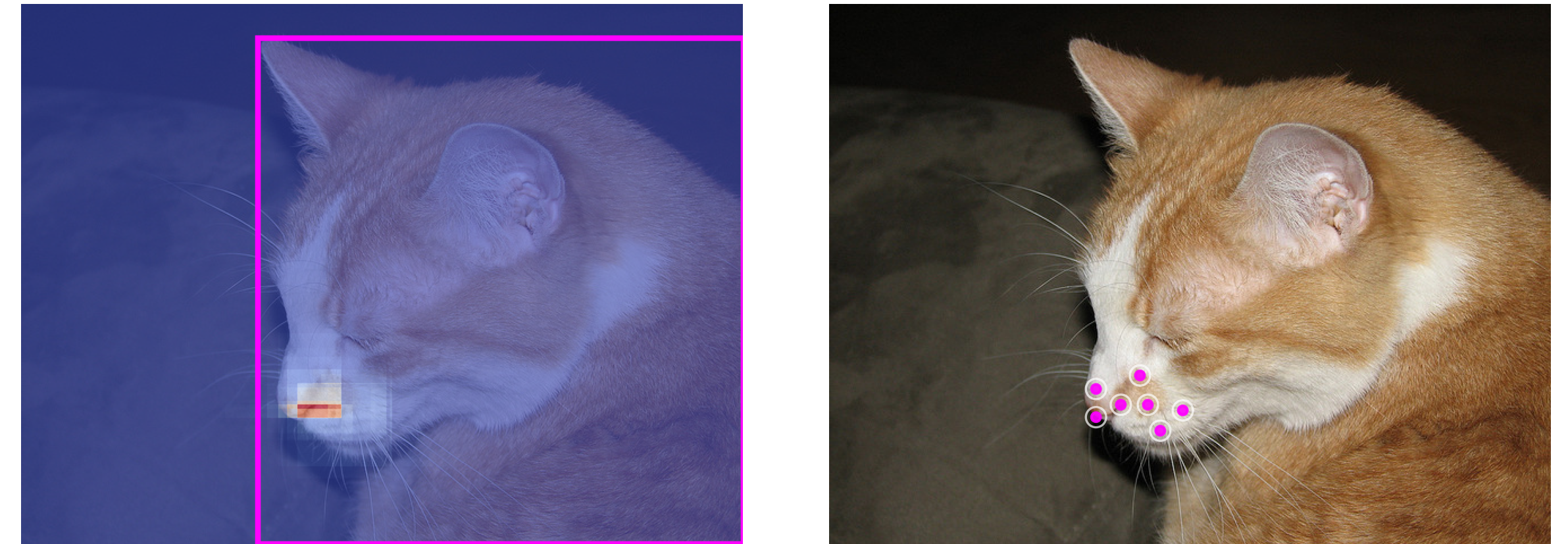} 
    \includegraphics[width=0.48\linewidth]{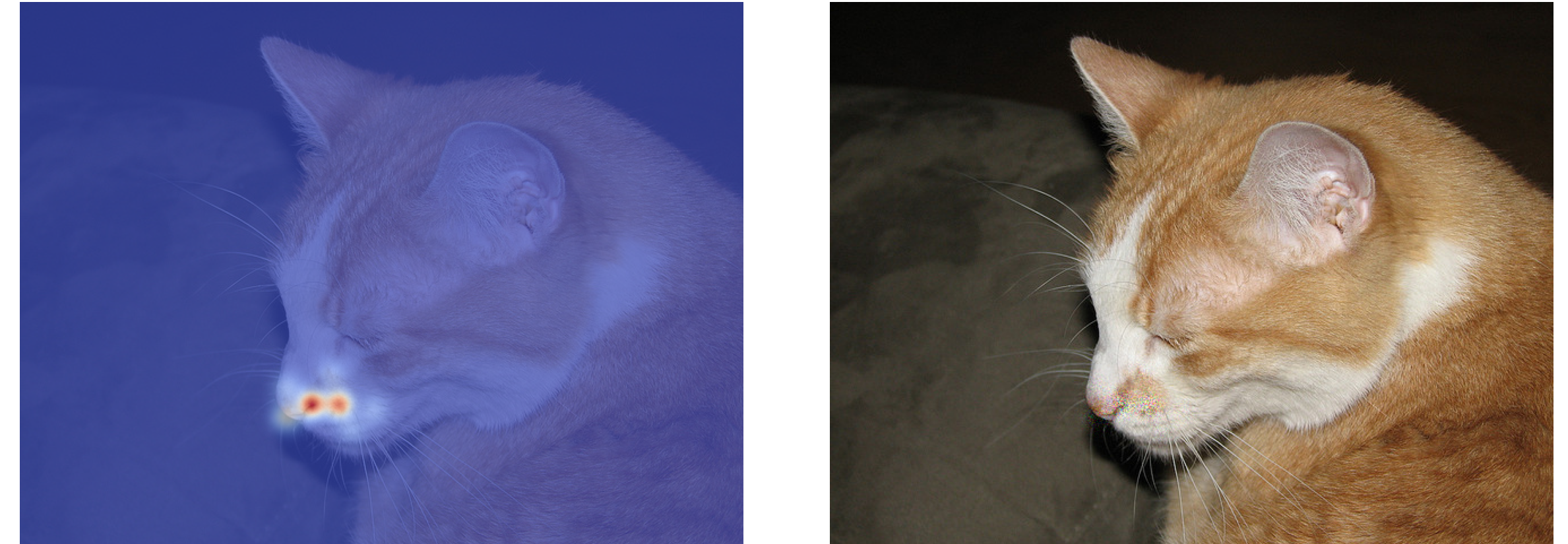}
    \caption{Example of an attack with \blackcattMOG{}. From left to right: original image overlaid with the responsibility restricted to \textit{inside-\msps} and bbox, the top 7 peaks extracted, fitted MoG mask and, finally, the attacked image with no detection.}
    \label{fig:MOGexample}
\end{figure}

\paragraph*{\textbf{\blackcatt}$_{\bm{MoG}}$} is a spatial and contour-aware algorithm that treats the map as a Mixture of Gaussians (MoG). 
\blackcattMOG generates a smooth, spatially coherent perturbation guided by the map rankings. Using the same modes as defined for~\Cref{algo:blended},~\Cref{algo:mog} searches for a given number of peaks and extracts them from within the \msps, using the map as a guide. 
Each peak is treated as the center of a 2D Gaussian kernel whose width reflects local map density. This produces a continuous MoG map:

\begin{equation}\label{eq:gaussian}
P(\mathbf{\mathcal{X}}) = \sum_{i=1}^{k} \alpha_i \exp\!\left(-\frac{\|\mathbf{\mathcal{X}}-\mathbf{p}_i\|^2}{2\sigma_i^2}\right),
\end{equation}

\noindent where $\mathbf{p}_i$ are the selected peaks, $\sigma_i$ controls spatial spread, and $\alpha_i$ scales by the normalized map value at each peak.~\Cref{algo:mog} iterates through a number of different peaks and returns the best-performing solution for each mode.
MoG is a soft mask defining where and how strongly to perturb. We take the Hadamard product (denoted $\odot$) of  $P(\mathbf{\mathcal{X}})$  and the attack perturbation $\delta$:
$\mathcal{X}_p = \mathcal{X} + \delta \odot P(\mathbf{\mathcal{X}})$, which perturbs the image at the location of the peak with the intensity indicated by $P(\mathbf{\mathcal{X}})$. 
\Cref{fig:MOGexample} shows an example of $7$ extracted peaks from inside-\msps, fit to a weighted mixture of 2D Gaussian kernels and attacked with noise.

\begin{algorithm}[t] 
    \caption{$\mathit{BlackCAtt_{MOG}} \\ \,(\mathcal{N}, \mathcal{X}, \text{SIGMAS}, \text{TOPK}, R(\mathcal{X}), M)$ }
    \label{algo:mog}
    \begin{flushleft}
        \textbf{INPUT:}\,\, 
        an object detector $\mathcal{N}$, an image $\mathcal{X}$, 
        a list of standard deviations for the Gaussian kernel SIGMAS, 
           a list of number of peaks \text{TOPK},
        a target map $R(\mathcal{X})$, the \msps, $M$, extracted from $R(\mathcal{X})$ 
        \\
        \textbf{OUTPUT:}\,\, metrics and masks of the best attacks for each goal(no\_pred, change\_pred and added\_new\_pred)
    \end{flushleft}     
        \begin{algorithmic}[1]
            \FORALL{topk $\in \text{TOPK}$}
                \FORALL{$mask \in [\text{outside-\msps}, \text{inside-\msps}, M]$}
                    \STATE peaks $\leftarrow$ topk\_peaks(mask, $r_1$)
                    \STATE Test each $\sigma_i$ in \text{SIGMAS} with:
                    \STATE $P(\mathbf{\mathcal{X}}) \leftarrow$ sum\_peaks(Gaussian2D(peak, $\sigma_i$) $\times$ mask[peak])
                    
                    \STATE $\mathcal{X}_p \leftarrow$ apply\_attack($P(\mathbf{\mathcal{X}}), \mathcal{X})$
                    
                    \STATE preds $\leftarrow \mathcal{N}(\mathcal{X}_p)$
                    \IF{metrics\_improve(preds)}
                        \STATE metrics $\leftarrow$ update\_metrics(preds)
                    \ENDIF
                \ENDFOR
            \ENDFOR
            \RETURN apply\_refinement($\mathcal{X}, P(\mathbf{\mathcal{X}}), \mathcal{N}$, metrics)
        \end{algorithmic}
\end{algorithm}

\section{Experiments}\label{sec:experiments}
In this section we describe the construction of our experiments, including the datasets and object detectors, along with details of hardware and reproducibility. We start with \blackcatt attacks using only label and bbox information, then move onto \blackcatt attacks which also have access to a loss function. We report full results in the supplementary material.

\textbf{Datasets and Models}: We conduct all experiments on the COCO test2017 split. From the full split, we select images that produce exactly one valid detection (\Cref{sec:results}).  To ensure reproducibility, we fix seeds and store detection outputs for all models across the entire dataset. We test three \od{}s representative of different model families: YOLOv11, FASTER R\_CNN and the transformer-based RT-DETR (see the supplementary material for full model details).
All experiments were run on an NVIDIA H100 80GB GPU and 4 Intel Xeon Platinum 8468 CPU.

\textbf{Experiment Setup}: We compare our method against two undirected baselines: (i) a standard global additive-noise attack---\textbf{Noise} and (ii) a boundary-band attack which places spatial disturbances in a bandwidth around the annotated bbox---\textbf{Noise targeted}. \textbf{Noise} perturbs all pixels uniformly across the image. This represents a strong but non-localized perturbation baseline, unconstrained by spatial priors. \textbf{Noise targeted} is a spatially localized attack similar in spirit to other boundary-focused attacks\cite{brendel2018decisionbased}. It therefore tests if merely disrupting the edges of localized regions achieves the attack goal. Both baselines operate with no causal information and no model score and serve as contrastive references for our \msps leveraged attacks. 

For our proposed methods, we restrict perturbations to additive noise for all experiments. This is to maintain comparability with the baselines and isolate the effect of spatial guidance rather than perturbation type. We search over a noise scale of $(5, 40)$ across all experiments and seek to maximize success under the constraints of $L_2$ and LPIPS distance measurements. For \blackcattMOG, we search through the number of peaks in the range $(3,15)$.
We run \blackcattMOG with two black box \xai tools, DRISE and a bbox aware version of \rex. DRISE does not produce \mspss, so we apply \rex{'s} causal extraction algorithm to the DRISE map. We do this to evaluate if \msps priors can influence the efficacy and distortion regardless of the underlying pixel ranking used. 

In the setting where the threat model has access to the scores, we inject restrictions to state-of-the-art (SOTA) attacks by limiting them to causal pixels: Square Attack (SA), Parallel Rectangle Flip Attack (PRFA) and Sparse Attack. We give each tool a maximum of $3500$ iterations of their respective algorithms  and the same for their \blackcatt-variant, enforcing identical $L_0$ and $L_2$ constraints. We also introduce a version of \blackcatt that is optimized using the appropriate loss functions for each attack goal. These are denoted BlackCAtt$_{\text{loss}}$ for $L_2$ adversarial attack and BlackCAtt$_{MoG_{\text{loss}}}$ as a sparse attack. This experiment highlights the effect of the spatial guidance in the presence of stronger score-based optimizer.
\section{Results}\label{sec:results}
\begin{figure*}[t]
    \centering
    \begin{subfigure}[t]{0.32\textwidth}
        \centering
        \pgfplotsset{width=4.55cm, height=3.5cm}
        \centering
        \begin{tikzpicture}
            \begin{axis}[
            ylabel={Success Rate \%},
            xlabel={$L_2$ threshold},
            grid=major,
            xticklabels={},
            extra x ticks={1, 2, 3, 4},
            xmin=1,
            xmax=4,
            extra x tick labels={2/255, 4/255, 8/255, 10/255}]
                \addplot[red,thick] coordinates {
                (1,0.0)
                (2,0.0)
                (3,(0.0)
                (4, 0.04081632653061225)
                };
                \addplot[blue,thick] coordinates {
                (1, 9.122448979591837)
                (2, 18.46938775510204)
                (3, 27.55102040816326)
                (4, 30.0)
                };
                \addplot[green,thick] coordinates {
                (1, 21.428571428571427)
                (2, 35.63265306122449)
                (3, 50.89795918367347)
                (4, 55.95918367346939)
                };
                \addplot[orange,thick] coordinates {
                (1, 14.510204081632653)
                (2, 25.142857142857146)
                (3, 42.44897959183673)
                (4, 47.48979591836734)
                };
                \addplot[purple,thick] coordinates {
                (1, 12.693877551020408)
                (2, 21.857142857142858)
                (3, 37.55102040816327)
                (4, 40.87755102040816)
                };
            \end{axis}
        \end{tikzpicture}
        \caption{\yolo}
    \end{subfigure}
    \hfill
    \begin{subfigure}[t]{0.32\textwidth}
    \pgfplotsset{width=4.55cm, height=3.5cm}
    \begin{tikzpicture}
        \begin{axis}[
        xlabel={$L_2$ threshold},
        grid=major,
        xticklabels={},
        extra x ticks={1, 2, 3, 4},
        xmin=1,
        xmax=4,
        extra x tick labels={2/255, 4/255, 8/255, 10/255}]
            \addplot[red,thick] coordinates {
            (1,0.0)
            (2,0.0)
            (3,(0.0)
            (4, 0.0)
            };
            \addplot[blue,thick] coordinates {
            (1, 7.15)
            (2, 12.7)
            (3, 18.48)
            (4, 20.04)
            };
            \addplot[green,thick] coordinates {
            (1, 14.63)
            (2, 27.0)
            (3, 40.0)
            (4, 44.41)
            };
            \addplot[orange,thick] coordinates {
            (1, 15.19)
            (2, 28.74)
            (3, 47.67)
            (4, 52.59)
            };
            \addplot[purple,thick] coordinates {
            (1, 14.74)
            (2, 27.89)
            (3, 47.89)
            (4, 52.74)
            };
        \end{axis}
    \end{tikzpicture}
    \caption{RT-DETR}
    \end{subfigure}
    \begin{subfigure}[t]{0.3\textwidth}
     \pgfplotsset{width=4.55cm, height=3.5cm}
    \centering
    \begin{tikzpicture}
        \begin{axis}[
        xlabel={$L_2$ threshold},
        grid=major,
        xticklabels={},
        extra x ticks={1, 2, 3, 4},
        xmin=1,
        xmax=4,
        extra x tick labels={2/255, 4/255, 8/255, 10/255}]
            \addplot[red,thick] coordinates {
            (1,0.0)
            (2,0.0)
            (3,(0.0)
            (4, 0.36)
            };
            \addplot[blue,thick] coordinates {
            (1, 30.36)
            (2, 51.09)
            (3, 66.0)
            (4, 68.73)
            };
            \addplot[green,thick] coordinates {
            (1, 51.09)
            (2, 79.64)
            (3, 96.0)
            (4, 98.0)
            };
            \addplot[orange,thick] coordinates {
            (1, 47.27)
            (2, 67.09)
            (3, 87.64)
            (4, 90.55)
            };
            \addplot[purple,thick] coordinates {
            (1, 59.82)
            (2, 80.91)
            (3, 92.73)
            (4, 94.55)
            };
        \end{axis}
    \end{tikzpicture}
    \caption{FASTER-RCNN}
    \end{subfigure}
    \caption{Success rate of different approaches in adding new spurious detection, with different models on COCO dataset, for different thresholds of L$_2$ norm. The different techniques are
     \textcolor{red}{noise}, \textcolor{blue}{targeted noise}, \textcolor{green}{blended},  \textcolor{orange}{DRISE$_{MoG}$} and \textcolor{purple}{MoG}. Note that the y axes differ.}%
    \label{fig:results:yolo}
\end{figure*}

\begin{table}[t]
    \centering
    \caption{Success of different adversarial attacks (in \%) with different thresholds, with YOLO on COCO dataset.}
    \label{tab:YOLO_outcome_sucess}
    \resizebox{\textwidth}{!}{
    \begin{subtable}{0.5\linewidth}\caption{LPIPS $\leq 0.005$}
           \centering
            \begin{tabular}{l||r|r|r}
            \toprule
                & no-pred & change & add \\
            \midrule\midrule
            noise & $0.29$ & $0.12$ & $0.59$  \\
            noise\_targeted & $9.96$ & $4.43$ & $23.08$ \\
            Blended & $28.39$ & $17.2$ & $49.8$  \\
            DRISE\_MoG & $\textbf{32.67}$ & $\textbf{21.82}$ & $\textbf{53.31}$ \\
            MoG & $31.18$ & $18.61$ & $47.53$ \\
            \bottomrule
        \end{tabular}
    \end{subtable}
    \hfill
    \begin{subtable}{0.5\linewidth}\caption{$L_2 \leq \frac{10}{255}$}
        \centering
        \begin{tabular}{l||r|r|r}
    \toprule
          & no-pred & change & add \\
    \midrule\midrule
    noise & $0.04$ & $0.02$ & $0.04$ \\
    noise\_targeted & $11.8$ & $5.84$ & $4.59$ \\
    Blended & $\textbf{32.69}$ & $\textbf{22.55}$ & $25.27$ \\
    DRISE\_MoG & $27.84$ & $18.8$ & $\textbf{26.43}$ \\
    MoG & $25.06$ & $15.1$ & $19.76$ \\
    \bottomrule
    \end{tabular}
    \end{subtable}
    }
\end{table}

\begin{table*}[t]
    \centering
    \caption{The impact of using \blackcatt on YOLO. `Success' reports the attack success rate under the respective norm constraints. The \textbf{bold} values indicate the best value across method type; \underline{underlined} values indicate the best value across all method types within the constraint.}
    \label{tab:comparison}
    \resizebox{\textwidth}{!}{
    \begin{tabular}{ll cccccccc ||c}
    \toprule
    \textbf{Objective} & \textbf{Method} & \textbf{Constraint} &   
    \shortstack{\textbf{Avg.}\\\textbf{Queries}} &
    \shortstack{\textbf{Avg.}\\\bm{$L_0$}} & 
    \shortstack{\textbf{Std.}\\\bm{$L_0$}} & 
    \shortstack{\textbf{Avg.}\\\bm{$L_2$}} & 
    \shortstack{\textbf{Std.}\\\bm{$L_2$}} &
    \shortstack{\textbf{Avg.}\\\textbf{LPIPS}} & 
    \shortstack{\textbf{Std.}\\\textbf{LPIPS}} &
    \shortstack{\textbf{Success}\\(\bm{\%})} \\
    
    \midrule\midrule
    \multirow{6}{*}{\textbf{Remove Pred}}
    & SA & $L_2 \leq 4/255$ & 274.2 & 0.987 &0.019& 0.0148 &0.001& 0.023 &0.019& \underline{\textbf{27.7}} \\
    & SA$_{exp}$ & $L_2 \leq 4/255$ & \textbf{37.3} & \textbf{0.072} &0.097&\textbf{ 0.0145} &0.001& \textbf{0.014} &0.012& 24.0 \\ 
    \cdashline{2-11}
    & PRFA & $L_2 \leq 4/255$ & \textbf{1.03} & 0.989 &0.017& 0.015 &0.000& 0.116 &0.084& \textbf{17.7} \\
    & PRFA$_{exp}$ & $L_2 \leq 4/255$ & 79.97 & \textbf{0.105} &0.107& \underline{\textbf{0.005}} &0.003&\underline{\textbf{0.010}}&0.014& 11.2 \\
    \cdashline{2-11}
    & \blackcatt$_{\text{loss}}$ & $L_2 \leq 4/255$ & 261.3 & \underline{\textbf{0.033}} &0.048& 0.0139 &0.002& 0.015 &0.013& 22.8 \\
    \cmidrule{2-11}
    & SparseRS & $L_0 \leq 0.005$ & 431.7 & 0.005 &0.000& 0.041 &0.002& 0.241 &0.131& \textbf{76.9} \\
    & SparseRS$_{exp}$ & $L_0 \leq 0.005$ & \textbf{267.9} & 0.005 &0.000& 0.041 &0.002& \underline{\textbf{0.065}} &0.045& 64.1 \\
     \cdashline{2-11}
    & BlackCAtt$_{MoG_{\text{loss}}}$ & $L_0 \leq 0.005$ & \underline{\textbf{132.98}} & 0.005 &0.000& \underline{\textbf{0.031}} &0.004 & 0.069 &0.048& 58.2 \\
    \midrule
    
    \multirow{6}{*}{\shortstack{\textbf{Change}\\\textbf{Pred}}}
    & SA & $L_2 \leq 4/255$ & 839.6 & 0.985 &0.028& 0.014 &0.001& 0.029 &0.020& \underline{\textbf{5.4}} \\
    & SA$_{exp}$ & $L_2 \leq 4/255$ & \underline{\textbf{110.1}} & \underline{\textbf{0.056}} &0.037& \textbf{0.013} &0.003& \underline{\textbf{0.015}} &0.014& 1.3 \\
     \cdashline{2-11}
    & PRFA & $L_2 \leq 4/255$ & 3395 & 0.989 &0.001& 0.015&0.000 & 0.065&0.001& 0.2 \\
    & PRFA$_{exp}$ & $L_2 \leq 4/255$ & \textbf{547.7} & \textbf{0.087} &0.085& \underline{\textbf{0.008}} &0.004& \underline{\textbf{0.015}} &0.012& \textbf{0.3} \\
     \cdashline{2-11}
    & BlackCAtt$_{\text{loss}}$ & $L_2 \leq 4/255$ & 702.2 & 0.065 &0.063& 0.013 &0.004& 0.023 &0.026& 1.4 \\
    \cmidrule{2-11}
    & SparseRS & $L_0 \leq 0.005$ & 929.7 & 0.005 &0.000& 0.041 &0.002& 0.218 &0.105& \underline{\textbf{5.6}} \\
    & SparseRS$_{exp}$ & $L_0 \leq 0.005$ & \textbf{641.4} & 0.005 &0.000& 0.040 &0.002& \underline{\textbf{0.066}} &0.043& 3.2 \\
     \cdashline{2-11}
    & BlackCAtt$_{MoG_{\text{loss}}}$  &  $L_0 \leq 0.005$ & \underline{\textbf{203.5}} & 0.005 &0.000& \underline{\textbf{0.031}} &0.005& 0.081 &0.055& 4.6 \\
    \midrule
    
    \multirow{6}{*}{\shortstack{\textbf{Add}\\\textbf{New Pred}}}
    & SA & $L_2 \leq 4/255$ & 106.7 & 0.989 & 0.026 & 0.0153 & 0.001 &0.029 & 0.027 & \underline{\textbf{79.2}} \\
    & SA$_{exp}$ & $L_2 \leq 4/255$ & \underline{\textbf{23.11}} & \textbf{0.051} &  0.059 & \textbf{0.0133} & 0.002 & \textbf{0.011} & 0.011&70.4 \\
      \cdashline{2-11}
    & PRFA & $L_2 \leq 4/255$ & 75.6 & 0.989 &0.036 & 0.0155&0.000 & 0.083 &0.070& 55.6 \\
    & PRFA$_{exp}$ & $L_2 \leq 4/255$ & \textbf{62.4} & \textbf{0.045}&0.047 & \underline{\textbf{0.0029}} &0.002& \underline{\textbf{0.002}} &0.004& \textbf{56.4} \\
    \cdashline{2-11}
    & BlackCAtt$_{\text{loss}}$ & $L_2 \leq 4/255$ & 41.9 & \underline{\textbf{0.007}} &0.024& 0.0043 &0.006& 0.003&0.008 & 68.3 \\
    \cmidrule{2-11}
    & SparseRS & $L_0 \leq 0.005$ & \textbf{104.4} & 0.005 &0.000 & \textbf{0.0405} &0.002 & 0.223 &0.126& \underline{\textbf{88.2}} \\
    & SparseRS$_{exp}$ & $L_0 \leq 0.005$ & 173.4 & 0.005 &0.000& 0.0407 &0.002& \textbf{0.059} &0.039& 73.7 \\
      \cdashline{2-11}
    & BlackCAtt$_{MoG_{\text{loss}}}$  &  $L_0 \leq 0.005$ & 56.4 & 0.005 &0.000& \underline{\textbf{0.0292}} &0.004& \underline{\textbf{0.042}} &0.031& 65.3 \\
    \bottomrule
    \end{tabular}
    }
\end{table*}

Most black-box attacks require continuous feedback, such as bounding box coordinates or confidence scores, to optimize a loss function. In contrast, the native \blackcattGreedy{} and \blackcattMOG{} algorithms operate with access to only the discrete class labels.
\Cref{tab:YOLO_outcome_sucess} shows success rates (\%) on \yolo for the three attack goals under two distance thresholds (LPIPS $\leq 0.05$ and $L_2 \leq \frac{10}{255}$). Despite the lack of a continuous gradient or loss signal, \blackcattGreedy{} and \blackcattMOG{} substantially outperform undirected baselines, being at least $2.3\times$ better. Noise and targeted noise rarely exceed 10\% success at lower thresholds, whereas \blackcatt{} methods achieve up to \textbf{53\%} for adding new detections and over \textbf{30\%} no-prediction success, demonstrating that perturbation of causal pixels leads to more effective attack at equal perceptual cost. 


\Cref{fig:results:yolo} shows that success rates steadily increase with the increase in permitted distortion. \blackcatt attacks show substantially higher success rates for producing added spurious detections compared to the baselines. This pattern is present for all three attack goals and architectures, with Faster-R-CNN most easily driven to add detections. RT-DETR is the most robust overall. Notably, \blackcattMOG yields high success even on RT-DETR, suggesting the MoG method aligns well with the compositionality of transformer architectures~\cite{Jiang_2024_CVPR}.  

\textbf{Enhancing Score-Based SOTA with \mspss}: We now show that using causal spatial priors allows us to improve state-of-the-art (SOTA) black-box attacks. We compare Square Attack (SA)~\cite{andriushchenko2020square}, Parallel Rectangle Flip Attack (PRFA)~\cite{prfa} and Sparse-RS~\cite{croce2019sparse} against their \msps-guided variants (denoted $_{exp}$). The results are summarized in~\Cref{tab:comparison}.
Note that $_{exp}$ variants have an additional cost of $\approx2125$ model calls to generate \mspss. However, this is only computed once per image and can then be used for all the other attack modes, essentially amortizing the cost. The average number of queries for \blackcatt-variants is lower than the original tools in most cases (underlined values in~\Cref{tab:comparison}).

We show that we are able to reduce both $L_0$ and LPIPS by an order of magnitude, with the trade-off with being a slightly lower success rate some cases. We see a similar pattern occur with the RT-DETR. \Cref{tab:cdf-loss} shows that enforcing \mspss(dashed lines and green line) consistently shifts each line upwards, meaning that these variants are able to achieve higher success at substantially lower distortion thresholds. Note that \blackcatt{} (green line) dominates almost always at lower distortions.

Using \mspss as spatial priors and incorporating information regarding pixel importance from the initial maps can be actively exploited to achieve our stated attack goals. Disrupting the minimal sufficient causes for an output is a \emph{principled, transparent, explainable} way to corrupt that output.

\begin{figure}[t]
    \centering
\begin{subfigure}{0.32\textwidth}
\centering
\begin{tikzpicture}
\begin{axis}[
    width=\linewidth,
    height=4cm,
    xlabel={$L_2$},
    ylabel={Success rate},
    grid=major,
    legend to name=sharedlegend,
    legend columns=5,
    legend style={
        font=\small,
        text=black,
        draw=none
    }
]
\addplot[thick, blue, mark=*] coordinates {(0.00392156862745098, 0/1000) (0.00784313725490196, 0/1000) (0.011764705882352941, 8/1000) (0.01568627450980392, 792/1000) (0.0196078431372549, 811/1000)};
\addlegendentry{SA}
\addplot[thick, blue, dashed, mark=square*]  coordinates {(0.00392156862745098, 7/1000) (0.00784313725490196, 40/1000) (0.011764705882352941, 124/1000) (0.01568627450980392, 704/1000) (0.0196078431372549, 704/1000)};
\addlegendentry{$SA_{exp}$}
\addplot[thick, red, mark=triangle*]  coordinates {(0.00392156862745098, 0/1000) (0.00784313725490196, 0/1000) (0.011764705882352941, 0/1000) (0.01568627450980392, 272/1000) (0.0196078431372549, 406/1000)};
\addlegendentry{PRFA}
\addplot[thick, red, dashed, mark=diamond*]  coordinates {(0.00392156862745098, 370/1000) (0.00784313725490196, 502/1000) (0.011764705882352941, 507/1000) (0.01568627450980392, 508/1000) (0.0196078431372549, 508/1000)};
\addlegendentry{PRFA exp}
\addplot[thick, green!60!black, mark=star]  coordinates {(0.00392156862745098, 462/1000) (0.00784313725490196, 483/1000) (0.011764705882352941, 517/1000) (0.01568627450980392, 683/1000) (0.0196078431372549, 683/1000)};
\addlegendentry{\blackcatt{}}
\end{axis}
\end{tikzpicture}
\end{subfigure}
\begin{subfigure}{0.32\textwidth}
\centering
\centering
\begin{tikzpicture}
\begin{axis}[
    width=\linewidth,
    height=4cm,
    xlabel={$L_0$},
    ylabel={Success rate},
    grid=major,
]
\addplot[thick, blue, mark=*] coordinates {(0.02, 0/1000) (0.01, 0/1000) (0.006666666666666667, 0/1000) (0.005, 0/1000) (0.004, 0/1000)};
\addplot[thick, blue, dashed, mark=square*]  coordinates {(0.02, 227/1000) (0.01, 142/1000) (0.006666666666666667, 93/1000) (0.005, 71/1000) (0.004, 60/1000)};
\addplot[thick, red, mark=triangle*]  coordinates {(0.02, 0/1000) (0.01, 0/1000) (0.006666666666666667, 0/1000) (0.005, 0/1000) (0.004, 0/1000)};
\addplot[thick, red, dashed, mark=diamond*]  coordinates {(0.02, 174/1000) (0.01, 106/1000) (0.006666666666666667, 71/1000) (0.005, 57/1000) (0.004, 48/1000)};
\addplot[thick, green!60!black, mark=star]  coordinates {(0.02, 624/1000) (0.01, 581/1000) (0.006666666666666667, 551/1000) (0.005, 528/1000) (0.004, 518/1000)};
\end{axis}
\end{tikzpicture}
\end{subfigure}
\centering
\begin{subfigure}{0.32\textwidth}
    \begin{tikzpicture}
\begin{axis}[
    width=\linewidth,
    height=4cm,
    xlabel={$LPIPS$},
    ylabel={Success rate},
    grid=major,
]
\addplot[thick, blue, mark=*] coordinates {(0.02, 0.0/1000) (0.01, 0.0/1000) (0.006666666666666667, 0.0/1000) (0.005, 0.0/1000) (0.004, 0.0/1000)};
\addplot[thick, blue, dashed, mark=square*]  coordinates {(0.02, 603/1000) (0.01, 411/1000) (0.006666666666666667, 288/1000) (0.005, 211/1000) (0.004, 164/1000)};
\addplot[thick, red, mark=triangle*]  coordinates {(0.02, 52/1000) (0.01, 7/1000) (0.006666666666666667, 2/1000) (0.005, 0/1000) (0.004, 0/1000)};
\addplot[thick, red, dashed, mark=diamond*]  coordinates {(0.02, 500/1000) (0.01, 490/1000) (0.006666666666666667, 475/1000) (0.005, 447/1000) (0.004, 430/1000)};
\addplot[thick, green!60!black, mark=star]  coordinates {(0.02, 658/1000) (0.01, 604/1000) (0.006666666666666667, 566/1000) (0.005, 540/1000) (0.004, 525/1000)};
\end{axis}
\end{tikzpicture}
\end{subfigure}
\ref{sharedlegend}
\caption{Cumulative success under $L_2, L_0$ and LPIPS distortion constraints. Dashed lines denote \blackcatt-guided variants.}
\label{tab:cdf-loss}
\end{figure}

\textbf{Limitations}:
\blackcatt requires \mspss. To the best of our knowledge, only \rex provides \mspss directly. 
The \rex algorithm is modular, however, allowing us to apply its \msps extraction mechanism to saliency
maps produced by other tools (such as DRISE). The extraction algorithm works best when the saliency
map is not `noisy', \ie containing spurious or misleading saliency attribution~\cite{chockler2024causal}. In the event that the \xai tool provides extremely noisy rankings, \blackcatt will not function well. This is a limitation of the \xai tool.
Absolute success rates and distortion budgets vary with architecture and domain: the transformer-based detector (RT-DETR) was empirically more robust to localized perturbations than single-stage detectors. Our experiments are on COCO; more cross-domain validation is required to claim generality.

The problem of \emph{over-determination} is well known in the literature of causality~\cite{Hal19}. As shown in~\cite{CKK25} for image classifiers, many images have multiple, independent, \mspss. \blackcatt only attacks one \msps as we know of no tool capable of produces multiple \mspss for \od. 
If the abundance of \mspss for \od is similar to that observed for image classifiers, this would go some way to explain why our causal attacks are not always successful: we have not discovered and attacked all relevant causes. In our main experiments, we limit ourselves to images with one detection only. This was to make result analysis easier. We also ran a small experiment ($200$ images) with multiple detections and found results similar to those present here (see supplementary material).

\section{Related Work}\label{sec:relwork}
 Due to the lack of space, we only overview the most relevant papers on \xai for object detectors, adversarial attacks on object detectors, and explainable attacks for object detectors. 
\blackcatt works in different modes, either decision-based or score-based attacks. We do not know of many other approach that does this. Decision-based attacks use label information only~\cite{brendel2018decisionbased,Chen2019HopSkipJumpAttackAQ}, however none are explainable.

\textbf{\xai for Images and Object Detectors}: There is a large body of work on explanations for image classifiers.  
Propagation-based explanation methods back-propagate a model's decision to the input layer to determine the weight of each input feature for the decision~\cite{springenberg2015striving,
sundararajan2017axiomatic,bach2015pixel,shrikumar2017learning,
nam2020relative}. \gradcam only needs one backward pass and propagates the class-specific gradient into the final convolutional layer of a DNN to coarsely highlight important regions of an input image~\cite{CAM}. 
Perturbation-based explanation approaches introduce perturbations to the 
input space directly in search for an explanation~\cite{lime,lundberg2017unified}.
None of the tools mentioned above work natively with \yolo, or any other object detector. \fsod~\cite{Kuroki2024FastDetection} adapts the loss function of \fastshap~\cite{fastshap} to train an explainer model with a UNet-inspired architecture for classification and location of the detected object~\cite{Jethani2022FastSHAP:Estimation}. This method however introduces a new model which itself would need explaining.

\textbf{Object Detection Models}: \od models come in several families that differ in architecture, speed and output structure and these differences affect the output of \xai tools. Single-stage, real-time detectors such as YOLO~\cite{yolo} and SSD~\cite{ssd} perform object localization and classification in a single unified pass, making them fast and suitable for real-time inference in, \eg, autonomous driving. Two stage object detectors, on the other hand, break the task into two distinct steps, generating spare sets of \emph{region proposals} and using a separate network to classify these regions. Transformer-based object detectors, like \underline{DE}tection \underline{TR}ansformer (DETR) and Real Time-DETR (RT-DETR)~\cite{lv2023detrs}, use a transformer encoder-decoder network, which removes the need for region proposals and non-maximum suppression (NMS). 

\textbf{Adversarial Attacks on Object Detectors}: Object detectors can fail catastrophically under small changes to their inputs. 
In a safety-critical scenario, 
a single missed detection may produce real harm. Attack goals for detectors commonly include removing detections, altering labels and adding spurious detections~\cite{surveyaaforobjectdtection}. 
Attack methods vary between white-box pixel-wise optimizers and black-box gradient-free optimizers NES\cite{ilyas2018black}, Square~\cite{andriushchenko2020square}, spare~\cite{croce2019sparse} and one-pixel~\cite{onepixel2019} attacks, context-aware~\cite{cai2022context} and patch-based~\cite{liu2018dpatch} attacks that place sticks in the scene. Standard evaluations include per-goal success rates under distortion budgets such as LPIPS, $L_1$, $L_2$ and $L_0$ and query or time budgets. Crucially, most such attacks demonstrate that detectors are vulnerable but do not explain \emph{why} a particular perturbation succeeds.


\textbf{Saliency in Adversarial Analysis}: There is a growing line of work that focus on the relationship between saliency methods and adversarial behavior~\cite{liu2021adversarial, ignatiev2019relating, mangla2020saliency, wang2018adversarial, gu2019saliencymethodsexplainingadversarial, papernot2015, Gao_2025_ICCV}. There is a theoretical relationship established between adversarial examples and explanations~\cite{ignatiev2019relating}, linked by a generalized form of hitting set duality. The formal logic requirement is that the ``counterexample'' must be a subset-minimal set of features that still guarantees the associated prediction or prediction error. Certain attack methodologies, such as the Jacobian-based Saliency Map Attack (JSMA)~\cite{JSMA}, explicitly define a saliency map based on the Jacobian matrix to identify input features that have the greatest influence on the model output. Prior work uses white box methods, such as Grad-CAM~\cite{wang2022adversarial}, and shows that feature statistics extracted from saliency maps, such as spatial concentration and entropy can reliably distinguish adversarial from clean inputs and thereby mitigate such attacks. 

The attacker model of these methods is unrealistic, given that the saliency generating processes are all white-box and require direct access either to the model internals or its gradient. \blackcatt, in contrast, is entirely black-box. Saliency Attack~\cite{dai2023} is probably the closest in spirit to our approach, however it is not intended for \od, so direct comparison is not possible. It attacks saliency regions by applying colored patches to an image. This has a high success rate, but is also computationally expensive
and not inherently explainable.

\section{Conclusions and Future Work}\label{sec:conclusion}
To the best of our knowledge, \blackcatt is the first explainable decision-based attack method for image classifiers. In this paper, we have focused on object detectors. \blackcatt is a meta-algorithm: we show that it also improves score-based attack techniques. 
Our approach is principled, transparent, explainable, and model-agnostic, treating the \od as a black box.  We show that parts of a cause are frequently outside the bounding box and that these causal pixels can then be exploited to create successful adversarial attacks. Moreover, we show that this works on three different architectures and for three different attack goals.

We will extend \blackcatt to manage the interaction between multiple \mspss. This will require a version of \rex that produces multiple \mspss for \od. We consider only sufficient pixel sets. \rex can also compute sufficient and necessary pixel sets~\cite{kelly2025causal}. These have more information than \mspss and should increase the success rate of \blackcatt.  Our findings suggest several practical directions for improving robustness and deployment practices. \blackcatt may be used in adversarial training to focus regularization on \mspss and surrounding context. Moreover, systematically analyzing \mspss that lie outside the bounding box may reveal more information about model behavior and bounding box calculation. 




%
%
\bibliographystyle{splncs04}
\bibliography{all}

@inproceedings{REOW2025,
    author = {Justyn Rodrigues and Kris Ehinger and Oliver Obst and Rosalind
              Wang},
    title = {Do Explanations Expose Bias? How Saliency Maps Affect Judgements of
             Biased Face-Recognition Models},
    booktitle = {Proceedings of the 25th European Conference on Artificial
                 Intelligence (ECAI 2025)},
    year = {2025},
}

@article{ignatiev2019relating,
    title = {On relating explanations and adversarial examples},
    author = {Ignatiev, Alexey and Narodytska, Nina and Marques-Silva, Joao},
    journal = {Advances in neural information processing systems},
    volume = {32},
    year = {2019},
}

@misc{gu2019saliencymethodsexplainingadversarial,
    title = {Saliency Methods for Explaining Adversarial Attacks},
    author = {Jindong Gu and Volker Tresp},
    year = {2019},
    eprint = {1908.08413},
    archivePrefix = {arXiv},
    primaryClass = {cs.CV},
    url = {https://arxiv.org/abs/1908.08413},
}

@misc{surveyaaforobjectdtection,
    title = {A Survey and Evaluation of Adversarial Attacks for Object Detection
             },
    author = {Khoi Nguyen Tiet Nguyen and Wenyu Zhang and Kangkang Lu and Yuhuan
              Wu and Xingjian Zheng and Hui Li Tan and Liangli Zhen},
    year = {2025},
    eprint = {2408.01934},
    archivePrefix = {arXiv},
    primaryClass = {cs.CV},
}

@inproceedings{kelly2025big,
    title = {I am big, you are little; I am right, you are wrong},
    author = {Kelly, David A and Chanchal, Akchunya and Blake, Nathan},
    booktitle = {Proceedings of the IEEE/CVF International Conference on
                 Computer Vision},
    pages = {817--826},
    year = {2025},
}

@inproceedings{ssd,
    author = {Wei Liu and Dragomir Anguelov and Dumitru Erhan and Christian
              Szegedy and Scott E. Reed and Cheng{-}Yang Fu and Alexander C. Berg
              },
    title = {{SSD:} Single Shot MultiBox Detector},
    booktitle = {Proceedings of European Conference in Computer Vision {ECCV},
                 Part {I}},
    series = {Lecture Notes in Computer Science},
    volume = {9905},
    pages = {21--37},
    publisher = {Springer},
    year = {2016},
}

@inproceedings{CKK25,
    title = {Multiple Different Explanations for Image Classifiers},
    author = {Chockler, Hana and Kelly, David A. and Kroening, Daniel},
    booktitle = {{ECAI} European Conference on Artificial Intelligence},
    year = {2025},
}

@article{Dic45,
    author = {Dice, Lee R.},
    title = {Measures of the Amount of Ecologic Association Between Species},
    journal = {Ecology},
    volume = 26,
    issue = 3,
    pages = {297–-302},
    year = 1945,
}

@book{Hal19,
    author = {Joseph Y. Halpern},
    title = {Actual Causality},
    year = 2019,
    publisher = {The MIT Press},
}

@inproceedings{springenberg2015striving,
    title = {Striving for Simplicity: The All Convolutional Net},
    author = {Jost Tobias Springenberg and Alexey Dosovitskiy and Thomas Brox
              and Martin A. Riedmiller},
    booktitle = {ICLR (Workshop Track)},
    url = {http://arxiv.org/abs/1412.6806},
    year = {2015},
}

@article{bach2015pixel,
    title = {On Pixel-wise Explanations for Non-linear Classifier Decisions by
             Layer-wise Relevance Propagation},
    author = {Bach, Sebastian and Binder, Alexander and Montavon, Gr{\'e}goire
              and Klauschen, Frederick and M{\"u}ller, Klaus-Robert and Samek,
              Wojciech},
    journal = {PLOS One},
    volume = {10},
    number = {7},
    year = {2015},
    publisher = {Public Library of Science},
}

@inproceedings{nam2020relative,
    title = {Relative Attributing Propagation: Interpreting the Comparative
             Contributions of Individual Units in Deep Neural Networks},
    author = {Nam, Woo-Jeoung and Gur, Shir and Choi, Jaesik and Wolf, Lior and
              Lee, Seong-Whan},
    booktitle = {AAAI Conference on Artificial Intelligence},
    volume = {34},
    pages = {2501--2508},
    year = {2020},
}

@inproceedings{sundararajan2017axiomatic,
    title = {Axiomatic Attribution for Deep Networks},
    author = {Sundararajan, Mukund and Taly, Ankur and Yan, Qiqi},
    booktitle = {International Conference on Machine Learning},
    pages = {3319--3328},
    year = {2017},
    organization = {PMLR},
}

@article{CH04,
    author = {Hana Chockler and Joseph Y. Halpern},
    title = {Responsibility and Blame: {A} Structural-Model Approach},
    journal = {J. Artif. Intell. Res.},
    volume = {22},
    pages = {93--115},
    year = {2004},
}

@article{Chen2019HopSkipJumpAttackAQ,
    title = {HopSkipJumpAttack: A Query-Efficient Decision-Based Attack},
    author = {Jianbo Chen and Michael I. Jordan},
    journal = {2020 IEEE Symposium on Security and Privacy (SP)},
    year = {2019},
    pages = {1277-1294},
}

@inproceedings{brendel2018decisionbased,
    title = {Decision-Based Adversarial Attacks: Reliable Attacks Against
             Black-Box Machine Learning Models},
    author = {Wieland Brendel and Jonas Rauber and Matthias Bethge},
    booktitle = {International Conference on Learning Representations},
    year = {2018},
}

@article{chockler2024causal,
    title = {Causal explanations for image classifiers},
    author = {Chockler, Hana and Kelly, David A and Kroening, Daniel and Sun,
              Youcheng},
    journal = {Journal of Artificial Intelligence Research},
    year = {2026},
}

@inproceedings{petsiuk2018rise,
    author = {Vitali Petsiuk and Abir Das and Kate Saenko},
    title = {{RISE:} Randomized Input Sampling for Explanation of Black-box
             Models},
    year = {2018},
    booktitle = {British Machine Vision Conference ({BMVC})},
    publisher = {{BMVA} Press},
}

@inproceedings{yolo,
    author = {Joseph Redmon and Santosh Kumar Divvala and Ross B. Girshick and
              Ali Farhadi},
    title = {You Only Look Once: Unified, Real-Time Object Detection},
    booktitle = {Proceedings of CVPR},
    pages = {779--788},
    year = {2016},
}

@inproceedings{shrikumar2017learning,
    title = {Learning Important Features Through Propagating Activation
             Differences},
    author = {Shrikumar, Avanti and Greenside, Peyton and Kundaje, Anshul},
    booktitle = {International Conference on Machine Learning (ICML)},
    volume = 70,
    publisher = {{PMLR}},
    pages = {3145--3153},
    year = {2017},
}

@inproceedings{CAM,
    title = {Grad-{CAM}: Visual Explanations from Deep Networks via
             Gradient-based Localization},
    author = {Selvaraju, Ramprasaath R. and Cogswell, Michael and Das, Abhishek
              and Vedantam, Ramakrishna and Parikh, Devi and Batra, Dhruv},
    booktitle = {International Conference on Computer Vision (ICCV)},
    publisher = {IEEE},
    pages = {618--626},
    year = {2017},
}

@inproceedings{lin2014microsoft,
    title = {Microsoft {COCO}: Common objects in context},
    author = {Lin, Tsung-Yi and Maire, Michael and Belongie, Serge and Hays,
              James and Perona, Pietro and Ramanan, Deva and Doll{\'a}r, Piotr
              and Zitnick, C Lawrence},
    booktitle = {European conference on computer vision},
    pages = {740--755},
    year = {2014},
    organization = {Springer},
}

@inproceedings{lundberg2017unified,
    title = {A Unified Approach to Interpreting Model Predictions},
    author = {Lundberg, Scott M. and Lee, Su-In},
    booktitle = {Advances in Neural Information Processing Systems (Neur{IPS})},
    pages = {4765--4774},
    volume = {30},
    year = {2017},
}

@inproceedings{lime,
    author = {Marco Tulio Ribeiro and Sameer Singh and Carlos Guestrin},
    title = {``{W}hy Should {I} Trust You?'' {E}xplaining the Predictions of Any
             Classifier},
    booktitle = {Knowledge Discovery and Data Mining (KDD)},
    pages = {1135--1144},
    publisher = {{ACM}},
    year = {2016},
}

@inproceedings{wang2018adversarial,
    title = {Adversarial Sample Detection for Deep Neural Network through Model
             Mutation Testing},
    author = {Wang, Jingyi and Dong, Guoliang and Sun, Jun and Wang, Xinyu and
              Zhang, Peixin},
    booktitle = {Proceedings of the 41st International Conference on Software
                 Engineering},
    year = {2019},
    organization = {ACM},
}

@article{JSMA,
    Author = {Nicolas Papernot and Patrick D. McDaniel and Somesh Jha and Matt
              Fredrikson and Z. Berkay Celik and Ananthram Swami},
    Bibsource = {dblp computer science bibliography, http://dblp.org},
    Journal = {CoRR},
    Title = {The Limitations of Deep Learning in Adversarial Settings},
    Url = {http://arxiv.org/abs/1511.07528},
    Volume = {abs/1511.07528},
    Year = {2015},
}

@inproceedings{Gao_2025_ICCV,
    author = {Gao, Zhenghao and Xu, Shengjie and Li, Zijing and Chen, Meixi and
              Yu, Chaojian and Shao, Yuanjie and Gao, Changxin},
    title = {FastJSMA: Accelerating Jacobian-based Saliency Map Attacks through
             Gradient Decoupling},
    booktitle = {Proceedings of the IEEE/CVF International Conference on
                 Computer Vision (ICCV)},
    month = {October},
    year = {2025},
    pages = {1506-1515},
}

@article{papernot2015,
    title = {The Limitations of Deep Learning in Adversarial Settings},
    author = {Nicolas Papernot and Patrick Mcdaniel and Somesh Jha and Matt
              Fredrikson and Z. Berkay Celik and Ananthram Swami},
    journal = {2016 IEEE European Symposium on Security and Privacy (EuroS\&P)},
    year = {2015},
    pages = {372-387},
}

@article{liu2021adversarial,
    title = {Adversarial attacks and defenses: An interpretation perspective},
    author = {Liu, Ninghao and Du, Mengnan and Guo, Ruocheng and Liu, Huan and
              Hu, Xia},
    journal = {ACM SIGKDD Explorations Newsletter},
    volume = {23},
    number = {1},
    pages = {86--99},
    year = {2021},
    publisher = {ACM New York, NY, USA},
}

@inproceedings{mangla2020saliency,
    title = {On saliency maps and adversarial robustness},
    author = {Mangla, Puneet and Singh, Vedant and Balasubramanian, Vineeth N},
    booktitle = {joint European conference on machine learning and knowledge
                 discovery in databases},
    pages = {272--288},
    year = {2020},
    organization = {Springer},
}

@article{Yahn2025AdversarialAP,
    title = {Adversarial Attention Perturbations for Large Object Detection
             Transformers},
    author = {Zachary Yahn and Selim Furkan Tekin and Fatih Ilhan and Sihao Hu
              and Tiansheng Huang and Yichang Xu and Margaret Loper and Ling Liu},
    journal = {ArXiv},
    year = {2025},
    volume = {abs/2508.02987},
}

@article{wang2022adversarial,
    title = {Adversarial example detection based on saliency map features},
    author = {Wang, Shen and Gong, Yuxin},
    journal = {Applied Intelligence},
    volume = {52},
    number = {6},
    pages = {6262--6275},
    year = {2022},
    publisher = {Springer},
}

@inproceedings{simba,
    author = {Narodytska, Nina and Kasiviswanathan, Shiva},
    booktitle = {2017 IEEE Conference on Computer Vision and Pattern Recognition
                 Workshops (CVPRW)},
    title = {Simple Black-Box Adversarial Attacks on Deep Neural Networks},
    year = {2017},
    volume = {},
    number = {},
    pages = {1310-1318},
    doi = {10.1109/CVPRW.2017.172},
}

@article{odattacks2025,
    author = {Nguyen, Khoi Nguyen Tiet and Zhang, Wenyu and Lu, Kangkang and Wu,
              Yu-Huan and Zheng, Xingjian and Li Tan, Hui and Zhen, Liangli},
    journal = {IEEE Transactions on Neural Networks and Learning Systems},
    title = {A Survey and Evaluation of Adversarial Attacks in Object Detection},
    year = {2025},
    volume = {36},
    number = {9},
    pages = {15706-15722},
}

@article{onepixel2019,
    author = {Su, Jiawei and Vargas, Danilo Vasconcellos and Sakurai, Kouichi},
    journal = {IEEE Transactions on Evolutionary Computation},
    title = {One Pixel Attack for Fooling Deep Neural Networks},
    year = {2019},
    volume = {23},
    number = {5},
    pages = {828-841},
}

@inproceedings{zhang25a,
    title = {{Saliency Maps Give a False Sense of Explanability to Image
             Classifiers}: {A}n Empirical Evaluation across Methods and Metrics},
    author = {Zhang, Hanwei and Figueroa, Felipe Torres and Hermanns, Holger},
    booktitle = {Proceedings of the 16th Asian Conference on Machine Learning},
    pages = {479--494},
    year = {2025},
    editor = {Nguyen, Vu and Lin, Hsuan-Tien},
    volume = {260},
    series = {Proceedings of Machine Learning Research},
    month = {05--08 Dec},
    publisher = {PMLR},
}

@inproceedings{drise,
    author = {Petsiuk, Vitali and Jain, Rajiv and Manjunatha, Varun and Morariu,
              Vlad I. and Mehra, Ashutosh and Ordonez, Vicente and Saenko, Kate},
    booktitle = {2021 IEEE/CVF Conference on Computer Vision and Pattern
                 Recognition (CVPR)},
    title = {Black-box Explanation of Object Detectors via Saliency Maps},
    year = {2021},
    volume = {},
    number = {},
    pages = {11438-11447},
}

@inproceedings{fastshap,
    title = {Fast{SHAP}: Real-Time Shapley Value Estimation},
    author = {Neil Jethani and Mukund Sudarshan and Ian Connick Covert and Su-In
              Lee and Rajesh Ranganath},
    booktitle = {International Conference on Learning Representations},
    year = {2022},
}

@article{Kuroki2024FastDetection,
    title = {{Fast Explanation Using Shapley Value for Object Detection}},
    year = {2024},
    journal = {IEEE Access},
    author = {Kuroki, Michihiro and Yamasaki, Toshihiko},
    pages = {31047--31054},
    volume = {12},
    publisher = {Institute of Electrical and Electronics Engineers Inc.},
}

@inproceedings{Jethani2022FastSHAP:Estimation,
    title = {{FastSHAP: Real-Time Shapley Value Estimation}},
    year = {2022},
    booktitle = {International Conference on Learning Representations},
    author = {Jethani, Neil and Sudarshan, Mukund and Connick Covert, Ian and
              Lee, Su-In and Ranganath, Rajesh},
    month = {7},
    url = {http://arxiv.org/abs/2107.07436},
    arxivId = {2107.07436},
}

@article{kelly2025causal,
    title = {Sufficient, Necessary and Complete Explanations in Image
             Classification},
    author = {Kelly, David A and Chockler, Hana},
    journal = {arXiv preprint arXiv:2507.23497},
    year = {2025},
}

@inproceedings{zhang2018perceptual,
    title = {The Unreasonable Effectiveness of Deep Features as a Perceptual
             Metric},
    author = {Zhang, Richard and Isola, Phillip and Efros, Alexei A and
              Shechtman, Eli and Wang, Oliver},
    booktitle = {CVPR},
    year = {2018},
}

@inproceedings{Jiang_2024_CVPR,
    author = {Jiang, Mingqi and Khorram, Saeed and Fuxin, Li},
    title = {Comparing the Decision-Making Mechanisms by Transformers and CNNs
             via Explanation Methods},
    booktitle = {Proceedings of the IEEE/CVF Conference on Computer Vision and
                 Pattern Recognition (CVPR)},
    month = {June},
    year = {2024},
    pages = {9546-9555},
}

@misc{lv2023detrs,
    title = {DETRs Beat YOLOs on Real-time Object Detection},
    author = {Wenyu Lv and Shangliang Xu and Yian Zhao and Guanzhong Wang and
              Jinman Wei and Cheng Cui and Yuning Du and Qingqing Dang and Yi Liu
              },
    year = {2023},
    eprint = {2304.08069},
    archivePrefix = {arXiv},
    primaryClass = {cs.CV},
}

@inproceedings{croce2019sparse,
    title = {Sparse and imperceivable adversarial attacks},
    author = {Croce, Francesco and Hein, Matthias},
    booktitle = {Proceedings of the IEEE/CVF international conference on
                 computer vision},
    pages = {4724--4732},
    year = {2019},
}

@inproceedings{ilyas2018black,
    title = {Black-box adversarial attacks with limited queries and information},
    author = {Ilyas, Andrew and Engstrom, Logan and Athalye, Anish and Lin,
              Jessy},
    booktitle = {International conference on machine learning},
    pages = {2137--2146},
    year = {2018},
    organization = {PMLR},
}

@inproceedings{andriushchenko2020square,
    title = {Square attack: a query-efficient black-box adversarial attack via
             random search},
    author = {Andriushchenko, Maksym and Croce, Francesco and Flammarion,
              Nicolas and Hein, Matthias},
    booktitle = {European conference on computer vision},
    pages = {484--501},
    year = {2020},
    organization = {Springer},
}

@inproceedings{cai2022context,
    title = {Context-aware transfer attacks for object detection},
    author = {Cai, Zikui and Xie, Xinxin and Li, Shasha and Yin, Mingjun and
              Song, Chengyu and Krishnamurthy, Srikanth V and Roy-Chowdhury, Amit
              K and Asif, M Salman},
    booktitle = {Proceedings of the AAAI Conference on Artificial Intelligence},
    volume = {36},
    pages = {149--157},
    year = {2022},
}

@article{liu2018dpatch,
    title = {Dpatch: An adversarial patch attack on object detectors},
    author = {Liu, Xin and Yang, Huanrui and Liu, Ziwei and Song, Linghao and Li
              , Hai and Chen, Yiran},
    journal = {arXiv preprint arXiv:1806.02299},
    year = {2018},
}

@article{dai2023,
    author = {Dai, Zeyu and Liu, Shengcai and Li, Qing and Tang, Ke},
    title = {Saliency Attack: Towards Imperceptible Black-box Adversarial Attack
             },
    year = {2023},
    journal = {ACM Transactions on Intelligent Systems and Technology},
    issue_date = {June 2023},
    publisher = {Association for Computing Machinery},
    address = {New York, NY, USA},
    volume = {14},
    number = {3},
    issn = {2157-6904},
    month = apr,
    articleno = {45},
    numpages = {20},
}

@inproceedings{bhusalface,
    title = {FACE: Faithful Automatic Concept Extraction},
    author = {Bhusal, Dipkamal and Clifford, Michael and Rampazzi, Sara and
              Rastogi, Nidhi},
    year = {2025},
    booktitle = {The Thirty-ninth Annual Conference on Neural Information
                 Processing Systems},
}

@inproceedings{prfa,
    author = {Liang, Siyuan and Wu, Baoyuan and Fan, Yanbo and Wei, Xingxing and
              Cao, Xiaochun},
    booktitle = {2021 IEEE/CVF International Conference on Computer Vision
                 (ICCV)},
    title = {Parallel Rectangle Flip Attack: A Query-based Black-box Attack
             against Object Detection},
    year = {2021},
    volume = {},
    number = {},
    pages = {7677-7687},
}
\newpage
\section*{Supplementary Materials}

\subsection{Models}
We evaluate our methods using pre-trained models accessed via Python's Ultralytics package for \yolo and RT-DETR. These specific architectures represent three distinct object detection paradigms:
\begin{enumerate}
    \item \textbf{Single-stage real-time detector:} YOLOv11 (checkpoint: \texttt{yolo11n.pt})
    \item \textbf{Two-stage detector:} Faster R-CNN \texttt{FasterRCNN\_ResNet50\_FPN\_V2\_Weight}, standard COCO pre-trained)
    \item \textbf{Transformer-based detector:} RT-DETR (checkpoint: \texttt{rtdetr-l.pt})
\end{enumerate}

\subsection{Preliminary Spatial Analysis}
In this section, we analyze the spatial relationship between the model's bounding box and causal pixels (\msps) for the detection. To capture both linear and non-linear monotonic relationships, we compute three statistical measures: Spearman's $\rho$, Kendall's $\tau$, and Pearson's $r$. We include Pearson to capture linear relationships, while Spearman and Kendall Tau are utilized to account for potential non-linear monotonic relationships.  

\begin{table*}[h!]
\caption{\textbf{Correlation between Detection Confidence and Spatial Metrics.} We evaluate the DICE coefficient (\text{DC}) and the fraction of the explanation inside the bounding box (\text{FIN}) across three detector architectures. P-values are reported to demonstrate statistical significance.}
    \centering
    \resizebox{\textwidth}{!}{
    \begin{tabular}{ll cccccc}
    \toprule
    \textbf{Model} & \textbf{Metric} & \textbf{Spearman} ($\rho$) & \textbf{p-value} & \textbf{Kendall} ($\tau$) & \textbf{p-value} & \textbf{Pearson} ($r$) & \textbf{p-value} \\
    \midrule
    \multirow{2}{*}{\textbf{YOLO} $(n=5437)$} 
    & \text{FIN} & $0.30$ & $1.9 \times 10^{-112}$ & $0.21$ & $2.1 \times 10^{-108}$ & $0.45$ & $5.6 \times 10^{-265}$ \\
    & \text{DC} & $-0.12$ & $4.3 \times 10^{-19}$ & $-0.08$ & $1.4 \times 10^{-18}$ & $-0.13$ & $9.4 \times 10^{-22}$ \\
    \midrule
    \multirow{2}{*}{\textbf{RT-DETR} $(n=2717)$} 
    & \text{FIN} & $0.27$ & $9.5 \times 10^{-46}$ & $0.19$ & $1.4 \times 10^{-45}$ & $0.34$ & $1.2 \times 10^{-76}$ \\
    & \text{DC} & $-0.12$ & $6.9 \times 10^{-10}$ & $-0.085$ & $2.7 \times 10^{-11}$ & $-0.07$ & $2.8 \times 10^{-4}$ \\
    \midrule
    \multirow{2}{*}{\textbf{Faster R-CNN} $(n=590)$} 
    & \text{FIN} & $0.11$ & $0.01$ & $0.08$ & $0.01$ & $0.37$ & $1.3 \times 10^{-20}$ \\
    & \text{DC} & $-0.18$ & $1.5 \times 10^{-5}$ & $-0.12$ & $1.6 \times 10^{-5}$ & $-0.081$ & $0.05$ \\
    \bottomrule
    \end{tabular}
    }
    \label{tab:combined_correlation}
\end{table*}

\subsubsection*{Discussion of Spatial Correlation Findings}
The statistical results in \Cref{tab:combined_correlation} reveal two critical insights regarding how object detectors utilize causal features, that high confidence leads to an internal focus (positive \text{FIN} correlation) and the contextual dependency is persistent (Negative DC correlation). Across all three architectures, there is a statistically significant positive correlation (Pearson $r$ ranging from $0.34$ to $0.45$) between detection confidence and \text{FIN}. This indicates that as a detector becomes more confident in its classification, a larger proportion of the causal pixels are located strictly within the predicted bounding box.

Despite the increase in \text{FIN}, the overall overlap between the bounding box and the explanation mask exhibits a weak \textit{negative} correlation across all models (Spearman $\rho$ between $-0.12$ and $-0.18$). This is a highly counter-intuitive finding: it proves that even for the most confident predictions ($>0.9$), the causal explanation mask does not perfectly converge to the shape of the bounding box. 

These finding provide the theoretical foundations for our \blackcatt{} methodology.  As it confirms that object detectors rely on contextual cues located outside the bounding box. regardless of architecture. Consequently, an attacker does not need to occlude the main subject to successfully deceive the model; targeting the highly responsible, contextual pixels identified by the \msps is a viable and potent adversarial strategy.

\subsection*{Distribution of Causal Pixels Across Confidence Intervals}
\Cref{tab:combined_overlap_dist} show the distribution of \msps across confidence intervals. All three models exhibit a large proportion of images were at least some \msps fall outside the predicted bounding box: \yolo has $62.42\%$, RT-DETR has $63.29\%$ and Faster-R-CNN has $74.58\%$. Despite many detections occurring at high confidence $0.7-1.0$, a substantial fraction still includes \msps outside the bbox. This strongly suggests that the models rely on contextual cues beyond the object itself when predicting labels. The vast majority of detections rely on pixels outside the bounding box (Partial Overlap), even at the highest confidence levels.

\begin{table*}[!ht]
\caption{\textbf{Distribution of \msps{}-Bounding Box Overlap.} The percentage of instances exhibiting No Overlap, Partial Overlap, and Full Overlap across different confidence bins on the COCO test dataset.}
    \label{tab:combined_overlap_dist}
    \centering
    \resizebox{\textwidth}{!}{
    \begin{tabular}{ll ccc}
    \toprule
    \textbf{Model} & \textbf{Conf Bin} & \textbf{No Overlap (\%)} & \textbf{Partial Overlap (\%)} & \textbf{Full Overlap (\%)} \\
    \midrule
    \multirow{5}{*}{\textbf{YOLO} $(n=5437)$} 
    & $0.25 - 0.40$ & $0.51$ & $\textbf{8.48}$ & $1.38$ \\
    & $0.40 - 0.55$ & $0.15$ & $\textbf{5.35}$ & $2.02$ \\
    & $0.55 - 0.70$ & $0.07$ & $\textbf{7.08}$ & $3.02$ \\
    & $0.70 - 0.85$ & $0.04$ & $\textbf{13.70}$ & $9.23$ \\
    & $0.85 - 1.00$ & $0.09$ & $\textbf{26.95}$ & $21.92$ \\
    \midrule
    \multirow{5}{*}{\textbf{RT-DETR} $(n=2717)$} 
    & $0.25 - 0.40$ & $0.55$ & $\textbf{1.10}$ & $0.26$ \\
    & $0.40 - 0.55$ & $0.18$ & $\textbf{0.59}$ & $0.11$ \\
    & $0.55 - 0.70$ & $0.04$ & $\textbf{0.74}$ & $0.29$ \\
    & $0.70 - 0.85$ & $0.40$ & $\textbf{1.58}$ & $0.70$ \\
    & $0.85 - 1.00$ & $5.52$ & $\textbf{52.59}$ & $35.33$ \\
    \midrule
    \multirow{5}{*}{\textbf{Faster R-CNN} $(n=530)$} 
    & $0.25 - 0.40$ & $\textbf{0.19}$ & $0.00$ & $0.00$ \\
    & $0.40 - 0.55$ & $0.00$ & $\textbf{0.19}$ & $\textbf{0.19}$ \\
    & $0.55 - 0.70$ & $0.00$ & $0.00$ & $0.00$ \\
    & $0.70 - 0.85$ & $0.00$ & $\textbf{0.19}$ & $0.00$ \\
    & $0.85 - 1.00$ & $0.19$ & $\textbf{73.58}$ & $31.13$ \\
    \bottomrule
    \end{tabular}
    }
\end{table*}

\subsection*{Spatial Metric Across Confidence Intervals}
\Cref{tab:combined_fin_dc} show exact means of the spatial metrics, FIN and DC introduced in the main paper, at different confidence ranges. Note that Faster-R-CNN has a skewed confidence distribution as it tends to be overconfident. Generally, the mean FIN has increased with confidence and DC has stayed same or decreased slightly. A naive assumption in object detection is that high-confidence predictions are driven entirely by the object's localized features. This reliance on external context does not diminish to none as the model becomes more certain.
\begin{table*}[!ht]
\caption{\textbf{Distribution of Spatial metrics} The mean confidence, FIN and DC across different confidence bins on the COCO test dataset.}
    \label{tab:combined_fin_dc}
   \centering
    \resizebox{\textwidth}{!}{
    \begin{tabular}{ll cccc}
    \toprule
    \textbf{Model} & \textbf{Conf Bin} & \textbf{N} & \textbf{Mean Conf} & \textbf{Mean FIN}  & \textbf{Mean DC}  \\
    \midrule
    \multirow{8}{*}{\textbf{YOLO} $(n=5437)$} 
    &0.20-0.29 & 194 & $0.27$ & $0.53 \pm 0.38$ & $0.32 \pm 0.22$ \\
    &0.29-0.38 & 313 & $0.33$ & $0.63 \pm 0.34$ & $0.37 \pm 0.23$ \\
    &0.38-0.47 & 236 & $0.43$ & $0.73 \pm 0.31$ & $0.34 \pm 0.22$ \\
    &0.47-0.56 & 245 & $0.51$ &$0.79 \pm 0.28$ & $0.34 \pm 0.21$ \\
    &0.56-0.64 & 293 & $0.60$ & $0.79 \pm 0.27$ & $0.35 \pm 0.21$ \\
    &0.64-0.73 & 427 & $0.69$ & $0.82 \pm 0.25$ & $0.36 \pm 0.22$ \\
    &0.73-0.82 & 664 & $0.78$ & $0.89 \pm 0.18$ & $0.32 \pm 0.21$ \\
    &0.82-0.91 & 1743 & $0.87$ & $0.92 \pm 0.15$ & $0.27 \pm 0.21$ \\
    &0.91-1.00 & 1322 & $0.93$ &$0.94 \pm 0.11$ & $0.28 \pm 0.20$ \\
    \midrule
    \multirow{5}{*}{\textbf{RT-DETR} $(n=2717)$} 
    &0.20-0.29 & 22 & $0.26$ & $0.22 \pm 0.32$ & $0.097 \pm 0.15$ \\
    &0.29-0.38 & 27 & $0.33$ & $0.34 \pm 0.39$ & $0.16 \pm 0.19$ \\
    &0.38-0.47 & 16 & $0.44$ & $0.34 \pm 0.35$ & $0.24 \pm 0.27$ \\
    &0.47-0.56 & 13 & $0.52$  & $0.55 \pm 0.42$ & $0.23 \pm 0.17$ \\
    &0.56-0.64 & 18 & $0.59$ & $0.64 \pm 0.39$ & $0.2 \pm 0.17$ \\
    &0.64-0.73 & 13 & $0.69$ & $0.48 \pm 0.43$ & $0.16 \pm 0.21$ \\
    &0.73-0.82 & 31 & $0.78$ & $0.73 \pm 0.32$ & $0.22 \pm 0.2$ \\
    &0.82-0.91 & 308 & $0.88$  & $0.65 \pm 0.41$ & $0.18 \pm 0.19$ \\
    &0.91-1.00 & 2269 & $0.95$  & $0.85 \pm 0.26$ & $0.15 \pm 0.14$ \\
    \midrule
    \multirow{5}{*}{\textbf{Faster R-CNN} $(n=530)$}
    &0.20-0.29 & 2 & $0.25$  & $0.35 \pm 0.5$ & $0.3 \pm 0.43$ \\
    &0.29-0.38 & 0 & $--$ & $--$  & $--$  \\
    &0.38-0.47 & 0 & $--$ & $--$ & $--$  \\
    &0.47-0.56 & 2 & $0.5$ & $0.98 \pm 0.024$ & $0.21 \pm 0.22$ \\
    &0.56-0.64 & 0 & $--$ & $--$  & $--$  \\
    &0.64-0.73 & 1 & $0.71$  & $0.6$  & $0.73$ \\
    &0.73-0.82 & 0 & $--$  & $--$ & $--$ \\
    &0.82-0.91 & 2 & $0.88$  & $0.58\pm0.25$ & $0.26 \pm 0.19$ \\
    &0.91-1.00 & 554 & $1$ & $0.88\pm0.16$ & $0.21 \pm 0.19$ \\
    \bottomrule
    \end{tabular}
    }
\end{table*}

\subsection*{Single Inference Attacks on MSPS}

\begin{figure*}[t!]
    \centering
    \begin{subfigure}[b]{0.25\textwidth}
        \includegraphics[scale=0.16]{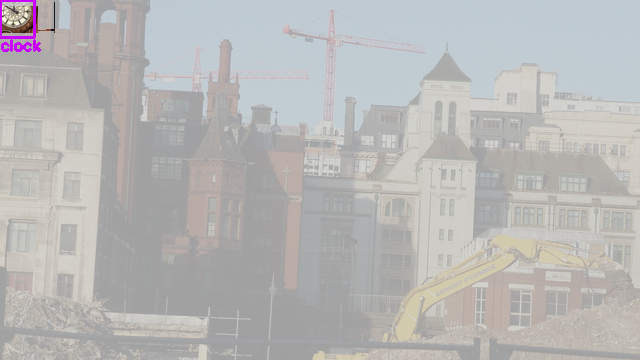}
        \caption{Clock with confidence $0.49$}
        \label{fig:clock:main}
    \end{subfigure}
    \hfill
    \begin{subfigure}[b]{0.13\textwidth}
        \centering
        \includegraphics[viewport=0 280 50 360,clip]{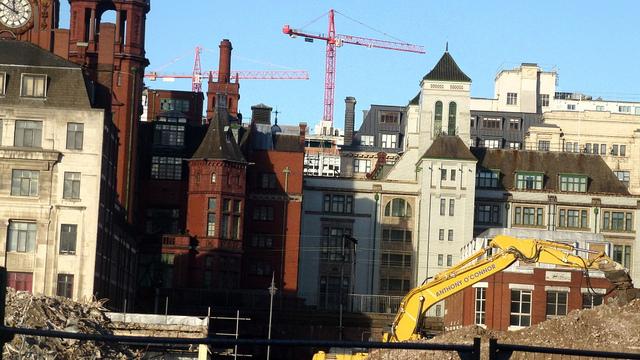}
        \caption{Blur}
    \end{subfigure}
    \hfill
    \begin{subfigure}{0.13\textwidth}
        \centering
        \includegraphics[viewport=0 280 50 360,clip]{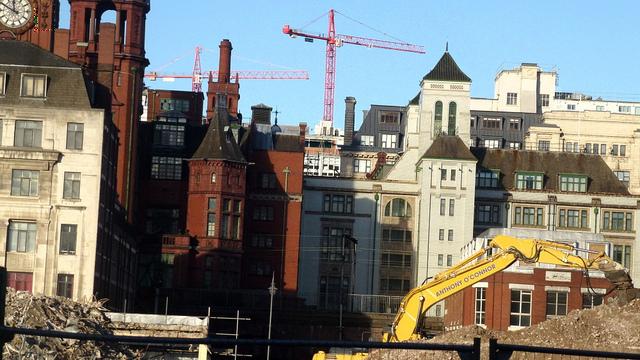}
        \caption{Noise}
    \end{subfigure}
    \hfill
    \begin{subfigure}{0.13\textwidth}
        \centering
        \includegraphics[viewport=0 280 50 360,clip]{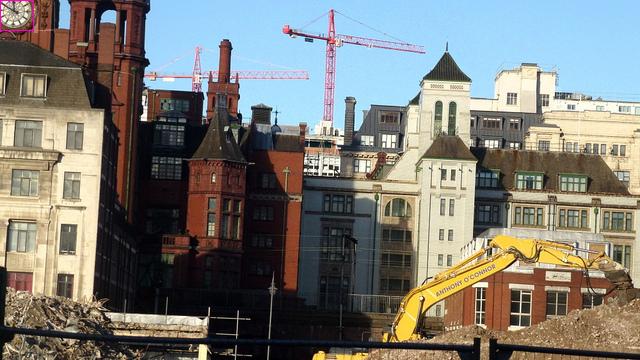}
        \caption{Colour}
    \end{subfigure}
    \hfill
    \begin{subfigure}{0.13\textwidth}
        \centering
        \includegraphics[viewport=0 280 50 360,clip]{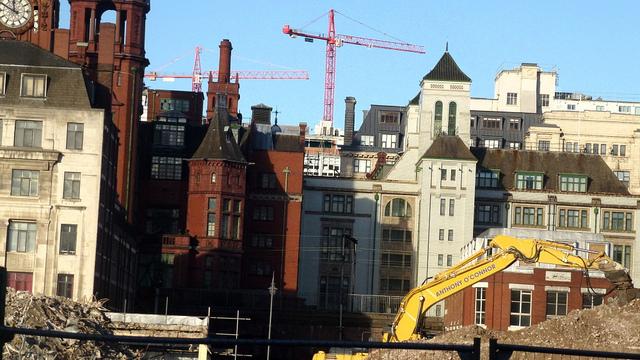}
        \caption{Shift}
    \end{subfigure}
\caption{Mutations on the pixels that are a part of the \msps but not in the bounding box. (zoomed in view)}
\label{fig:clock}
\end{figure*}

\begin{figure*}[t!]
    \centering 
    \begin{subfigure}[t]{\textwidth}
        \centering
        \includegraphics[width=0.33\linewidth]{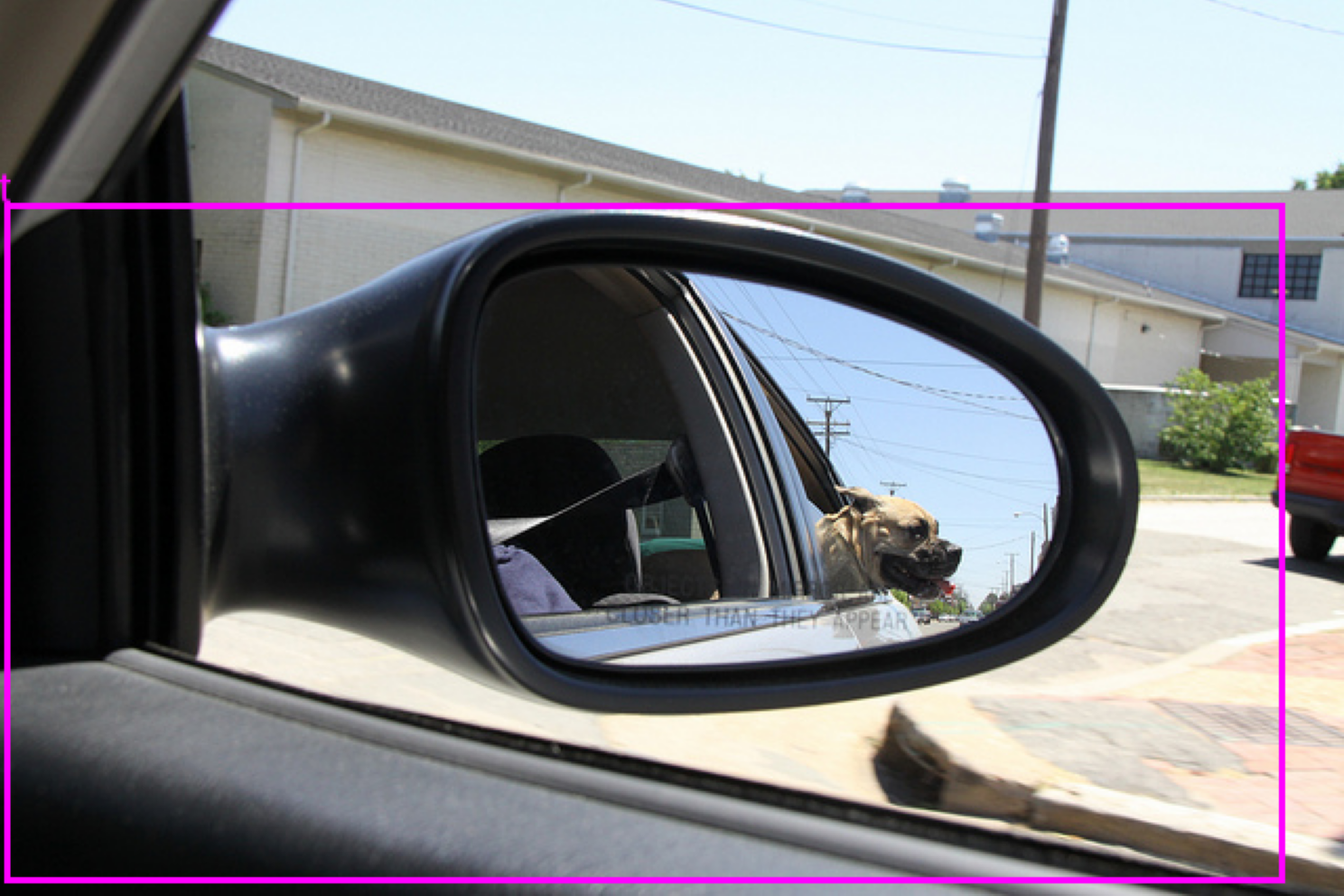}
        \caption{Car with $0.54$ confidence. }
        \label{fig:car:main}
    \end{subfigure}
    \begin{subfigure}[t]{\textwidth}
    \centering
     {\large\textbf{Inside}}\par\medskip
     \centering
    \begin{tabular}{@{}cc@{}}
      \subcaptionbox{Blur - No Detection\label{fig:car:inside:blur}}{%
        \includegraphics[width=0.33\linewidth]{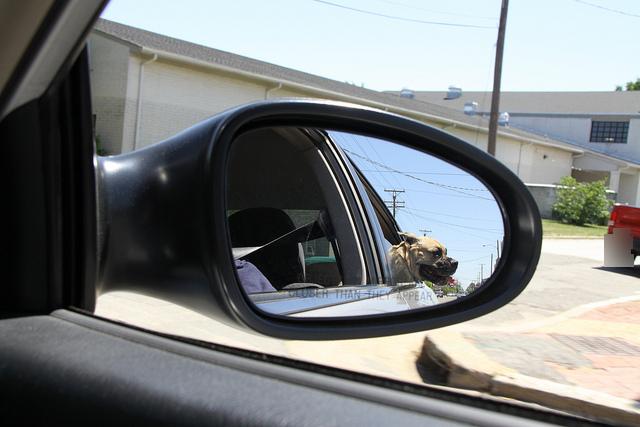}
      } &
      \subcaptionbox{Noise - No Detection\label{fig:car:inside:noise}}{%
        \includegraphics[width=0.33\linewidth]{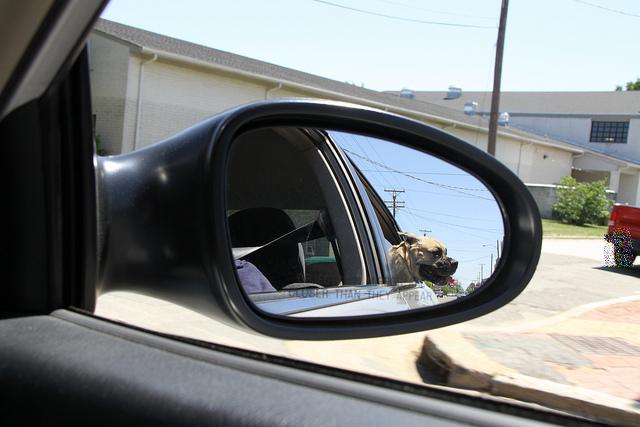}
      } \\
      \subcaptionbox{Pixel value - Detected Two Trucks with 0.345 and 0.312 confidence and car with 0.269 confidence \label{fig:car:inside:pixel}}{%
        \includegraphics[width=0.33\linewidth]{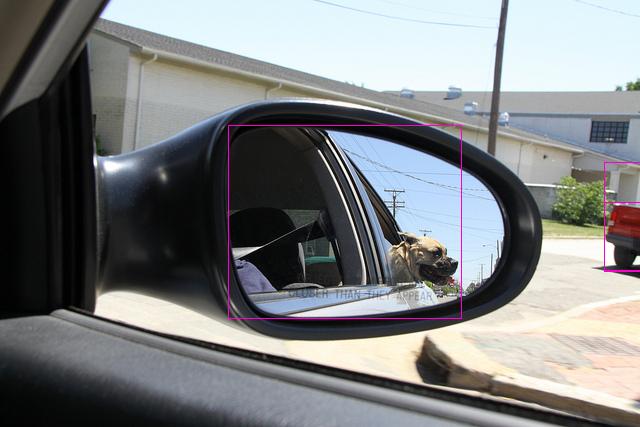}
      } &
      \subcaptionbox{Shift - Detected Truck with 0.417 confidence \label{fig:car:inside:shift}}{%
        \includegraphics[width=0.33\linewidth]{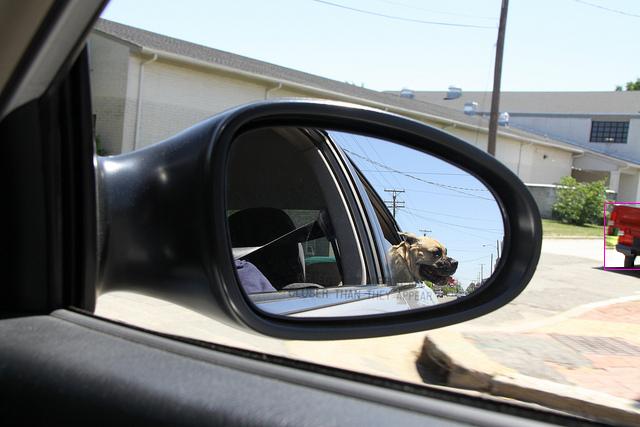}
      } \\
    \end{tabular}
    \centering \\
    {\large\textbf{Outside}}\par\medskip
    \begin{tabular}{@{}cc@{}}
      \subcaptionbox{Blur - Detected Bench with 0.286 confidence\label{fig:car:outside:blur}}{%
        \includegraphics[width=0.33\linewidth]{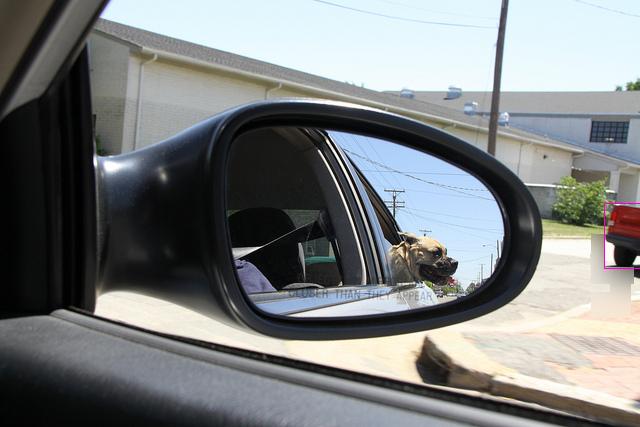}
      } &
      \subcaptionbox{Noise - No Detection\label{fig:car:outside:noise}}{%
        \includegraphics[width=0.33\linewidth]{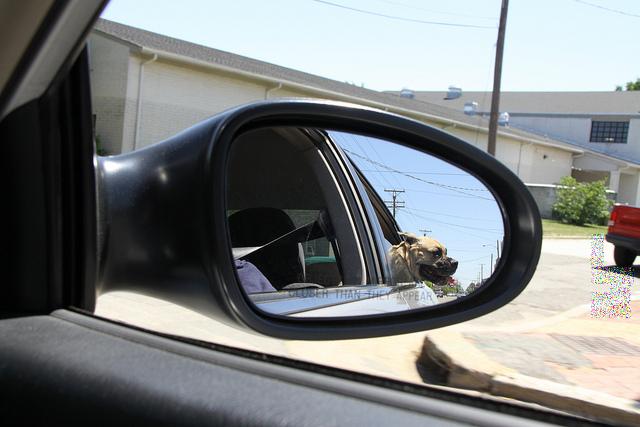}
      } \\
      \subcaptionbox{Pixel value - Detected Truck with 0.383 confidence and car with 0.264\label{fig:car:outside:pixel}}{%
        \includegraphics[width=0.33\linewidth]{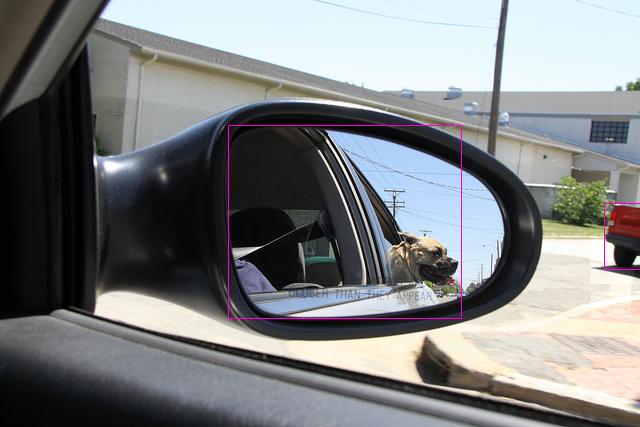}
      } &
      \subcaptionbox{Shift - Bench with 0.398 confidence\label{fig:car:outside:shift}}{%
        \includegraphics[width=0.33\linewidth]{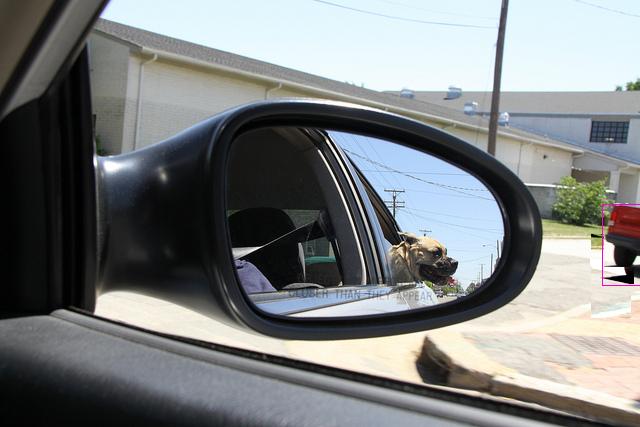}
      } \\
    \end{tabular}
    \end{subfigure}
    \caption{Single step attack on a \emph{car} using \yolo. Shows attacking the \msps that is \emph{inside} the bounding box and \emph{outside} the bounding box.}
    \label{fig:car}
\end{figure*}

\Cref{fig:car} shows a single-step attack using the \msps, where we add noise, blur, change pixel values, or shift pixels either on the part of the \msps inside the bbox or outside the bbox. All attacks were successful: we were able to add new detections (e.g. \Cref{fig:car:inside:pixel,fig:car:outside:pixel}), change detection labels (e.g. \Cref{fig:car:inside:shift}), and make detections disappear (e.g. \Cref{fig:car:outside:noise}). We also show an instance of a single-step attack on pixels outside the bbox in \Cref{fig:clock}, where all attacks are imperceptible and succeed (remove detection) except the pixel-value attack, which only lowers the confidence.

We show the full results of the single step attacks in \Cref{tab:yolo_full_initial_attack,tab:fasterrcnn_full_initial_attack,tab:rtdetr_full_initial_attack}. 

\begin{table}[t!]
\centering
\begin{tabularx}{\linewidth}{llXXX}
\toprule
&&\multicolumn{3}{c}{\textbf{Average Area}} \\ 
\cmidrule(lr){3-5}
method & attack & No Prediction  & Prediction Changed & Added New Prediction\\
\midrule
inside & blur & $5082.74$ & $2962.03$ & $1477.97$ \\
inside & noise & $3771.47$  & $2594.75$ & $1902.11$ \\
inside & shift & $1757.89$ & $1140.0$ & $2078.14$ \\
inside & pixel value & $1391.52$ & $412.05$ & $1983.83$ \\
outside & blur & $1150.03$ & $606.75$ & $365.87$ \\
outside & noise & $990.12$ & $620.4$ & $\textbf{\textit{434.08}}$ \\
outside & shift & $883.46$ & $379.11$ & $526.54$ \\
outside & pixel value & $\textbf{\textit{768.47}}$ & $\textbf{\textit{232.76}}$ & $499.96$ \\
\cmidrule(lr){3-5}
&&\multicolumn{3}{c}{\textbf{Number of Instances}} \\
\midrule
inside & blur & \textbf{\textit{1241}} & \textbf{\textit{643}} & 689 \\
inside & noise & 865 &  532 & 740 \\
inside & shift & 543 &  261 & \textbf{\textit{744}} \\
inside & pixel value & 297 & 109 & 683 \\
outside & blur & 337 &  172 & 658 \\
outside & noise & 358 &  177 & 679 \\
outside & shift & 303 &  141 & 686 \\
outside & pixel value & 254 &  99 & 685 \\
\bottomrule
\end{tabularx}
\caption{Dataset: \textbf{COCO}, Model: \textbf{YOLO} average area. The numbers in \textbf{\underline{bold and italics}} indicate the lowest average area change or highest number of images to get the given outcome (No Prediction (bbox removed), Prediction Changed and Adding a New Prediction). }
\label{tab:yolo_full_initial_attack}
\end{table}


\begin{table}[htb!]
\centering
\begin{tabularx}{\linewidth}{llXXX}
\toprule
&&\multicolumn{3}{c}{\textbf{Average Area}} \\ \\ 
\cmidrule(lr){3-5}
method & attack & No Prediction  & Prediction Changed & Added New Prediction\\
\midrule
inside & blur & $327.88$ & $434.61$ & $3653.85$ \\
inside & noise & $257.11$ & $278.22$ & $4046.76$ \\
inside & shift & $12.51$ & $152.94$ & $3454.94$ \\
inside & pixel value & $53.5$ &  $\textbf{\textit{112.63}}$ & $2420.93$ \\
outside & blur & $56.91$ & $200.19$ & $567.87$ \\
outside & noise & $\textbf{\textit{14.68}}$ &  $294.68$ & $552.12$ \\
outside & shift & $20.02$ &  $126.66$ & $585.8$ \\
outside & pixel value & $60.74$ & $192.23$ & $\textbf{\textit{438.26}}$ \\
\midrule
&&\multicolumn{3}{c}{\textbf{Number of Instances}} \\
\midrule
inside & blur & \textbf{\textit{17}} & \textbf{\textit{29}} & 306 \\
inside & noise & 8 & 20 & \textbf{\textit{328}} \\
inside & shift & 9  & 13 & 284 \\
inside & pixel value & 5 &  10 & 207 \\
outside & blur & 8  & 8 & 210 \\
outside & noise & 5  & 16 & 244 \\
outside & shift & 5  & 10 & 207 \\
outside & pixel value & 7 &  8 & 200 \\
\bottomrule
\end{tabularx}
\caption{\textbf{COCO} with \textbf{FASTER-RCNN} average area. The numbers in \textbf{\underline{bold and italics}} indicate the lowest average area change or highest number of images to get the given outcome (No Prediction (bbox removed), Prediction Changed and Adding a New Prediction).}
\label{tab:fasterrcnn_full_initial_attack}
\end{table}

\begin{table}[htb!]
\centering
\begin{tabularx}{\linewidth}{llXXX}
\toprule
&&\multicolumn{3}{c}{\textbf{Average Area}} \\
\cmidrule(lr){3-5}
method & attack & No Prediction  & Prediction Changed & Added New Prediction\\
\midrule
inside & blur & $3.93$ &  $438.54$ & $1140.85$ \\
inside & noise & $5.44$ &  $286.04$ & $1039.71$ \\
inside & shift & $5.2$ &  $165.8$ & $782.51$ \\
inside & pixel value & $\textbf{\textit{2.7}}$  & $128.27$ & $560.25$ \\
outside & blur & $7.2$ &  $241.19$ & $\textbf{\textit{157.23}}$ \\
outside & noise & $18.7$ &  $199.29$ & $281.29$ \\
outside & shift & $13.81$ &  $134.05$ & $277.9$ \\
outside & pixel value & $28.52$ & $\textbf{\textit{90.82}}$ & $174.35$ \\
\midrule
&&\multicolumn{3}{c}{\textbf{Number of Instances}} \\
\midrule
inside & blur & 17 &  \textbf{\textit{137}} & \textbf{\textit{428}} \\
inside & noise & 16 &  90 & 400 \\
inside & shift  & \textbf{\textit{18}} &  65 & 365 \\
inside & pixel value & 8 &  33 & 261 \\
outside & blur & 9 &  48 & 253 \\
outside & noise & 9 &  46 & 318 \\
outside & shift & 9 &  40 & 280 \\
outside & pixel value & 11 &  24 & 262 \\
\bottomrule
\end{tabularx}
\caption{\textbf{COCO} with \textbf{RT-DETR} average area. The numbers in \textbf{\underline{bold and italics}} indicate the lowest average area change or highest number of images to get the given outcome (No Prediction (bbox removed), Prediction Changed and Adding a New Prediction).}
\label{tab:rtdetr_full_initial_attack}
\end{table}

\subsection{Experiment}

\begin{figure*}
    \begin{subfigure}{\textwidth}
    \centering
            \begin{subfigure}[t]{0.32\textwidth}
        \centering
        \pgfplotsset{width=4.55cm, height=3.5cm}
        \centering
        \begin{tikzpicture}
            \begin{axis}[
            ylabel={Success Rate \%},
            xlabel={$L_0$ threshold},
            grid=major,
            xticklabels={},
            extra x ticks={1, 2, 3, 4},
            xmin=1,
            xmax=4,
            extra x tick labels={0.01, 0.05, 0.1, 0.2}]
                \addplot[red,thick] coordinates {
                (1,0.0)
                (2,0.0)
                (3,(0.0)
                (4, 0.0)
                };
                \addplot[blue,thick] coordinates {
                (1, 5.02)
                (2, 10.31)
                (3, 12.16)
                (4, 13.12)
                };
                \addplot[green,thick] coordinates {
                (1, 11.84)
                (2, 23.94)
                (3, 34.73)
                (4, 39.47)
                };
                \addplot[orange,thick] coordinates {
                (1, 16.94)
                (2, 34.8)
                (3, 42.69)
                (4, 52.61)
                };
                \addplot[purple,thick] coordinates {
                (1, 14.53)
                (2, 29.98)
                (3, 37.57)
                (4, 46.08)
                };
            \end{axis}
        \end{tikzpicture}
        \caption{\yolo}
    \end{subfigure}
    \hfill
    \centering
    \begin{subfigure}[t]{0.32\textwidth}
    \pgfplotsset{width=4.55cm, height=3.5cm}
    \centering
    \begin{tikzpicture}
        \begin{axis}[
        xlabel={$L_0$ threshold},
        grid=major,
        xticklabels={},
        extra x ticks={1, 2, 3, 4},
        xmin=1,
        xmax=4,
        extra x tick labels={0.01, 0.05, 0.1, 0.2}]
            \addplot[red,thick] coordinates {
            (1,0.0)
            (2,0.0)
            (3,(0.0)
            (4, 0.0)
            };
            \addplot[blue,thick] coordinates {
            (1, 0.93)
            (2, 1.41)
            (3, 1.52)
            (4, 1.59)
            };
            \addplot[green,thick] coordinates {
            (1, 1.19)
            (2, 1.74)
            (3, 2.19)
            (4, 2.3)
            };
            \addplot[orange,thick] coordinates {
            (1, 1.93)
            (2, 4.07)
            (3, 5.93)
            (4, 7.56)
            };
            \addplot[purple,thick] coordinates {
            (1, 1.89)
            (2, 2.7)
            (3, 3.81)
            (4, 4.96)
            };
        \end{axis}
    \end{tikzpicture}
        \caption{RT-DETR}
    \end{subfigure}
    \hfill
    \centering
    \begin{subfigure}[t]{0.3\textwidth}
     \pgfplotsset{width=4.55cm, height=3.5cm}
    \centering
    \begin{tikzpicture}
        \begin{axis}[
        xlabel={$L_0$ threshold},
        grid=major,
        xticklabels={},
        extra x ticks={1, 2, 3, 4},
        xmin=1,
        xmax=4,
        extra x tick labels={0.01, 0.05, 0.1, 0.2}]
            \addplot[red,thick] coordinates {
            (1,0.0)
            (2,0.0)
            (3,(0.0)
            (4, 0.0)
            };
            \addplot[blue,thick] coordinates {
            (1, 1.64)
            (2, 3.82)
            (3, 4.36)
            (4, 4.36)
            };
            \addplot[green,thick] coordinates {
            (1, 2.73)
            (2, 4.36)
            (3, 5.27)
            (4, 5.45)
            };
            \addplot[orange,thick] coordinates {
            (1, 0.55)
            (2, 4.18)
            (3, 7.45)
            (4, 11.45)
            };
            \addplot[purple,thick] coordinates {
            (1, 4.36)
            (2, 9.09)
            (3, 11.64)
            (4, 16.0)
            };
        \end{axis}
    \end{tikzpicture}
    \hfill
        \caption{FASTER-RCNN}
    \end{subfigure}
            \begin{subfigure}[t]{0.32\textwidth}
        \centering
        \pgfplotsset{width=4.55cm, height=3.5cm}
        \centering
        \begin{tikzpicture}
            \begin{axis}[
            ylabel={Success Rate \%},
            xlabel={$L_1$ threshold},
            grid=major,
            xticklabels={},
            extra x ticks={1, 2, 3, 4},
            xmin=1,
            xmax=4,
            extra x tick labels={2/255, 4/255, 8/255, 10/255}]
                \addplot[red,thick] coordinates {
                (1,0.0)
                (2,0.0)
                (3,0.16)
                (4, 8.08)
                };
                \addplot[blue,thick] coordinates {
                (1, 11.67)
                (2, 12.9)
                (3, 13.14)
                (4, 13.14)
                };
                \addplot[green,thick] coordinates {
                (1, 31.37)
                (2, 38.06)
                (3, 40.67)
                (4, 40.92)
                };
                \addplot[orange,thick] coordinates {
                (1, 32.65)
                (2, 38.51)
                (3, 47.45)
                (4, 51.1)
                };
                \addplot[purple,thick] coordinates {
                (1, 30.1)
                (2, 36.31)
                (3, 43.84)
                (4, 46.24)
                };
            \end{axis}
        \end{tikzpicture}
        \caption{ \yolo}
    \end{subfigure}
    \hfill
    \centering
    \begin{subfigure}[t]{0.32\textwidth}
    \pgfplotsset{width=4.55cm, height=3.5cm}
    \centering
    \begin{tikzpicture}
        \begin{axis}[
        xlabel={$L_1$ threshold},
        grid=major,
        xticklabels={},
        extra x ticks={1, 2, 3, 4},
        xmin=1,
        xmax=4,
        extra x tick labels={2/255, 4/255, 8/255, 10/255}]
            \addplot[red,thick] coordinates {
            (1,0.0)
            (2,0.0)
            (3,0.0)
            (4, 0.52)
            };
            \addplot[blue,thick] coordinates {
            (1, 1.52)
            (2, 1.59)
            (3, 1.59)
            (4, 1.59)
            };
            \addplot[green,thick] coordinates {
            (1, 2.15)
            (2, 2.37)
            (3, 2.44)
            (4, 2.44)
            };
            \addplot[orange,thick] coordinates {
            (1, 3.67)
            (2, 5.26)
            (3, 6.89)
            (4, 7.37)
            };
            \addplot[purple,thick] coordinates {
            (1, 2.67)
            (2, 3.59)
            (3, 4.67)
            (4, 4.96)
            };
        \end{axis}
    \end{tikzpicture}
        \caption{ RT-DETR}
    \end{subfigure}
    \hfill
    \centering
    \begin{subfigure}[t]{0.3\textwidth}
     \pgfplotsset{width=4.55cm, height=3.5cm}
    \centering
    \begin{tikzpicture}
        \begin{axis}[
        xlabel={$L_1$ threshold},
        grid=major,
        xticklabels={},
        extra x ticks={1, 2, 3, 4},
        xmin=1,
        xmax=4,
        extra x tick labels={2/255, 4/255, 8/255, 10/255}]
            \addplot[red,thick] coordinates {
            (1,0.0)
            (2,0.0)
            (3,0.0)
            (4, 0.73)
            };
            \addplot[blue,thick] coordinates {
            (1, 4.36)
            (2, 4.36)
            (3, 4.36)
            (4, 4.36)
            };
            \addplot[green,thick] coordinates {
            (1, 4.73)
            (2, 5.45)
            (3, 5.45)
            (4, 5.45)
            };
            \addplot[orange,thick] coordinates {
            (1, 3.64)
            (2, 6.36)
            (3, 9.45)
            (4, 10.73)
            };
            \addplot[purple,thick] coordinates {
            (1, 8.91)
            (2, 11.82)
            (3, 15.64)
            (4, 16.18)
            };
        \end{axis}
    \end{tikzpicture}
        \caption{ FASTER-RCNN}
    \end{subfigure}
            \begin{subfigure}[t]{0.32\textwidth}
        \centering
        \pgfplotsset{width=4.55cm, height=3.5cm}
        \centering
        \begin{tikzpicture}
            \begin{axis}[
            ylabel={Success Rate \%},
            xlabel={$L_2$ threshold},
            grid=major,
            xticklabels={},
            extra x ticks={1, 2, 3, 4},
            xmin=1,
            xmax=4,
            extra x tick labels={2/255, 4/255, 8/255, 10/255}]
                \addplot[red,thick] coordinates {
                (1, 0.0)
                (2, 0.0)
                (3, 0.0)
                (4, 0.04081632653061225)
                };
                \addplot[blue,thick] coordinates {
                (1, 4.612244897959183)
                (2, 8.26530612244898)
                (3, 10.877551020408163)
                (4, 11.795918367346939)
                };
                \addplot[green,thick] coordinates {
                (1, 10.346938775510205)
                (2, 18.26530612244898)
                (3, 28.448979591836736)
                (4, 32.69387755102041)
                };
                \addplot[orange,thick] coordinates {
                (1, 7.081632653061225)
                (2, 12.775510204081634)
                (3, 24.142857142857142)
                (4, 27.836734693877553)
                };
                \addplot[purple,thick] coordinates {
                (1, 6.877551020408164)
                (2, 13.061224489795919)
                (3, 22.75510204081633)
                (4, 25.06122448979592)
                };
            \end{axis}
        \end{tikzpicture}
        \caption{ \yolo}
    \end{subfigure}
    \hfill
    \centering
    \begin{subfigure}[t]{0.32\textwidth}
    \pgfplotsset{width=4.55cm, height=3.5cm}
    \centering
    \begin{tikzpicture}
        \begin{axis}[
        xlabel={$L_2$ threshold},
        grid=major,
        xticklabels={},
        extra x ticks={1, 2, 3, 4},
        xmin=1,
        xmax=4,
        extra x tick labels={2/255, 4/255, 8/255, 10/255}]
            \addplot[red,thick] coordinates {
            (1,0.0)
            (2,0.0)
            (3,0.0)
            (4, 0.0)
            };
            \addplot[blue,thick] coordinates {
            (1, 0.9259259259259258)
            (2, 1.2962962962962963)
            (3, 1.4444444444444444)
            (4, 1.5185185185185186)
            };
            \addplot[green,thick] coordinates {
            (1, 1.1851851851851851)
            (2, 1.6296296296296295)
            (3, 2.037037037037037)
            (4, 2.185185185185185)
            };
            \addplot[orange,thick] coordinates {
            (1, 1.2222222222222223)
            (2, 1.5925925925925926)
            (3, 2.2962962962962963)
            (4, 2.5925925925925926)
            };
            \addplot[purple,thick] coordinates {
            (1, 1.2962962962962963)
            (2, 1.7777777777777777)
            (3, 2.148148148148148)
            (4, 2.2962962962962963)
            };
        \end{axis}
    \end{tikzpicture}
        \caption{ RT-DETR}
    \end{subfigure}
    \hfill
    \centering
    \begin{subfigure}[t]{0.3\textwidth}
     \pgfplotsset{width=4.55cm, height=3.5cm}
    \centering
    \begin{tikzpicture}
        \begin{axis}[
        xlabel={$L_1$ threshold},
        grid=major,
        xticklabels={},
        extra x ticks={1, 2, 3, 4},
        xmin=1,
        xmax=4,
        extra x tick labels={2/255, 4/255, 8/255, 10/255}]
            \addplot[red,thick] coordinates {
            (1,0.0)
            (2,0.0)
            (3,0.0)
            (4, 0.0)
            };
            \addplot[blue,thick] coordinates {
            (1, 2.3636363636363638)
            (2, 3.6363636363636362)
            (3, 4.0)
            (4, 4.363636363636364)
            };
            \addplot[green,thick] coordinates {
            (1, 3.090909090909091)
            (2, 3.4545454545454546)
            (3, 4.545454545454546)
            (4, 4.909090909090909)
            };
            \addplot[orange,thick] coordinates {
            (1, 0.0)
            (2, 0.18181818181818182)
            (3, 1.090909090909091)
            (4, 1.2727272727272727)
            };
            \addplot[purple,thick] coordinates {
            (1, 4.0)
            (2, 4.181818181818182)
            (3, 5.090909090909091)
            (4, 6.0)
            };
        \end{axis}
    \end{tikzpicture}
        \caption{FASTER-RCNN}
    \end{subfigure}
            \begin{subfigure}[t]{0.32\textwidth}
        \centering
        \pgfplotsset{width=4.55cm, height=3.5cm}
        \centering
        \begin{tikzpicture}
            \begin{axis}[
            ylabel={Success Rate \%},
            xlabel={$LPIPS$ threshold},
            grid=major,
            xticklabels={},
            extra x ticks={1, 2, 3, 4},
            xmin=1,
            xmax=4,
            extra x tick labels={0.01 , 0.05 , 0.1 , 0.2}]
                \addplot[red,thick] coordinates {
                (1, 0.0)
                (2, 0.29)
                (3, 1.8)
                (4, 7.41)
                };
                \addplot[blue,thick] coordinates {
                (1, 5.67)
                (2, 9.96)
                (3, 11.82)
                (4, 12.92)
                };
                \addplot[green,thick] coordinates {
                (1, 16.02)
                (2, 28.39)
                (3, 35.63)
                (4, 39.98)
                };
                \addplot[orange,thick] coordinates {
                (1, 14.63)
                (2, 32.67)
                (3, 41.12)
                (4, 51.8)
                };
                \addplot[purple,thick] coordinates {
                (1, 15.27)
                (2, 31.18)
                (3, 38.88)
                (4, 46.71)
                };
            \end{axis}
        \end{tikzpicture}
        \caption{ \yolo}
    \end{subfigure}
    \hfill
    \centering
    \begin{subfigure}[t]{0.32\textwidth}
    \pgfplotsset{width=4.55cm, height=3.5cm}
    \centering
    \begin{tikzpicture}
        \begin{axis}[
        xlabel={$LPIPS$ threshold},
        grid=major,
        xticklabels={},
        extra x ticks={1, 2, 3, 4},
        xmin=1,
        xmax=4,
        extra x tick labels={0.01 , 0.05 , 0.1 , 0.2}]
            \addplot[red,thick] coordinates {
            (1,0.0)
            (2,0.0)
            (3,0.22)
            (4, 0.52)
            };
            \addplot[blue,thick] coordinates {
            (1, 1.22)
            (2, 1.44)
            (3, 1.48)
            (4, 1.59)
            };
            \addplot[green,thick] coordinates {
            (1, 1.48)
            (2, 2.07)
            (3, 2.19)
            (4, 2.44)
            };
            \addplot[orange,thick] coordinates {
            (1, 1.89)
            (2, 4.11)
            (3, 6.15)
            (4, 7.56)
            };
            \addplot[purple,thick] coordinates {
            (1, 1.89)
            (2, 2.96)
            (3, 3.96)
            (4, 5.07)
            };
        \end{axis}
    \end{tikzpicture}
        \caption{ RT-DETR}
    \end{subfigure}
    \hfill
    \centering
    \begin{subfigure}[t]{0.3\textwidth}
     \pgfplotsset{width=4.55cm, height=3.5cm}
    \centering
    \begin{tikzpicture}
        \begin{axis}[
        xlabel={$LPIPS$ threshold},
        grid=major,
        xticklabels={},
        extra x ticks={1, 2, 3, 4},
        xmin=1,
        xmax=4,
        extra x tick labels={0.01 , 0.05 , 0.1 , 0.2}]
            \addplot[red,thick] coordinates {
            (1,0.0)
            (2,0.0)
            (3,0.36)
            (4, 0.91)
            };
            \addplot[blue,thick] coordinates {
            (1, 3.09)
            (2, 3.45)
            (3, 4.36)
            (4, 4.36)
            };
            \addplot[green,thick] coordinates {
            (1, 3.27)
            (2, 4.36)
            (3, 5.09)
            (4, 5.45)
            };
            \addplot[orange,thick] coordinates {
            (1, 0.91)
            (2, 4.0)
            (3, 8.18)
            (4, 11.27)
            };
            \addplot[purple,thick] coordinates {
            (1, 4.36)
            (2, 8.91)
            (3, 13.64)
            (4, 16.73)
            };
        \end{axis}
    \end{tikzpicture}
        \caption{FASTER-RCNN}
    \end{subfigure}
            \begin{subfigure}[t]{0.32\textwidth}
        \centering
        \pgfplotsset{width=4.55cm, height=3.5cm}
        \centering
        \begin{tikzpicture}
            \begin{axis}[
            ylabel={Success Rate \%},
            xlabel={$SSIM$ threshold},
            grid=major,
            xticklabels={},
            extra x ticks={1, 2, 3, 4},
            xmin=1,
            xmax=4,
            extra x tick labels={0.99 , 0.98 , 0.95 , 0.9}]
                \addplot[red,thick] coordinates {
                (1, 0.0)
                (2, 0.0)
                (3, 0.0)
                (4, 0.06)
                };
                \addplot[blue,thick] coordinates {
                (1, 5.24)
                (2, 8.22)
                (3, 11.2)
                (4, 12.53)
                };
                \addplot[green,thick] coordinates {
                (1, 14.73)
                (2, 19.73)
                (3, 30.12)
                (4, 37.63)
                };
                \addplot[orange,thick] coordinates {
                (1, 22.71)
                (2, 30.0)
                (3, 38.84)
                (4, 48.51)
                };
                \addplot[purple,thick] coordinates {
                (1, 20.82)
                (2, 26.86)
                (3, 35.73)
                (4, 43.86)
                };
            \end{axis}
        \end{tikzpicture}
        \caption{ \yolo}
    \end{subfigure}
    \hfill
    \centering
    \begin{subfigure}[t]{0.32\textwidth}
    \pgfplotsset{width=4.55cm, height=3.5cm}
    \centering
    \begin{tikzpicture}
        \begin{axis}[
        xlabel={$SSIM$ threshold},
        grid=major,
        xticklabels={},
        extra x ticks={1, 2, 3, 4},
        xmin=1,
        xmax=4,
        extra x tick labels={0.01 , 0.05 , 0.1 , 0.2}]
            \addplot[red,thick] coordinates {
            (1,0.0)
            (2,0.0)
            (3,0.0)
            (4, 0.0)
            };
            \addplot[blue,thick] coordinates {
            (1, 1.19)
            (2, 1.37)
            (3, 1.41)
            (4, 1.59)
            };
            \addplot[green,thick] coordinates {
            (1, 1.33)
            (2, 1.59)
            (3, 2.07)
            (4, 2.3)
            };
            \addplot[orange,thick] coordinates {
            (1, 2.33)
            (2, 3.07)
            (3, 5.22)
            (4, 7.15)
            };
            \addplot[purple,thick] coordinates {
            (1, 2.11)
            (2, 2.37)
            (3, 3.48)
            (4, 4.81)
            };
        \end{axis}
    \end{tikzpicture}
        \caption{ RT-DETR}
    \end{subfigure}
    \hfill
    \centering
    \begin{subfigure}[t]{0.3\textwidth}
     \pgfplotsset{width=4.55cm, height=3.5cm}
    \centering
    \begin{tikzpicture}
        \begin{axis}[
        xlabel={$SSIM$ threshold},
        grid=major,
        xticklabels={},
        extra x ticks={1, 2, 3, 4},
        xmin=1,
        xmax=4,
        extra x tick labels={0.01 , 0.05 , 0.1 , 0.2}]
            \addplot[red,thick] coordinates {
            (1,0.0)
            (2,0.0)
            (3,0.0)
            (4, 0.0)
            };
            \addplot[blue,thick] coordinates {
            (1, 2.73)
            (2, 3.09)
            (3, 4.0)
            (4, 4.36)
            };
            \addplot[green,thick] coordinates {
            (1, 3.64)
            (2, 3.82)
            (3, 4.91)
            (4, 5.45)
            };
            \addplot[orange,thick] coordinates {
            (1, 1.27)
            (2, 2.55)
            (3, 6.36)
            (4, 10.18)
            };
            \addplot[purple,thick] coordinates {
            (1, 5.64)
            (2, 7.27)
            (3, 11.45)
            (4, 15.64)
            };
        \end{axis}
    \end{tikzpicture}
        \caption{FASTER-RCNN}
    \end{subfigure}
    \end{subfigure}
    \caption{Success rate of different approaches in \emph{removing a detection}, with different models on COCO dataset, for different thresholds with $L_0, L_1, L_2, LPIPS, SSIM$. The different techniques are
     \textcolor{red}{noise}, \textcolor{blue}{targeted noise}, \textcolor{green}{blended},  \textcolor{orange}{DRISE$_{MoG}$} and \textcolor{purple}{MoG}.}
    \label{fig:success_rate_no_pred}
\end{figure*}

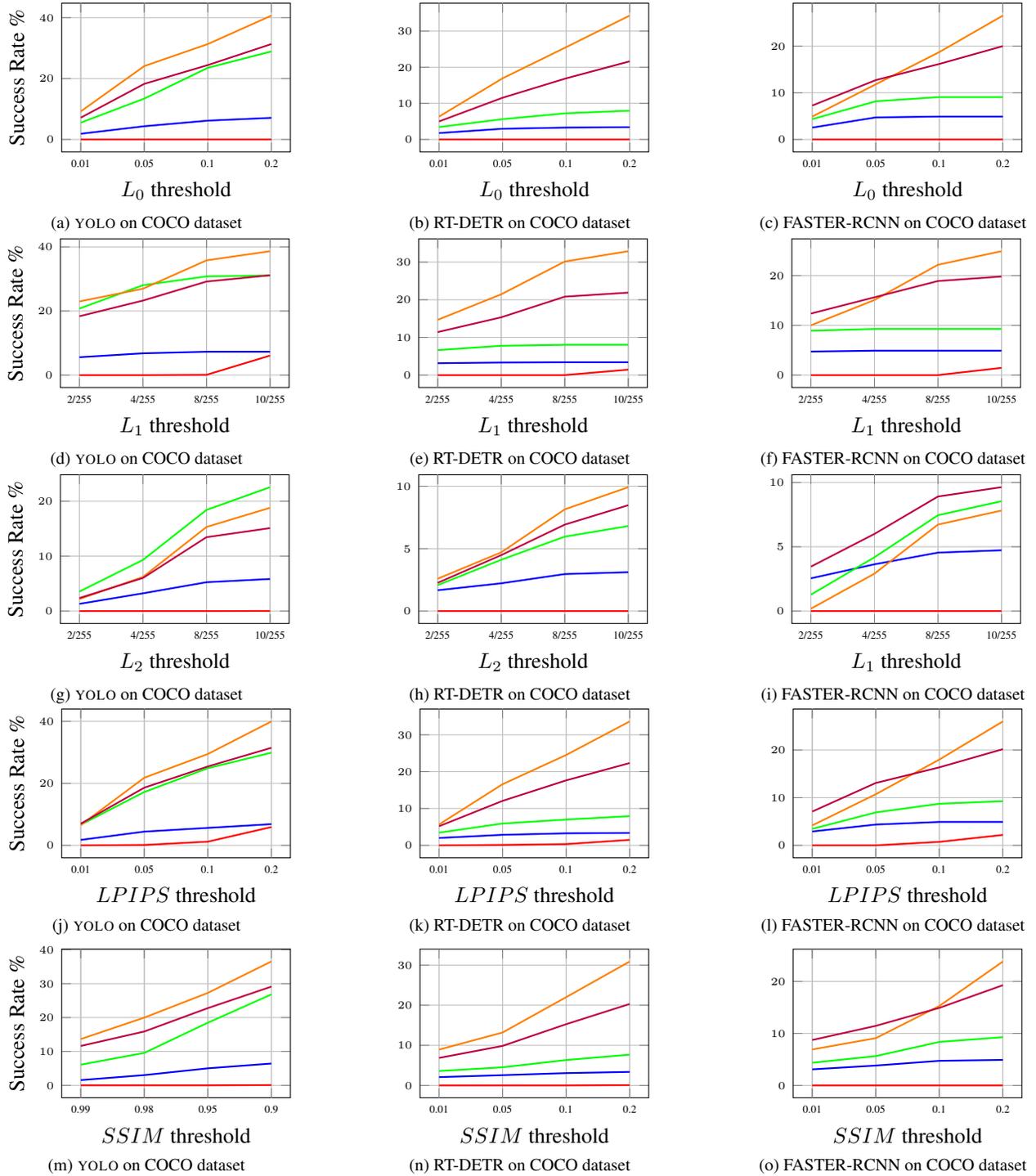
\begin{figure*}
    \begin{subfigure}{\textwidth}
        \centering
            \begin{subfigure}[t]{0.3\textwidth}
        \centering
        \pgfplotsset{width=4.55cm, height=3.5cm}
        \centering
        \begin{tikzpicture}
            \begin{axis}[
            ylabel={Success Rate \%},
            xlabel={$L_0$ threshold},
            grid=major,
            xticklabels={},
            extra x ticks={1, 2, 3, 4},
            extra x tick labels={0.01, 0.05, 0.1, 0.2}]
                \addplot[red,thick] coordinates {
                (1,0.0)
                (2,0.0)
                (3,(0.0)
                (4, 0.0)
                };
                \addplot[blue,thick] coordinates {
                (1, 1.8775510204081631)
                (2, 4.346938775510204)
                (3, 6.183673469387755)
                (4, 7.102040816326531)
                };
                \addplot[green,thick] coordinates {
                (1, 5.530612244897959)
                (2, 13.428571428571429)
                (3, 23.53061224489796)
                (4, 28.89795918367347)
                };
                \addplot[orange,thick] coordinates {
                (1, 9.224489795918366)
                (2, 24.102040816326532)
                (3, 31.326530612244895)
                (4, 40.6734693877551)
                };
                \addplot[purple,thick] coordinates {
                (1, 7.163265306122449)
                (2, 18.26530612244898)
                (3, 24.428571428571427)
                (4, 31.326530612244895)
                };
            \end{axis}
        \end{tikzpicture}
        \caption{ \yolo}
    \end{subfigure}
    \hfill
    \centering
    \begin{subfigure}[t]{0.3\textwidth}
    \pgfplotsset{width=4.55cm, height=3.5cm}
    \centering
    \begin{tikzpicture}
        \begin{axis}[
        xlabel={$L_0$ threshold},
        grid=major,
        xticklabels={},
        extra x ticks={1, 2, 3, 4},
        extra x tick labels={0.01, 0.05, 0.1, 0.2}]
            \addplot[red,thick] coordinates {
            (1,0.0)
            (2,0.0)
            (3,(0.0)
            (4, 0.0)
            };
            \addplot[blue,thick] coordinates {
            (1, 1.7777777777777777)
            (2, 2.9629629629629632)
            (3, 3.2962962962962963)
            (4, 3.4074074074074074)
            };
            \addplot[green,thick] coordinates {
            (1, 3.4444444444444446)
            (2, 5.62962962962963)
            (3, 7.2592592592592595)
            (4, 7.962962962962964)
            };
            \addplot[orange,thick] coordinates {
            (1, 6.333333333333334)
            (2, 16.925925925925924)
            (3, 25.555555555555554)
            (4, 34.2962962962963)
            };
            \addplot[purple,thick] coordinates {
            (1, 5.0)
            (2, 11.555555555555555)
            (3, 16.925925925925924)
            (4, 21.62962962962963)
            };
        \end{axis}
    \end{tikzpicture}
        \caption{ RT-DETR}
    \end{subfigure}
    \hfill
    \centering
    \begin{subfigure}[t]{0.3\textwidth}
     \pgfplotsset{width=4.55cm, height=3.5cm}
    \centering
    \begin{tikzpicture}
        \begin{axis}[
        xlabel={$L_0$ threshold},
        grid=major,
        xticklabels={},
        extra x ticks={1, 2, 3, 4},
        extra x tick labels={0.01, 0.05, 0.1, 0.2}]
            \addplot[red,thick] coordinates {
            (1,0.0)
            (2,0.0)
            (3,(0.0)
            (4, 0.0)
            };
            \addplot[blue,thick] coordinates {
            (1, 2.5454545454545454)
            (2, 4.7272727272727275)
            (3, 4.909090909090909)
            (4, 4.909090909090909)
            };
            \addplot[green,thick] coordinates {
            (1, 4.363636363636364)
            (2, 8.181818181818182)
            (3, 9.090909090909092)
            (4, 9.090909090909092)
            };
            \addplot[orange,thick] coordinates {
            (1, 4.909090909090909)
            (2, 11.818181818181818)
            (3, 18.72727272727273)
            (4, 26.545454545454543)
            };
            \addplot[purple,thick] coordinates {
            (1, 7.2727272727272725)
            (2, 12.727272727272727)
            (3, 16.18181818181818)
            (4, 20.0)
            };
        \end{axis}
    \end{tikzpicture}
        \caption{ FASTER-RCNN}
    \end{subfigure}
            \begin{subfigure}[t]{0.3\textwidth}
        \centering
        \pgfplotsset{width=4.55cm, height=3.5cm}
        \centering
        \begin{tikzpicture}
            \begin{axis}[
            ylabel={Success Rate \%},
            xlabel={$L_1$ threshold},
            grid=major,
            xticklabels={},
            extra x ticks={1, 2, 3, 4},
            extra x tick labels={2/255, 4/255, 8/255, 10/255}]
                \addplot[red,thick] coordinates {
                (1,0.0)
                (2,0.0)
                (3,0.0816326530612245)
                (4, 6.081632653061225)
                };
                \addplot[blue,thick] coordinates {
                (1, 5.571428571428571)
                (2, 6.795918367346938)
                (3, 7.285714285714286)
                (4, 7.285714285714286)
                };
                \addplot[green,thick] coordinates {
                (1, 20.755102040816325)
                (2, 28.04081632653061)
                (3, 30.795918367346935)
                (4, 31.081632653061224)
                };
                \addplot[orange,thick] coordinates {
                (1, 22.979591836734695)
                (2, 26.93877551020408)
                (3, 35.79591836734694)
                (4, 38.6530612244898)
                };
                \addplot[purple,thick] coordinates {
                (1, 18.346938775510203)
                (2, 23.244897959183675)
                (3, 29.183673469387756)
                (4, 31.142857142857146)
                };
            \end{axis}
        \end{tikzpicture}
        \caption{ \yolo}
    \end{subfigure}
    \hfill
    \centering
    \begin{subfigure}[t]{0.3\textwidth}
    \pgfplotsset{width=4.55cm, height=3.5cm}
    \centering
    \begin{tikzpicture}
        \begin{axis}[
        xlabel={$L_1$ threshold},
        grid=major,
        xticklabels={},
        extra x ticks={1, 2, 3, 4},
        extra x tick labels={2/255, 4/255, 8/255, 10/255}]
            \addplot[red,thick] coordinates {
            (1,0.0)
            (2,0.0)
            (3,0.0)
            (4, 1.4444444444444444)
            };
            \addplot[blue,thick] coordinates {
            (1, 3.148148148148148)
            (2, 3.3333333333333335)
            (3, 3.4074074074074074)
            (4, 3.4074074074074074)
            };
            \addplot[green,thick] coordinates {
            (1, 6.62962962962963)
            (2, 7.777777777777778)
            (3, 8.037037037037036)
            (4, 8.037037037037036)
            };
            \addplot[orange,thick] coordinates {
            (1, 14.666666666666666)
            (2, 21.444444444444443)
            (3, 30.148148148148145)
            (4, 32.88888888888889)
            };
            \addplot[purple,thick] coordinates {
            (1, 11.407407407407408)
            (2, 15.333333333333332)
            (3, 20.814814814814813)
            (4, 21.88888888888889)
            };
        \end{axis}
    \end{tikzpicture}
        \caption{ RT-DETR}
    \end{subfigure}
    \hfill
    \centering
    \begin{subfigure}[t]{0.3\textwidth}
     \pgfplotsset{width=4.55cm, height=3.5cm}
    \centering
    \begin{tikzpicture}
        \begin{axis}[
        xlabel={$L_1$ threshold},
        grid=major,
        xticklabels={},
        extra x ticks={1, 2, 3, 4},
        extra x tick labels={2/255, 4/255, 8/255, 10/255}]
            \addplot[red,thick] coordinates {
            (1,0.0)
            (2,0.0)
            (3,0.0)
            (4, 1.4545454545454546)
            };
            \addplot[blue,thick] coordinates {
            (1, 4.7272727272727275)
            (2, 4.909090909090909)
            (3, 4.909090909090909)
            (4, 4.909090909090909)
            };
            \addplot[green,thick] coordinates {
            (1, 8.90909090909091)
            (2, 9.272727272727273)
            (3, 9.272727272727273)
            (4, 9.272727272727273)
            };
            \addplot[orange,thick] coordinates {
            (1, 10.0)
            (2, 15.090909090909092)
            (3, 22.181818181818183)
            (4, 24.90909090909091)
            };
            \addplot[purple,thick] coordinates {
            (1, 12.363636363636363)
            (2, 15.636363636363637)
            (3, 18.90909090909091)
            (4, 19.818181818181817)
            };
        \end{axis}
    \end{tikzpicture}
        \caption{ FASTER-RCNN}
    \end{subfigure}
            \begin{subfigure}[t]{0.3\textwidth}
        \centering
        \pgfplotsset{width=4.55cm, height=3.5cm}
        \centering
        \begin{tikzpicture}
            \begin{axis}[
            ylabel={Success Rate \%},
            xlabel={$L_2$ threshold},
            grid=major,
            xticklabels={},
            extra x ticks={1, 2, 3, 4},
            extra x tick labels={2/255, 4/255, 8/255, 10/255}]
                \addplot[red,thick] coordinates {
                (1, 0.0)
                (2, 0.0)
                (3, 0.0)
                (4, 0.020408163265306124)
                };
                \addplot[blue,thick] coordinates {
                (1, 1.306122448979592)
                (2, 3.224489795918367)
                (3, 5.244897959183673)
                (4, 5.836734693877551)
                };
                \addplot[green,thick] coordinates {
                (1, 3.5510204081632657)
                (2, 9.306122448979593)
                (3, 18.408163265306122)
                (4, 22.551020408163268)
                };
                \addplot[orange,thick] coordinates {
                (1, 2.183673469387755)
                (2, 6.204081632653061)
                (3, 15.306122448979592)
                (4, 18.79591836734694)
                };
                \addplot[purple,thick] coordinates {
                (1, 2.36734693877551)
                (2, 6.020408163265306)
                (3, 13.448979591836736)
                (4, 15.10204081632653)
                };
            \end{axis}
        \end{tikzpicture}
        \caption{ \yolo}
    \end{subfigure}
    \hfill
    \centering
    \begin{subfigure}[t]{0.3\textwidth}
    \pgfplotsset{width=4.55cm, height=3.5cm}
    \centering
    \begin{tikzpicture}
        \begin{axis}[
        xlabel={$L_2$ threshold},
        grid=major,
        xticklabels={},
        extra x ticks={1, 2, 3, 4},
        extra x tick labels={2/255, 4/255, 8/255, 10/255}]
            \addplot[red,thick] coordinates {
            (1,0.0)
            (2,0.0)
            (3,0.0)
            (4, 0.0)
            };
            \addplot[blue,thick] coordinates {
            (1, 1.6666666666666667)
            (2, 2.2222222222222223)
            (3, 2.9629629629629632)
            (4, 3.111111111111111)
            };
            \addplot[green,thick] coordinates {
            (1, 2.074074074074074)
            (2, 4.111111111111112)
            (3, 5.962962962962963)
            (4, 6.814814814814815)
            };
            \addplot[orange,thick] coordinates {
            (1, 2.5925925925925926)
            (2, 4.703703703703703)
            (3, 8.148148148148149)
            (4, 9.925925925925926)
            };
            \addplot[purple,thick] coordinates {
            (1, 2.259259259259259)
            (2, 4.481481481481482)
            (3, 6.925925925925926)
            (4, 8.481481481481481)
            };
        \end{axis}
    \end{tikzpicture}
        \caption{ RT-DETR}
    \end{subfigure}
    \hfill
    \centering
    \begin{subfigure}[t]{0.3\textwidth}
     \pgfplotsset{width=4.55cm, height=3.5cm}
    \centering
    \begin{tikzpicture}
        \begin{axis}[
        xlabel={$L_2$ threshold},
        grid=major,
        xticklabels={},
        extra x ticks={1, 2, 3, 4},
        extra x tick labels={2/255, 4/255, 8/255, 10/255}]
            \addplot[red,thick] coordinates {
            (1,0.0)
            (2,0.0)
            (3,0.0)
            (4, 0.0)
            };
            \addplot[blue,thick] coordinates {
            (1, 2.5454545454545454)
            (2, 3.6363636363636362)
            (3, 4.545454545454546)
            (4, 4.7272727272727275)
            };
            \addplot[green,thick] coordinates {
            (1, 1.2727272727272727)
            (2, 4.181818181818182)
            (3, 7.454545454545454)
            (4, 8.545454545454545)
            };
            \addplot[orange,thick] coordinates {
            (1, 0.18181818181818182)
            (2, 2.909090909090909)
            (3, 6.7272727272727275)
            (4, 7.818181818181818)
            };
            \addplot[purple,thick] coordinates {
            (1, 3.4545454545454546)
            (2, 6.0)
            (3, 8.90909090909091)
            (4, 9.636363636363637)
            };
        \end{axis}
    \end{tikzpicture}
        \caption{FASTER-RCNN}
    \end{subfigure}
            \begin{subfigure}[t]{0.3\textwidth}
        \centering
        \pgfplotsset{width=4.55cm, height=3.5cm}
        \centering
        \begin{tikzpicture}
            \begin{axis}[
            ylabel={Success Rate \%},
            xlabel={$LPIPS$ threshold},
            grid=major,
            xticklabels={},
            extra x ticks={1, 2, 3, 4},
            extra x tick labels={0.01 , 0.05 , 0.1 , 0.2}]
                \addplot[red,thick] coordinates {
                (1, 0.0)
                (2, 0.12244897959183673)
                (3, 1.183673469387755)
                (4, 5.8979591836734695)
                };
                \addplot[blue,thick] coordinates {
                (1, 1.7551020408163265)
                (2, 4.428571428571428)
                (3, 5.63265306122449)
                (4, 6.836734693877551)
                };
                \addplot[green,thick] coordinates {
                (1, 6.673469387755102)
                (2, 17.20408163265306)
                (3, 24.897959183673468)
                (4, 29.897959183673468)
                };
                \addplot[orange,thick] coordinates {
                (1, 6.551020408163264)
                (2, 21.816326530612244)
                (3, 29.46938775510204)
                (4, 39.9795918367347)
                };
                \addplot[purple,thick] coordinates {
                (1, 7.000000000000001)
                (2, 18.612244897959187)
                (3, 25.448979591836736)
                (4, 31.46938775510204)
                };
            \end{axis}
        \end{tikzpicture}
        \caption{ \yolo}
    \end{subfigure}
    \hfill
    \centering
    \begin{subfigure}[t]{0.3\textwidth}
    \pgfplotsset{width=4.55cm, height=3.5cm}
    \centering
    \begin{tikzpicture}
        \begin{axis}[
        xlabel={$LPIPS$ threshold},
        grid=major,
        xticklabels={},
        extra x ticks={1, 2, 3, 4},
        extra x tick labels={0.01 , 0.05 , 0.1 , 0.2}]
            \addplot[red,thick] coordinates {
            (1,0.0)
            (2,0.1111111111111111)
            (3,0.33333333333333337)
            (4, 1.4814814814814816)
            };
            \addplot[blue,thick] coordinates {
            (1, 2.0)
            (2, 2.851851851851852)
            (3, 3.259259259259259)
            (4, 3.3703703703703702)
            };
            \addplot[green,thick] coordinates {
            (1, 3.4444444444444446)
            (2, 5.9259259259259265)
            (3, 7.000000000000001)
            (4, 7.9259259259259265)
            };
            \addplot[orange,thick] coordinates {
            (1, 5.666666666666666)
            (2, 16.59259259259259)
            (3, 24.51851851851852)
            (4, 33.62962962962963)
            };
            \addplot[purple,thick] coordinates {
            (1, 5.185185185185185)
            (2, 12.074074074074074)
            (3, 17.62962962962963)
            (4, 22.333333333333332)
            };
        \end{axis}
    \end{tikzpicture}
        \caption{ RT-DETR}
    \end{subfigure}
    \hfill
    \centering
    \begin{subfigure}[t]{0.3\textwidth}
     \pgfplotsset{width=4.55cm, height=3.5cm}
    \centering
    \begin{tikzpicture}
        \begin{axis}[
        xlabel={$LPIPS$ threshold},
        grid=major,
        xticklabels={},
        extra x ticks={1, 2, 3, 4},
        extra x tick labels={0.01 , 0.05 , 0.1 , 0.2}]
            \addplot[red,thick] coordinates {
            (1,0.0)
            (2,0.0)
            (3,0.7272727272727273)
            (4,2.181818181818182)
            };
            \addplot[blue,thick] coordinates {
            (1, 2.909090909090909)
            (2, 4.363636363636364)
            (3, 4.909090909090909)
            (4, 4.909090909090909)
            };
            \addplot[green,thick] coordinates {
            (1, 3.4545454545454546)
            (2, 6.909090909090909)
            (3, 8.727272727272728)
            (4, 9.272727272727273)
            };
            \addplot[orange,thick] coordinates {
            (1, 4.181818181818182)
            (2, 10.727272727272727)
            (3, 18.0)
            (4, 26.0)
            };
            \addplot[purple,thick] coordinates {
            (1, 7.090909090909091)
            (2, 13.090909090909092)
            (3, 16.363636363636363)
            (4, 20.18181818181818)
            };
        \end{axis}
    \end{tikzpicture}
        \caption{FASTER-RCNN}
    \end{subfigure}
            \begin{subfigure}[t]{0.3\textwidth}
        \centering
        \pgfplotsset{width=4.55cm, height=3.5cm}
        \centering
        \begin{tikzpicture}
            \begin{axis}[
            ylabel={Success Rate \%},
            xlabel={$SSIM$ threshold},
            grid=major,
            xticklabels={},
            extra x ticks={1, 2, 3, 4},
            extra x tick labels={0.99 , 0.98 , 0.95 , 0.9}]
                \addplot[red,thick] coordinates {
                (1, 0.0)
                (2, 0.0)
                (3, 0.0)
                (4, 0.08)
                };
                \addplot[blue,thick] coordinates {
                (1, 1.55)
                (2, 3.02)
                (3, 5.02)
                (4, 6.43)
                };
                \addplot[green,thick] coordinates {
                (1, 6.1)
                (2, 9.57)
                (3, 18.47)
                (4, 26.82)
                };
                \addplot[orange,thick] coordinates {
                (1, 13.65)
                (2, 19.98)
                (3, 27.27)
                (4, 36.55)
                };
                \addplot[purple,thick] coordinates {
                (1, 11.61)
                (2, 15.88)
                (3, 22.78)
                (4, 29.12)
                };
            \end{axis}
        \end{tikzpicture}
        \caption{ \yolo}
    \end{subfigure}
    \hfill
    \centering
    \begin{subfigure}[t]{0.3\textwidth}
    \pgfplotsset{width=4.55cm, height=3.5cm}
    \centering
    \begin{tikzpicture}
        \begin{axis}[
        xlabel={$SSIM$ threshold},
        grid=major,
        xticklabels={},
        extra x ticks={1, 2, 3, 4},
        extra x tick labels={0.01 , 0.05 , 0.1 , 0.2}]
            \addplot[red,thick] coordinates {
            (1,0.0)
            (2,0.0)
            (3,0.0)
            (4, 0.07)
            };
            \addplot[blue,thick] coordinates {
            (1, 2.07)
            (2, 2.56)
            (3, 3.07)
            (4, 3.37)
            };
            \addplot[green,thick] coordinates {
            (1, 3.59)
            (2, 4.52)
            (3, 6.33)
            (4, 7.67)
            };
            \addplot[orange,thick] coordinates {
            (1, 8.93)
            (2, 13.19)
            (3, 22.0)
            (4, 30.89)
            };
            \addplot[purple,thick] coordinates {
            (1, 6.85)
            (2, 9.85)
            (3, 15.26)
            (4, 20.3)
            };
        \end{axis}
    \end{tikzpicture}
        \caption{ RT-DETR}
    \end{subfigure}
    \hfill
    \centering
    \begin{subfigure}[t]{0.3\textwidth}
     \pgfplotsset{width=4.55cm, height=3.5cm}
    \centering
    \begin{tikzpicture}
        \begin{axis}[
        xlabel={$SSIM$ threshold},
        grid=major,
        xticklabels={},
        extra x ticks={1, 2, 3, 4},
        extra x tick labels={0.01 , 0.05 , 0.1 , 0.2}]
            \addplot[red,thick] coordinates {
            (1,0.0)
            (2,0.0)
            (3,0.0)
            (4, 0.0)
            };
            \addplot[blue,thick] coordinates {
            (1, 3.09)
            (2, 3.82)
            (3, 4.73)
            (4, 4.91)
            };
            \addplot[green,thick] coordinates {
            (1, 4.36)
            (2, 5.64)
            (3, 8.36)
            (4, 9.27)
            };
            \addplot[orange,thick] coordinates {
            (1, 6.91)
            (2, 9.09)
            (3, 15.27)
            (4, 23.82)
            };
            \addplot[purple,thick] coordinates {
            (1, 8.73)
            (2, 11.45)
            (3, 14.91)
            (4, 19.27)
            };
        \end{axis}
    \end{tikzpicture}
        \caption{FASTER-RCNN}
    \end{subfigure}
    \end{subfigure}
    \caption{Success rate of different approaches in \emph{changing the label of the detection}, with different models on COCO dataset, for different thresholds with $L_0, L_1, L_2, LPIPS, SSIM$. The different techniques are
     \textcolor{red}{noise}, \textcolor{blue}{targeted noise}, \textcolor{green}{blended},  \textcolor{orange}{DRISE$_{MoG}$} and \textcolor{purple}{MoG}.}
    \label{fig:success_rate_change_pred}
\end{figure*}

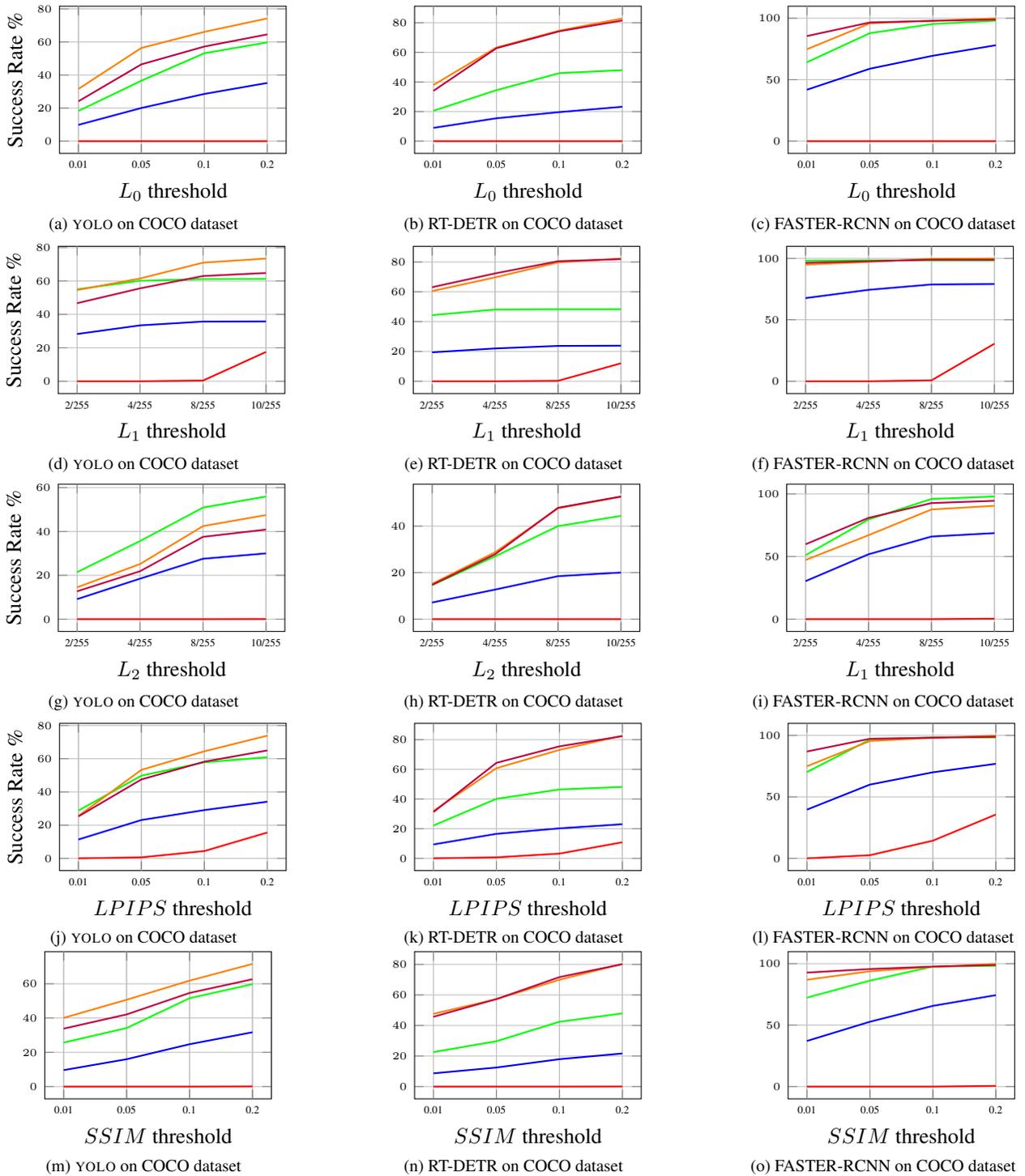
\begin{figure*}
    \begin{subfigure}{\textwidth}
        \centering
            \begin{subfigure}[t]{0.32\textwidth}
        \centering
        \pgfplotsset{width=4.55cm, height=3.5cm}
        \centering
        \begin{tikzpicture}
            \begin{axis}[
            ylabel={Success Rate \%},
            xlabel={$L_0$ threshold},
            grid=major,
            xticklabels={},
            extra x ticks={1, 2, 3, 4},
            extra x tick labels={0.01, 0.05, 0.1, 0.2}]
                \addplot[red,thick] coordinates {
                (1,0.0)
                (2,0.0)
                (3,0.0)
                (4, 0.0)
                };
                \addplot[blue,thick] coordinates {
                (1, 9.83673469387755)
                (2, 20.06122448979592)
                (3, 28.57142857142857)
                (4, 35.224489795918366)
                };
                \addplot[green,thick] coordinates {
                (1, 18.26530612244898)
                (2, 36.57142857142857)
                (3, 53.204081632653065)
                (4, 59.75510204081632)
                };
                \addplot[orange,thick] coordinates {
                (1, 31.591836734693878)
                (2, 56.40816326530612)
                (3, 66.14285714285715)
                (4, 74.26530612244898)
                };
                \addplot[purple,thick] coordinates {
                (1, 24.122448979591837)
                (2, 46.46938775510204)
                (3, 57.244897959183675)
                (4, 64.59183673469387)
                };
            \end{axis}
        \end{tikzpicture}
        \caption{ \yolo}
    \end{subfigure}
    \hfill
    \centering
    \begin{subfigure}[t]{0.32\textwidth}
    \pgfplotsset{width=4.55cm, height=3.5cm}
    \centering
    \begin{tikzpicture}
        \begin{axis}[
        xlabel={$L_0$ threshold},
        grid=major,
        xticklabels={},
        extra x ticks={1, 2, 3, 4},
        extra x tick labels={0.01, 0.05, 0.1, 0.2}]
            \addplot[red,thick] coordinates {
            (1,0.0)
            (2,0.0)
            (3,(0.0)
            (4, 0.0)
            };
            \addplot[blue,thick] coordinates {
            (1, 8.962962962962964)
            (2, 15.444444444444445)
            (3, 19.62962962962963)
            (4, 23.22222222222222)
            };
            \addplot[green,thick] coordinates {
            (1, 20.51851851851852)
            (2, 34.44444444444444)
            (3, 45.96296296296296)
            (4, 48.0)
            };
            \addplot[orange,thick] coordinates {
            (1, 37.96296296296296)
            (2, 63.2962962962963)
            (3, 74.70370370370371)
            (4, 82.96296296296296)
            };
            \addplot[purple,thick] coordinates {
            (1, 34.0)
            (2, 62.81481481481481)
            (3, 74.25925925925925)
            (4, 81.55555555555556)
            };
        \end{axis}
    \end{tikzpicture}
        \caption{ RT-DETR}
    \end{subfigure}
    \hfill
    \centering
    \begin{subfigure}[t]{0.32\textwidth}
     \pgfplotsset{width=4.55cm, height=3.5cm}
    \centering
    \begin{tikzpicture}
        \begin{axis}[
        xlabel={$L_0$ threshold},
        grid=major,
        xticklabels={},
        extra x ticks={1, 2, 3, 4},
        extra x tick labels={0.01, 0.05, 0.1, 0.2}]
            \addplot[red,thick] coordinates {
            (1,0.0)
            (2,0.0)
            (3,0.0)
            (4, 0.0)
            };
            \addplot[blue,thick] coordinates {
            (1, 41.81818181818181)
            (2, 58.909090909090914)
            (3, 69.45454545454545)
            (4, 78.0)
            };
            \addplot[green,thick] coordinates {
            (1, 64.18181818181819)
            (2, 87.81818181818181)
            (3, 95.27272727272728)
            (4, 98.0)
            };
            \addplot[orange,thick] coordinates {
            (1, 74.72727272727273)
            (2, 95.81818181818181)
            (3, 98.0)
            (4, 99.81818181818181)
            };
            \addplot[purple,thick] coordinates {
            (1, 85.45454545454545)
            (2, 96.54545454545455)
            (3, 97.81818181818181)
            (4, 98.9090909090909)
            };
        \end{axis}
    \end{tikzpicture}
        \caption{ FASTER-RCNN}
    \end{subfigure}
            \begin{subfigure}[t]{0.3\textwidth}
        \centering
        \pgfplotsset{width=4.55cm, height=3.5cm}
        \centering
        \begin{tikzpicture}
            \begin{axis}[
            ylabel={Success Rate \%},
            xlabel={$L_1$ threshold},
            grid=major,
            xticklabels={},
            extra x ticks={1, 2, 3, 4},
            extra x tick labels={2/255, 4/255, 8/255, 10/255}]
                \addplot[red,thick] coordinates {
                (1,0.0)
                (2,0.0)
                (3,0.40816326530612246)
                (4, 17.53061224489796)
                };
                \addplot[blue,thick] coordinates {
                (1, 28.244897959183675)
                (2, 33.40816326530612)
                (3, 35.69387755102041)
                (4, 35.775510204081634)
                };
                \addplot[green,thick] coordinates {
                (1, 54.9795918367347)
                (2, 60.10204081632653)
                (3, 61.12244897959184)
                (4, 61.244897959183675)
                };
                \addplot[orange,thick] coordinates {
                (1, 54.55102040816327)
                (2, 61.44897959183674)
                (3, 70.93877551020408)
                (4, 73.38775510204081)
                };
                \addplot[purple,thick] coordinates {
                (1, 46.6734693877551)
                (2, 55.55102040816327)
                (3, 62.93877551020408)
                (4, 64.75510204081633)
                };
            \end{axis}
        \end{tikzpicture}
        \caption{ \yolo}
    \end{subfigure}
    \hfill
    \centering
    \begin{subfigure}[t]{0.3\textwidth}
    \pgfplotsset{width=4.55cm, height=3.5cm}
    \centering
    \begin{tikzpicture}
        \begin{axis}[
        xlabel={$L_1$ threshold},
        grid=major,
        xticklabels={},
        extra x ticks={1, 2, 3, 4},
        extra x tick labels={2/255, 4/255, 8/255, 10/255}]
            \addplot[red,thick] coordinates {
            (1,0.0)
            (2,0.0)
            (3,0.2962962962962963)
            (4, 12.037037037037036)
            };
            \addplot[blue,thick] coordinates {
            (1, 19.407407407407405)
            (2, 22.0)
            (3, 23.703703703703706)
            (4, 23.85185185185185)
            };
            \addplot[green,thick] coordinates {
            (1, 44.333333333333336)
            (2, 48.11111111111111)
            (3, 48.2962962962963)
            (4, 48.2962962962963)
            };
            \addplot[orange,thick] coordinates {
            (1, 60.48148148148148)
            (2, 69.66666666666667)
            (3, 79.74074074074075)
            (4, 82.2962962962963)
            };
            \addplot[purple,thick] coordinates {
            (1, 63.07407407407407)
            (2, 72.33333333333334)
            (3, 80.51851851851852)
            (4, 81.92592592592592)
            };
        \end{axis}
    \end{tikzpicture}
        \caption{ RT-DETR}
    \end{subfigure}
    \hfill
    \centering
    \begin{subfigure}[t]{0.3\textwidth}
     \pgfplotsset{width=4.55cm, height=3.5cm}
    \centering
    \begin{tikzpicture}
        \begin{axis}[
        xlabel={$L_1$ threshold},
        grid=major,
        xticklabels={},
        extra x ticks={1, 2, 3, 4},
        extra x tick labels={2/255, 4/255, 8/255, 10/255}]
            \addplot[red,thick] coordinates {
            (1,0.0)
            (2,0.0)
            (3,0.7272727272727273)
            (4, 30.545454545454547)
            };
            \addplot[blue,thick] coordinates {
            (1, 67.63636363636364)
            (2, 74.36363636363636)
            (3, 78.72727272727272)
            (4, 79.0909090909091)
            };
            \addplot[green,thick] coordinates {
            (1, 98.0)
            (2, 98.36363636363636)
            (3, 98.36363636363636)
            (4, 98.36363636363636)
            };
            \addplot[orange,thick] coordinates {
            (1, 94.9090909090909)
            (2, 97.0909090909091)
            (3, 99.63636363636364)
            (4, 99.81818181818181)
            };
            \addplot[purple,thick] coordinates {
            (1, 96.36363636363636)
            (2, 97.63636363636363)
            (3, 98.9090909090909)
            (4, 98.9090909090909)
            };
        \end{axis}
    \end{tikzpicture}
        \caption{ FASTER-RCNN}
    \end{subfigure}
            \begin{subfigure}[t]{0.3\textwidth}
        \centering
        \pgfplotsset{width=4.55cm, height=3.5cm}
        \centering
        \begin{tikzpicture}
            \begin{axis}[
            ylabel={Success Rate \%},
            xlabel={$L_2$ threshold},
            grid=major,
            xticklabels={},
            extra x ticks={1, 2, 3, 4},
            extra x tick labels={2/255, 4/255, 8/255, 10/255}]
                \addplot[red,thick] coordinates {
                (1, 0.0)
                (2, 0.0)
                (3, 0.0)
                (4, 0.04081632653061225)
                };
                \addplot[blue,thick] coordinates {
                (1, 9.122448979591837)
                (2, 18.46938775510204)
                (3, 27.55102040816326)
                (4, 30.0)
                };
                \addplot[green,thick] coordinates {
                (1, 21.428571428571427)
                (2, 35.63265306122449)
                (3, 50.89795918367347)
                (4, 55.95918367346939)
                };
                \addplot[orange,thick] coordinates {
                (1, 14.510204081632653)
                (2, 25.142857142857146)
                (3, 42.44897959183673)
                (4, 47.48979591836734)
                };
                \addplot[purple,thick] coordinates {
                (1, 12.693877551020408)
                (2, 21.857142857142858)
                (3, 37.55102040816327)
                (4, 40.87755102040816)
                };
            \end{axis}
        \end{tikzpicture}
        \caption{ \yolo}
    \end{subfigure}
    \hfill
    \centering
    \begin{subfigure}[t]{0.3\textwidth}
    \pgfplotsset{width=4.55cm, height=3.5cm}
    \centering
    \begin{tikzpicture}
        \begin{axis}[
        xlabel={$L_2$ threshold},
        grid=major,
        xticklabels={},
        extra x ticks={1, 2, 3, 4},
        extra x tick labels={2/255, 4/255, 8/255, 10/255}]
            \addplot[red,thick] coordinates {
            (1,0.0)
            (2,0.0)
            (3,0.0)
            (4, 0.0)
            };
            \addplot[blue,thick] coordinates {
            (1, 7.148148148148148)
            (2, 12.703703703703702)
            (3, 18.48148148148148)
            (4, 20.037037037037038)
            };
            \addplot[green,thick] coordinates {
            (1, 14.629629629629628)
            (2, 27.0)
            (3, 40.0)
            (4, 44.407407407407405)
            };
            \addplot[orange,thick] coordinates {
            (1, 15.185185185185185)
            (2, 28.74074074074074)
            (3, 47.66666666666667)
            (4, 52.59259259259259)
            };
            \addplot[purple,thick] coordinates {
            (1, 14.74074074074074)
            (2, 27.88888888888889)
            (3, 47.888888888888886)
            (4, 52.74074074074074)
            };
        \end{axis}
    \end{tikzpicture}
        \caption{ RT-DETR}
    \end{subfigure}
    \hfill
    \centering
    \begin{subfigure}[t]{0.3\textwidth}
     \pgfplotsset{width=4.55cm, height=3.5cm}
    \centering
    \begin{tikzpicture}
        \begin{axis}[
        xlabel={$L_1$ threshold},
        grid=major,
        xticklabels={},
        extra x ticks={1, 2, 3, 4},
        extra x tick labels={2/255, 4/255, 8/255, 10/255}]
            \addplot[red,thick] coordinates {
            (1,0.0)
            (2,0.0)
            (3,0.0)
            (4, 0.36363636363636365)
            };
            \addplot[blue,thick] coordinates {
            (1, 30.363636363636363)
            (2, 51.81818181818182)
            (3, 66.0)
            (4, 68.72727272727272)
            };
            \addplot[green,thick] coordinates {
            (1, 51.090909090909086)
            (2, 79.63636363636364)
            (3, 96.0)
            (4, 98.0)
            };
            \addplot[orange,thick] coordinates {
            (1, 47.27272727272727)
            (2, 67.0909090909091)
            (3, 87.63636363636364)
            (4, 90.54545454545455)
            };
            \addplot[purple,thick] coordinates {
            (1, 59.81818181818181)
            (2, 80.9090909090909)
            (3, 92.72727272727272)
            (4, 94.54545454545455)
            };
        \end{axis}
    \end{tikzpicture}
        \caption{FASTER-RCNN}
    \end{subfigure}
            \begin{subfigure}[t]{0.3\textwidth}
        \centering
        \pgfplotsset{width=4.55cm, height=3.5cm}
        \centering
        \begin{tikzpicture}
            \begin{axis}[
            ylabel={Success Rate \%},
            xlabel={$LPIPS$ threshold},
            grid=major,
            xticklabels={},
            extra x ticks={1, 2, 3, 4},
            extra x tick labels={0.01 , 0.05 , 0.1 , 0.2}]
                \addplot[red,thick] coordinates {
                (1, 0.0)
                (2, 0.5918367346938775)
                (3, 4.387755102040816)
                (4, 15.551020408163266)
                };
                \addplot[blue,thick] coordinates {
                (1, 11.346938775510203)
                (2, 23.081632653061224)
                (3, 29.081632653061224)
                (4, 34.12244897959184)
                };
                \addplot[green,thick] coordinates {
                (1, 28.816326530612248)
                (2, 49.795918367346935)
                (3, 57.83673469387756)
                (4, 60.897959183673464)
                };
                \addplot[orange,thick] coordinates {
                (1, 25.53061224489796)
                (2, 53.306122448979586)
                (3, 64.46938775510203)
                (4, 73.89795918367346)
                };
                \addplot[purple,thick] coordinates {
                (1, 25.265306122448976)
                (2, 47.53061224489796)
                (3, 58.224489795918366)
                (4, 65.0)
                };
            \end{axis}
        \end{tikzpicture}
        \caption{ \yolo}
    \end{subfigure}
    \hfill
    \centering
    \begin{subfigure}[t]{0.3\textwidth}
    \pgfplotsset{width=4.55cm, height=3.5cm}
    \centering
    \begin{tikzpicture}
        \begin{axis}[
        xlabel={$LPIPS$ threshold},
        grid=major,
        xticklabels={},
        extra x ticks={1, 2, 3, 4},
        extra x tick labels={0.01 , 0.05 , 0.1 , 0.2}]
            \addplot[red,thick] coordinates {
            (1,0.0)
            (2,0.6666666666666667)
            (3,3.185185185185185)
            (4, 10.814814814814815)
            };
            \addplot[blue,thick] coordinates {
            (1, 9.37037037037037)
            (2, 16.51851851851852)
            (3, 20.22222222222222)
            (4, 23.0)
            };
            \addplot[green,thick] coordinates {
            (1, 22.14814814814815)
            (2, 40.11111111111111)
            (3, 46.44444444444444)
            (4, 48.148148148148145)
            };
            \addplot[orange,thick] coordinates {
            (1, 31.925925925925924)
            (2, 60.77777777777777)
            (3, 73.07407407407408)
            (4, 82.7037037037037)
            };
            \addplot[purple,thick] coordinates {
            (1, 31.222222222222225)
            (2, 64.37037037037037)
            (3, 75.48148148148148)
            (4, 82.33333333333334)
            };
        \end{axis}
    \end{tikzpicture}
        \caption{ RT-DETR}
    \end{subfigure}
    \hfill
    \centering
    \begin{subfigure}[t]{0.3\textwidth}
     \pgfplotsset{width=4.55cm, height=3.5cm}
    \centering
    \begin{tikzpicture}
        \begin{axis}[
        xlabel={$LPIPS$ threshold},
        grid=major,
        xticklabels={},
        extra x ticks={1, 2, 3, 4},
        extra x tick labels={0.01 , 0.05 , 0.1 , 0.2}]
            \addplot[red,thick] coordinates {
            (1,0.0)
            (2,2.5454545454545454)
            (3,14.363636363636365)
            (4,35.63636363636364)
            };
            \addplot[blue,thick] coordinates {
            (1, 39.63636363636363)
            (2, 60.0)
            (3, 70.0)
            (4, 76.9090909090909)
            };
            \addplot[green,thick] coordinates {
            (1, 70.18181818181817)
            (2, 96.54545454545455)
            (3, 98.36363636363636)
            (4, 98.36363636363636)
            };
            \addplot[orange,thick] coordinates {
            (1, 74.90909090909092)
            (2, 95.45454545454545)
            (3, 98.0)
            (4, 99.81818181818181)
            };
            \addplot[purple,thick] coordinates {
            (1, 86.9090909090909)
            (2, 97.27272727272728)
            (3, 98.18181818181819)
            (4, 98.9090909090909)
            };
        \end{axis}
    \end{tikzpicture}
        \caption{FASTER-RCNN}
    \end{subfigure}
            \begin{subfigure}[t]{0.3\textwidth}
        \centering
        \pgfplotsset{width=4.55cm, height=3.5cm}
        \centering
        \begin{tikzpicture}
            \begin{axis}[
        xlabel={$SSIM$ threshold},
        grid=major,
        xticklabels={},
        extra x ticks={1, 2, 3, 4},
        extra x tick labels={0.01 , 0.05 , 0.1 , 0.2}]
            \addplot[red,thick] coordinates {
            (1,0.0)
            (2,0.0)
            (3,0.0)
            (4,0.1)
            };
            \addplot[blue,thick] coordinates {
            (1,  9.61 )
            (2,  15.96 )
            (3,  24.71 )
            (4,  31.69 )
            };
            \addplot[green,thick] coordinates {
            (1,  25.67 )
            (2,  34.14 )
            (3,  51.55 )
            (4,  59.78 )
            };
            \addplot[orange,thick] coordinates {
            (1,  40.08 )
            (2,  50.65 )
            (3,  61.8 )
            (4,  71.59 )
            };
            \addplot[purple,thick] coordinates {
            (1,  33.8 )
            (2,  42.08 )
            (3,  54.63 )
            (4,  62.65 )
            };
        \end{axis}
        \end{tikzpicture}
        \caption{ \yolo}
    \end{subfigure}
    \hfill
    \centering
    \begin{subfigure}[t]{0.3\textwidth}
    \pgfplotsset{width=4.55cm, height=3.5cm}
    \centering
    \begin{tikzpicture}
        \begin{axis}[
        xlabel={$SSIM$ threshold},
        grid=major,
        xticklabels={},
        extra x ticks={1, 2, 3, 4},
        extra x tick labels={0.01 , 0.05 , 0.1 , 0.2}]
            \addplot[red,thick] coordinates {
            (1,0.0)
            (2,0.0)
            (3,0.0)
            (4, 0.07)
            };
            \addplot[blue,thick] coordinates {
            (1,  8.67 )
            (2,  12.48 )
            (3,  17.93 )
            (4,  21.67 )
            };
            \addplot[green,thick] coordinates {
            (1,  22.56 )
            (2,  29.7 )
            (3,  42.41 )
            (4,  47.89 )
            };
            \addplot[orange,thick] coordinates {
            (1,  47.59 )
            (2,  57.26 )
            (3,  69.93 )
            (4,  80.33 )
            };
            \addplot[purple,thick] coordinates {
            (1,  45.7 )
            (2,  57.26 )
            (3,  71.67 )
            (4,  80.15 )
            };
        \end{axis}
    \end{tikzpicture}
        \caption{ RT-DETR}
    \end{subfigure}
    \hfill
    \centering
    \begin{subfigure}[t]{0.3\textwidth}
     \pgfplotsset{width=4.55cm, height=3.5cm}
    \centering
    \begin{tikzpicture}
        \begin{axis}[
        xlabel={$SSIM$ threshold},
        grid=major,
        xticklabels={},
        extra x ticks={1, 2, 3, 4},
        extra x tick labels={0.01 , 0.05 , 0.1 , 0.2}]
            \addplot[red,thick] coordinates {
            (1,0.0)
            (2,0.0)
            (3,0.0)
            (4,0.55)
            };
            \addplot[blue,thick] coordinates {
            (1,  37.09 )
            (2,  52.73 )
            (3,  65.64 )
            (4,  74.36 )
            };
            \addplot[green,thick] coordinates {
            (1,  72.36 )
            (2,  86.18 )
            (3,  97.64 )
            (4,  98.36 )
            };
            \addplot[orange,thick] coordinates {
            (1,  86.91 )
            (2,  93.82 )
            (3,  97.45 )
            (4,  99.82 )
            };
            \addplot[purple,thick] coordinates {
            (1,  92.73 )
            (2,  95.64 )
            (3,  97.64 )
            (4,  98.91 )
            };
        \end{axis}
    \end{tikzpicture}
        \caption{FASTER-RCNN}
    \end{subfigure}
    \end{subfigure}
    \caption{Success rate of different approaches in \emph{adding new spurious detection}, with different models on COCO dataset, for different thresholds with $L_0, L_1, L_2, LPIPS, SSIM$. The different techniques are
     \textcolor{red}{noise}, \textcolor{blue}{targeted noise}, \textcolor{green}{blended},  \textcolor{orange}{DRISE$_{MoG}$} and \textcolor{purple}{MoG}.}
    \label{fig:success_rate_added_pred}
\end{figure*}

\Cref{fig:success_rate_no_pred,fig:success_rate_change_pred,fig:success_rate_added_pred} show that success rates grow monotonically with larger distortion budgets for all similarity measures, $L_0, L_1, L_2, LPIPS, SSIM$ and that \msps-guided methods consistently outperform undirected baselines. The performance gap is model-dependent: \yolo and FASTER-R-CNN exhibit the largest gains from causal priors, while RT-DETR is noticeably more robust (lower absolute success). \blackcattMOG{} deliver the best label-change performance across thresholds, while \blackcattGreedy{} and \blackcattMOG{} lead in producing spurious detections — all markedly outperforming Noise and Targeted Noise. For removal attacks, \blackcattMOG{}-DRISE performs best overall, except on Faster-R-CNN where \blackcattMOG{}-\rex is superior. Although the top-performing \blackcatt{} variant varies slightly by metric, every \blackcatt{} method reliably exceeds baseline performance.

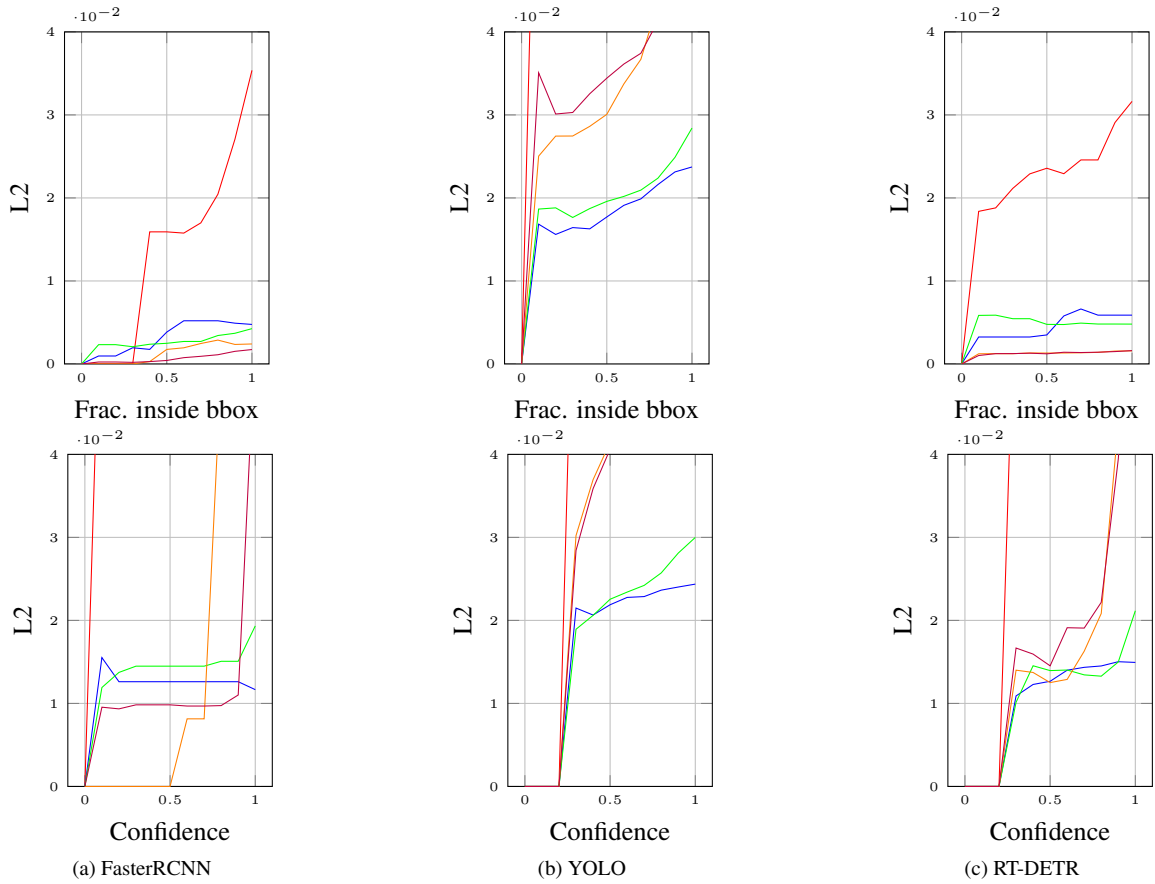
\begin{figure*}[t!]
    \centering
        \begin{subfigure}{0.33\textwidth}
            \centering
            \pgfplotsset{height=6cm, width=4.3cm}
    \begin{tikzpicture}
        \begin{axis} [grid=major, xlabel=Frac. inside bbox, ylabel=L2, ymin=0, ymax=0.04]
            \addplot[red] coordinates {
                (0. , 0)
                (0.1, 0) 
                (0.2, 0)
                (0.3, 0)
                (0.4, 0.015914)
                (0.5, 0.015914)
                (0.6, 0.015765)
                (0.7, 0.016995)
                (0.8, 0.020443)
                (0.9, 0.027063)
                (1.,  0.035365) };
                \addplot[blue] coordinates {
                (0. , 0)
                (0.1, 0.000961956) 
                (0.2, 0.000961956)
                (0.3, 0.001954361)
                (0.4, 0.001755963)
                (0.5, 0.003839624)
                (0.6, 0.005204102)
                (0.7, 0.005204102)
                (0.8, 0.005204102)
                (0.9, 0.004915274)
                (1.,  0.004764165) };
                \addplot[green] coordinates {
                (0. , 0)
                (0.1, 0.002327560365481779) 
                (0.2, 0.002327560365481779)
                (0.3, 0.002077236891168139)
                (0.4, 0.0023774594052821904)
                (0.5, 0.002490924351539378)
                (0.6, 0.002713283261354016)
                (0.7, 0.002713283261354016)
                (0.8, 0.0034264653063773177)
                (0.9, 0.0036989677459241155)
                (1.,  0.004256596384309099) };
                \addplot[orange] coordinates {
                (0. , 0)
                (0.1, 0.) 
                (0.2, 0.)
                (0.3, 0.)
                (0.4, 0.00028915405374049097)
                (0.5, 0.0017529213621349895)
                (0.6, 0.001949008409906373)
                (0.7, 0.0024672099929920034)
                (0.8, 0.002877406163801381)
                (0.9, 0.0023533504563393905)
                (1.,  0.002410848874711278) };
                \addplot[purple] coordinates {
                (0. , 0)
                (0.1, 0.00023897399569163094) 
                (0.2, 0.00023897399569163094)
                (0.3,  0.00019640708002642755)
                (0.4, 0.00028915405374049097)
                (0.5, 0.00042346427316682973)
                (0.6, 0.0007614008934484683)
                (0.7, 0.0009242985028427012)
                (0.8, 0.0011135128288331671)
                (0.9, 0.0015230654167688873)
                (1.,  0.0017377439444084917) };
        \end{axis}
    \end{tikzpicture}
            \label{fig:results:fasterrcnn}
        \end{subfigure}
        \hfill 
        \begin{subfigure}{0.33\textwidth}
            \centering

\pgfplotsset{height=6cm,width=4.3cm}
    \begin{tikzpicture}
        \begin{axis} [grid=major,ylabel=L2, xlabel=Frac. inside bbox, ymin=0, ymax=0.04]
            \addplot[red] coordinates {
                (0, 0) 
                (0.1, 0.08320852581462737)
                (0.2, 0.08195108485976847)
                (0.3, 0.08572651466353923)
                (0.4, 0.08921979299270241)
                (0.5, 0.09034667435477314)
                (0.6, 0.09304108097837226)
                (0.7, 0.09585395118864608)
                (0.8, 0.10011240706598877)
                (0.9, 0.10501295182108981)
                (1, 0.11117842872841553) };
                \addplot[blue] coordinates {
                ((0, 0)
                (0.1, 0.016837871764769537)
                (0.2, 0.015601966517291293)
                (0.3, 0.016427738104390098)
                (0.4, 0.01627566904655837)
                (0.5, 0.01770405103577653)
                (0.6, 0.01910576832097923)
                (0.7, 0.019886603677440964)
                (0.8, 0.021607573982107987)
                (0.9, 0.02313109024407403)
                (1, 0.023731792680532073) };
                \addplot[green] coordinates {
                (0, 0)
                (0.1, 0.018658528249575024)
                (0.2, 0.01880501685822469)
                (0.3, 0.01764743396610397)
                (0.4, 0.018716429687854433)
                (0.5, 0.01958262842179587)
                (0.6, 0.02018175525364561)
                (0.7, 0.02094077877996134)
                (0.8, 0.02237725164850826)
                (0.9, 0.024884658662103855)
                (1, 0.02841476325018589) };
                \addplot[orange] coordinates {
                (0, 0)
                (0.1, 0.025033790277161426)
                (0.2, 0.027448110429004784)
                (0.3, 0.027458358321597962)
                (0.4, 0.02862171181711228)
                (0.5, 0.030065703515847034)
                (0.6, 0.03372566000106226)
                (0.7, 0.03669225467560543)
                (0.8, 0.04345383696475202)
                (0.9, 0.051142633353923946)
                (1, 0.06408498647954163) };
                \addplot[purple] coordinates {
                (0, 0)
                (0.1, 0.035069095610972115)
                (0.2, 0.030107535412214087)
                (0.3, 0.030293230895908834)
                (0.4, 0.03254877961922632)
                (0.5, 0.03442084861968929)
                (0.6, 0.03612441381671631)
                (0.7, 0.03742092840640978)
                (0.8, 0.04129748622479607)
                (0.9, 0.045357336759398445)
                (1, 0.05533661327367787) };
        \end{axis}
    \end{tikzpicture}
            \label{fig:results:yolo}
        \end{subfigure}
        \hfill
         \begin{subfigure}{0.33\textwidth}
            \centering
            \pgfplotsset{height=6cm, width=4.3cm}
    \begin{tikzpicture}
        \begin{axis} [grid=major, ylabel=L2, xlabel=Frac. inside bbox, ymin=0, ymax=0.04]
            \addplot[red] coordinates {
                (0., 0)
                (0.1, 0.018383087279681888)
                (0.2, 0.018802674049475908)
                (0.3, 0.02114513327061485)
                (0.4, 0.02289699030957801)
                (0.5, 0.02357770271013998)
                (0.6, 0.02291937584399454)
                (0.7, 0.024577311628466193)
                (0.8, 0.024577311628466193)
                (0.9, 0.029082649794570568)
                (1., 0.03162633632874811)
                };
                \addplot[blue] coordinates {
                (0,0)
                (0.1 ,0.0032494943297804294)
                (0.2 ,0.0032494943297804294)
                (0.3 ,0.0032494943297804294)
                (0.4 ,0.0032494943297804294)
                (0.5 ,0.0034976883695046826)
                (0.6 ,0.005786427893057766)
                (0.7 ,0.006624042450225668)
                (0.8 ,0.0058823376830597)
                (0.9 ,0.0058823376830597)
                (1. ,0.0058823376830597)
                };
                \addplot[green] coordinates {
                (0, 0)
                (0.1 ,0.005851876009849297)
                (0.2 ,0.005878692474313147)
                (0.3 ,0.005454130799875688)
                (0.4 ,0.005454130799875688)
                (0.5 ,0.0047714446646767725)
                (0.6 ,0.00475213252268092)
                (0.7 ,0.0049236908421139555)
                (0.8 ,0.00481170321810741)
                (0.9 ,0.00481170321810741)
                (1. ,0.004803146689633664)
                };
                \addplot[orange] coordinates {
                (0. ,0)
                (0.1 ,0.0012271063178076926)
                (0.2 ,0.0012439633579700785)
                (0.3 ,0.0012369190137019857)
                (0.4 ,0.0013257894725545382)
                (0.5 ,0.0013146806422796005)
                (0.6 ,0.0013267777971442468)
                (0.7 ,0.0013491424220169028)
                (0.8 ,0.0013981312675933723)
                (0.9 ,0.0015546416866645677)
                (1. ,0.0015803730102993646)
                };
                \addplot[purple] coordinates {
                (0., 0)
                (0.1, 0.0010189578509370806)
                (0.2, 0.0012455647584033247)
                (0.3, 0.0012492007247717374)
                (0.4, 0.001299574865756173)
                (0.5, 0.0012228956990146835)
                (0.6, 0.001413937185847095)
                (0.7, 0.0013786093303605083)
                (0.8, 0.0014187570370630422)
                (0.9, 0.0014929942333575309)
                (1., 0.001610418857459393)
                };
        \end{axis}
    \end{tikzpicture}
            \label{fig:results:rfdetr}
        \end{subfigure}  
        \hfill
        \begin{subfigure}{0.33\textwidth}
            \centering
            \pgfplotsset{height=6cm, width=4.3cm}
    \begin{tikzpicture}
        \begin{axis} [grid=major, xlabel=Confidence, ylabel=L2, ymin=0, ymax=0.04]
            \addplot[red] coordinates {
               (0, 0)
                (0.1, 0.06710005685397821)
                (0.2, 0.06579866779765357)
                (0.3, 0.06814938747754105)
                (0.4, 0.06814938747754105)
                (0.5, 0.0721096996844108)
                (0.6, 0.07604623112418209)
                (0.7, 0.07604623112418209)
                (0.8, 0.0759596121196998)
                (0.9, 0.07898261241699801)
                (1, 0.1463763388898223)
                };
                \addplot[blue] coordinates {
                (0, 0)
                (0.1, 0.01551789742886317)
                (0.2, 0.012614598612566792)
                (0.3, 0.012614598612566792)
                (0.4, 0.012614598612566792)
                (0.5, 0.012614598612566792)
                (0.6, 0.012614598612566792)
                (0.7, 0.012614598612566792)
                (0.8, 0.012614598612566792)
                (0.9, 0.012614598612566792)
                (1, 0.01166086700422956)
                };
                \addplot[green] coordinates {
               (0, 0)
                (0.1, 0.011916102272535405)
                (0.2, 0.013734810553878207)
                (0.3, 0.014487722990776138)
                (0.4, 0.014487722990776138)
                (0.5, 0.014487722990776138)
                (0.6, 0.014487722990776138)
                (0.7, 0.014487722990776138)
                (0.8, 0.01507352784529017)
                (0.9, 0.01507352784529017)
                (1, 0.019334819478764576)
                };
                \addplot[orange] coordinates {
                (0, 0)
                (0.1, 0)
                (0.2, 0)
                (0.3, 0)
                (0.4, 0)
                (0.5, 0)
                (0.6, 0.0081433188759352)
                (0.7, 0.0081433188759352)
                (0.8, 0.0503078081638497)
                (0.9, 0.053763873640567374)
                (1, 0.08111058285286023)
                };
                \addplot[purple] coordinates {
                (0, 0)
                (0.1, 0.00954617108562282)
                (0.2, 0.00934464773684114)
                (0.3, 0.009829340848742837)
                (0.4, 0.009829340848742837)
                (0.5, 0.009829340848742837)
                (0.6, 0.009685190295808721)
                (0.7, 0.009685190295808721)
                (0.8, 0.0097397690849514)
                (0.9, 0.011015352843186211)
                (1, 0.05444450963941236)
                };
        \end{axis}
    \end{tikzpicture}
            \caption{FasterRCNN}
            \label{fig:results:fasterrcnnl2}
        \end{subfigure}
        \hfill 
        \begin{subfigure}{0.33\textwidth}
            \centering
            \pgfplotsset{height=6cm,width=4.3cm}
    \begin{tikzpicture}
        \begin{axis} [grid=major,xlabel=Confidence, ylabel=L2, ymin=0, ymax=0.04]
            \addplot[red] coordinates {
             (0, 0)
            (0.1, 0)
            (0.2, 0)
            (0.3, 0.07561102842196397)
            (0.4, 0.07621735790534129)
            (0.5, 0.07918117700436428)
            (0.6, 0.08334867229253694)
            (0.7, 0.08696460277295222)
            (0.8, 0.09358053593283723)
            (0.9, 0.10761007106408299)
            (1, 0.11642174024526991)
             };
            \addplot[blue] coordinates {
            (0, 0)
            (0.1, 0)
            (0.2, 0)
            (0.3, 0.021486763010886697)
            (0.4, 0.020646275004092017)
            (0.5, 0.02188376015663315)
            (0.6, 0.022766944164491872)
            (0.7, 0.02287053057999055)
            (0.8, 0.02364306879042214)
            (0.9, 0.024019969150339115)
            (1, 0.02436998884717868)
            };
            \addplot[green] coordinates {
            (0, 0)
            (0.1, 0)
            (0.2, 0)
            (0.3, 0.018934673729144817)
            (0.4, 0.020602186772253676)
            (0.5, 0.022548110339763755)
            (0.6, 0.02340841009406534)
            (0.7, 0.024217409003153743)
            (0.8, 0.025716424222408673)
            (0.9, 0.02810720271965609)
            (1, 0.029991746505225056)
            };
            \addplot[orange] coordinates {
            (0, 0)
            (0.1, 0)
            (0.2, 0)
            (0.3, 0.030215408925560135)
            (0.4, 0.03694250446247326)
            (0.5, 0.0414490777799228)
            (0.6, 0.04410668931545206)
            (0.7, 0.04722942146071358)
            (0.8, 0.05207981294475096)
            (0.9, 0.06249148078058696)
            (1, 0.07258532504484974)
            };
            \addplot[purple] coordinates {
            (0, 0)
            (0.1, 0)
            (0.2, 0)
            (0.3, 0.028418338200613964)
            (0.4, 0.03586155742718268)
            (0.5, 0.040622279034035064)
            (0.6, 0.04171452593262738)
            (0.7, 0.04329875803721434)
            (0.8, 0.04701458245293713)
            (0.9, 0.054556666481716726)
            (1, 0.06151077615232936)
            };
        \end{axis}
    \end{tikzpicture}
            \caption{YOLO}
            \label{fig:results:yolol2}
        \end{subfigure}
        \hfill
         \begin{subfigure}{0.33\textwidth}
            \centering
            \pgfplotsset{height=6cm, width=4.3cm}
    \begin{tikzpicture}
        \begin{axis} [grid=major, xlabel=Confidence, ylabel=L2, ymin=0, ymax=0.04]
            \addplot[red] coordinates {
               (0, 0)
                (0.1, 0)
                (0.2, 0)
                (0.3, 0.06637182416692027)
                (0.4, 0.07717553740469309)
                (0.5, 0.0778892500627432)
                (0.6, 0.08111268547510071)
                (0.7, 0.08000994021416011)
                (0.8, 0.08451698036698184)
                (0.9, 0.10060708691399328)
                (1, 0.13692544684013347)
                };
                \addplot[blue] coordinates {
                (0, 0)
                (0.1, 0)
                (0.2, 0)
                (0.3, 0.010898690188090011)
                (0.4, 0.012283300418007594)
                (0.5, 0.012665336918269534)
                (0.6, 0.014002274574963681)
                (0.7, 0.014347004208945335)
                (0.8, 0.01451278366467128)
                (0.9, 0.015024500390735544)
                (1, 0.014938850153666356)
                };
                \addplot[green] coordinates {
                (0, 0)
                (0.1, 0)
                (0.2, 0)
                (0.3, 0.010135309756712561)
                (0.4, 0.014535139751521604)
                (0.5, 0.013953445593531842)
                (0.6, 0.014022727617249181)
                (0.7, 0.01343859098043584)
                (0.8, 0.013280873334959442)
                (0.9, 0.014992070349980739)
                (1, 0.02116266556728006)
                };
                \addplot[orange] coordinates {
                (0, 0)
                (0.1, 0)
                (0.2, 0)
                (0.3, 0.014006119968925474)
                (0.4, 0.013728192578695444)
                (0.5, 0.0125091180103828)
                (0.6, 0.012893029103644305)
                (0.7, 0.016287858033588485)
                (0.8, 0.02083112508566292)
                (0.9, 0.043622224301955614)
                (1, 0.08118064096758768)
                };
                \addplot[purple] coordinates {
                (0, 0)
                (0.1, 0)
                (0.2, 0)
                (0.3, 0.016672410797245265)
                (0.4, 0.01594232070818335)
                (0.5, 0.014528584306775009)
                (0.6, 0.01911857929462138)
                (0.7, 0.019073687090162696)
                (0.8, 0.022177324901140955)
                (0.9, 0.03943182299759994)
                (1, 0.0670101820972801)
                };
        \end{axis}
    \end{tikzpicture}
            \caption{RT-DETR}
            \label{fig:results:rfdetrl2}
        \end{subfigure}  
        \hfill
    \caption{Complete results comparing $L_2$ against confidence for Faster-R-CNN, \yolo and RT-DETR the different approaches are \textcolor{red}{noise}, \textcolor{blue}{targeted noise}, \textcolor{green}{blended},  \textcolor{orange}{DRISE$_{MoG}$} and \textcolor{purple}{MoG}.}
    \label{fig:resultscomplete}
\end{figure*}

\Cref{fig:resultscomplete} shows the relationship between confidence and FIN and $L_2$ score across the board for all of the methods for the successful attack only. We are able to see that $L_2$ generally increases with confidence and FIN suggesting more effort is required to get a successful attack for all methods.

\begin{figure*}[t!]
    \centering
    \begin{subfigure}[t]{0.24\linewidth}\centering
    \includegraphics[width=\linewidth]{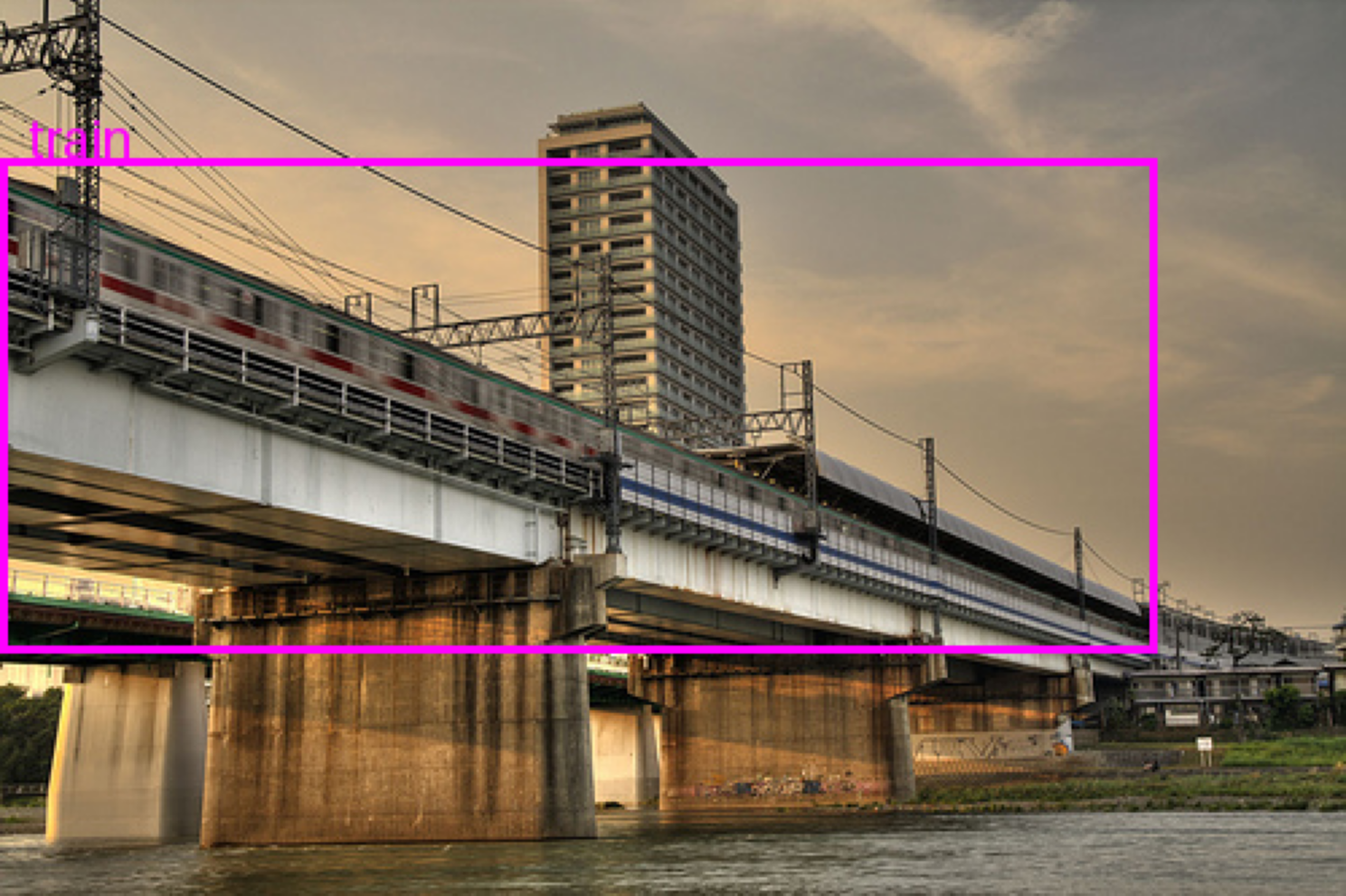}
    \caption{Original (bbox)}
     \end{subfigure}%
  \begin{subfigure}[t]{0.24\linewidth}\centering
    \includegraphics[width=\linewidth]{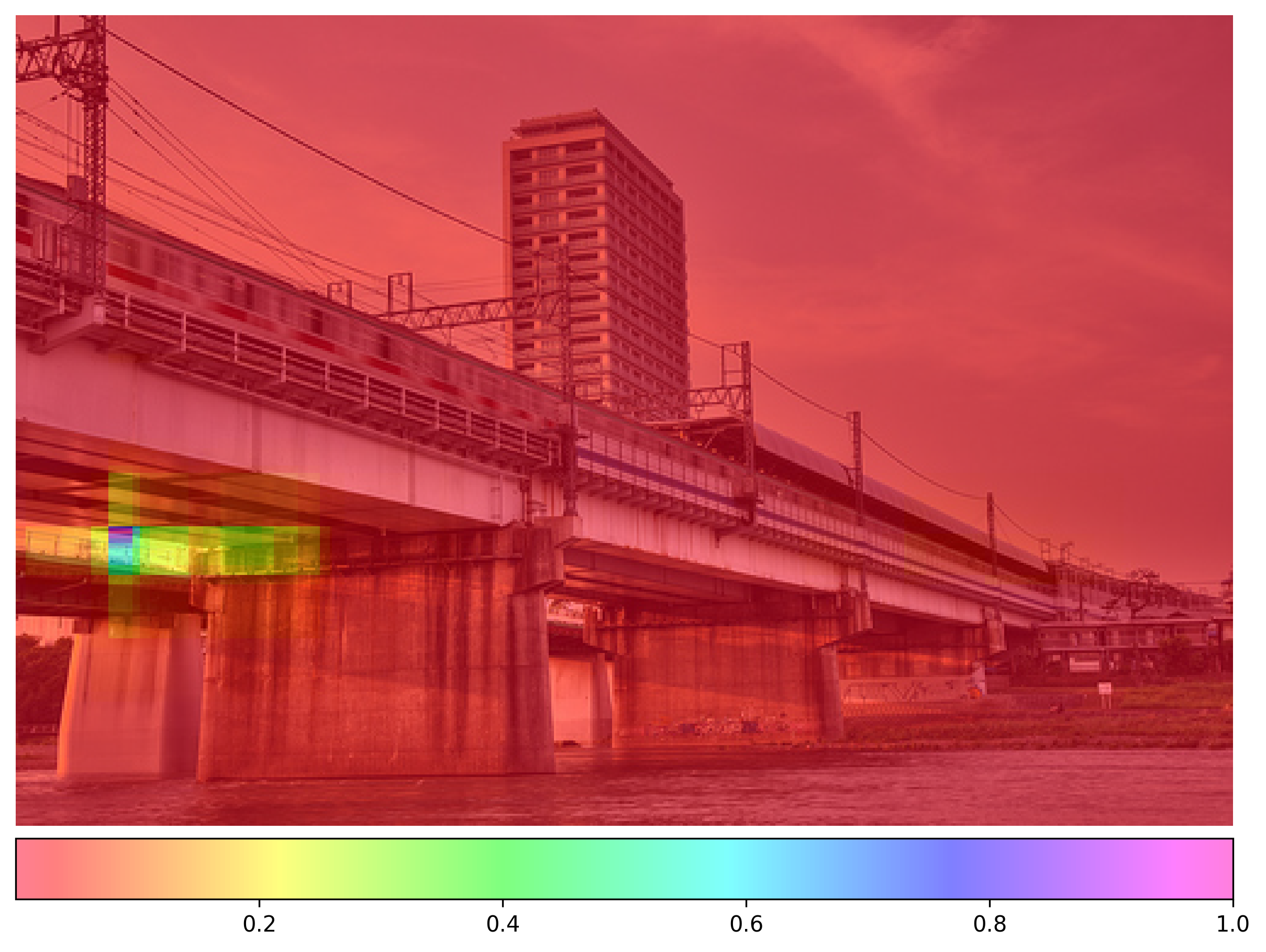}
    \caption{ReX Responsibility Map}
  \end{subfigure}%
  \begin{subfigure}[t]{0.24\linewidth}\centering
    \includegraphics[width=\linewidth]{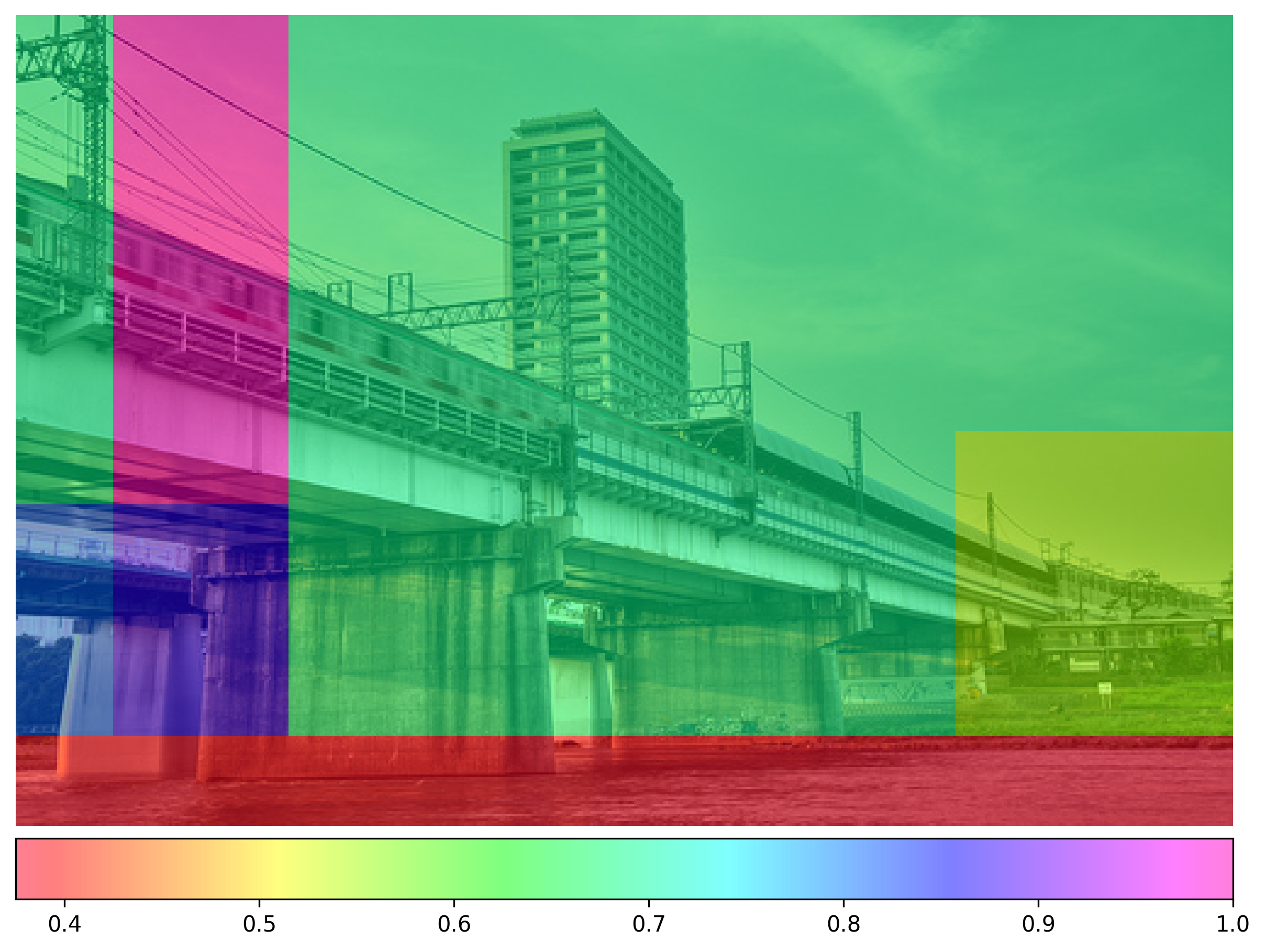}
    \caption{ReX Negative Responsibility Map}
  \end{subfigure}%
  \begin{subfigure}[t]{0.24\linewidth}\centering
    \includegraphics[width=\linewidth]{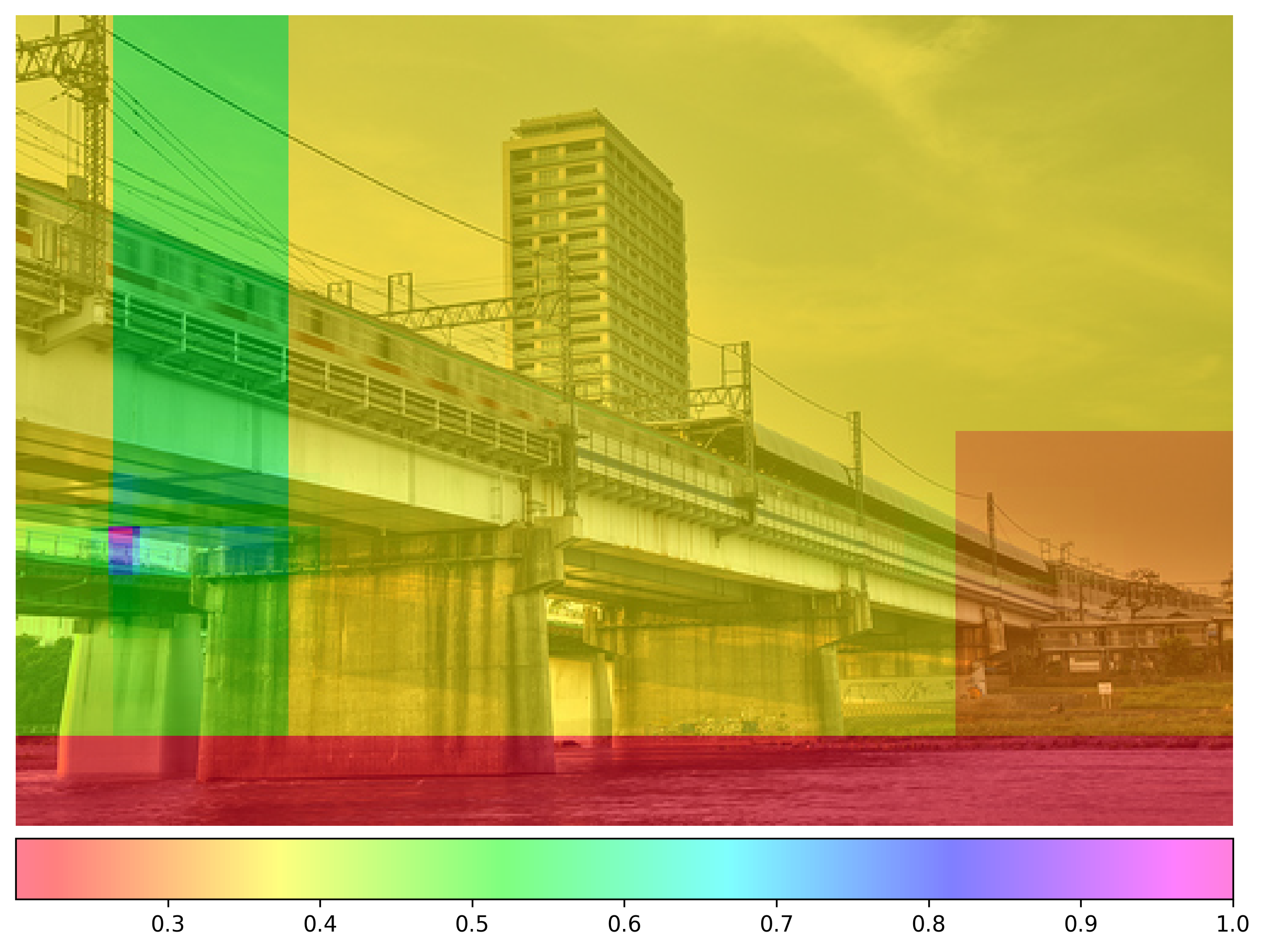}
    \caption{Combined responsibility}
  \end{subfigure}

  \begin{subfigure}[t]{0.24\linewidth}\centering
    \includegraphics[width=\linewidth]{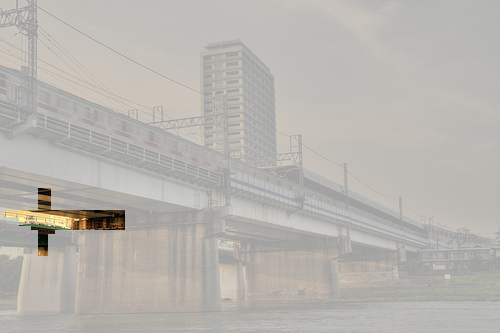}
    \caption{ReX \msps (overlay)}
  \end{subfigure}%
  \begin{subfigure}[t]{0.24\linewidth}\centering
    \includegraphics[width=\linewidth]{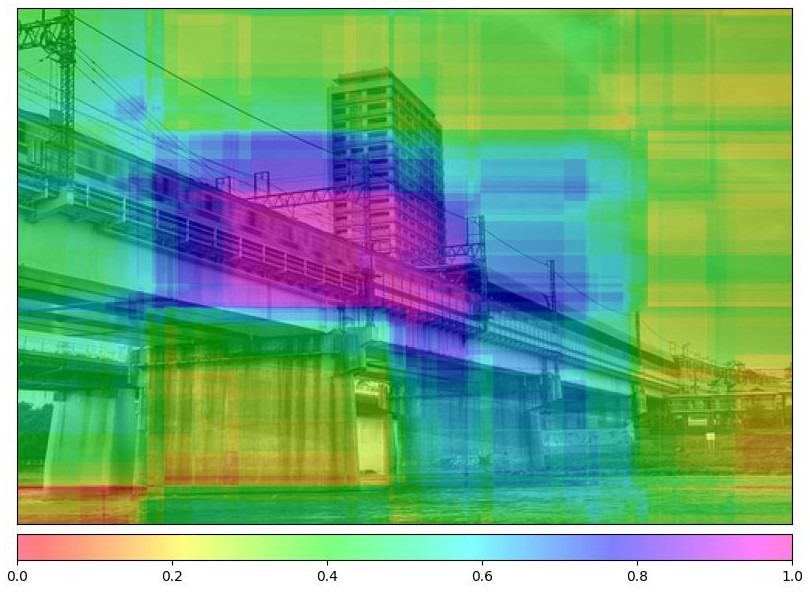}
    \caption{DRISE Saliency map}
  \end{subfigure}%
  \begin{subfigure}[t]{0.24\linewidth}\centering
    \includegraphics[width=\linewidth]{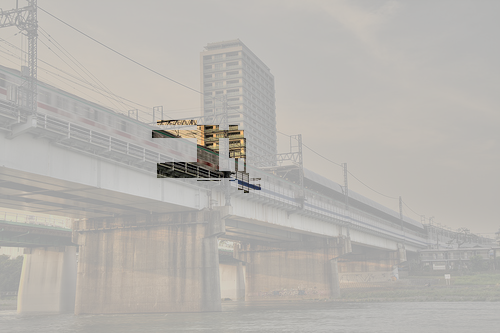}
    \caption{DRISE \msps (overlay)}
  \end{subfigure}

    \begin{subfigure}[ht]{0.29\linewidth}\centering
    \includegraphics[width=\linewidth]{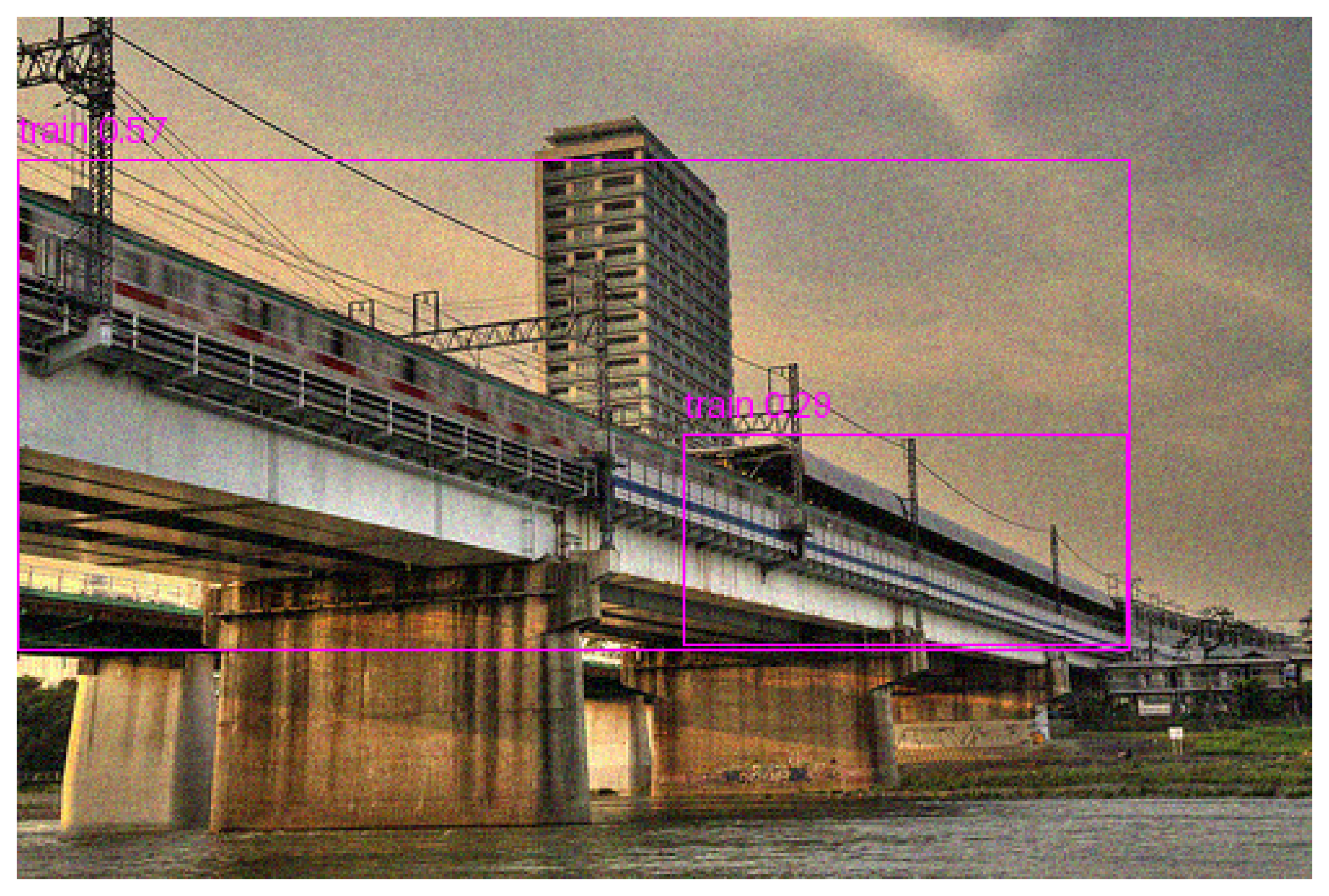}
    \caption{Noise \\ Outcome: added another train \\ LPIPS: 0.926  $L_2$:  0.1934}
    \end{subfigure}
  \begin{subfigure}[ht]{0.29\linewidth}\centering
    \includegraphics[width=\linewidth]{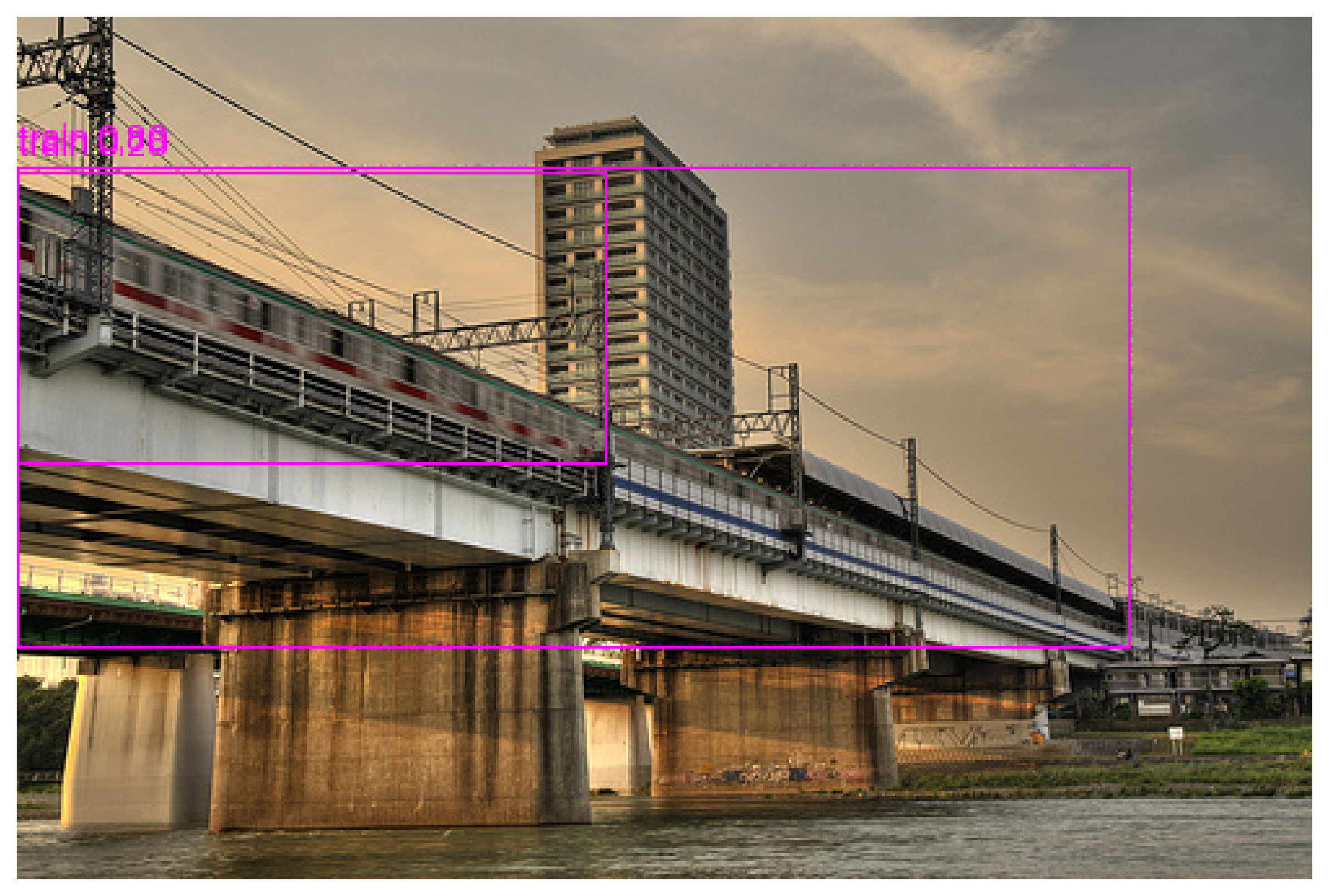}
    \caption{Targeted noise \\ Outcome: add another train\\ LPIPS: 0.019   $L_2$:  0.0062}
  \end{subfigure}
  \centering
  \begin{subfigure}[ht]{0.29\linewidth}\centering
    \includegraphics[width=\linewidth]{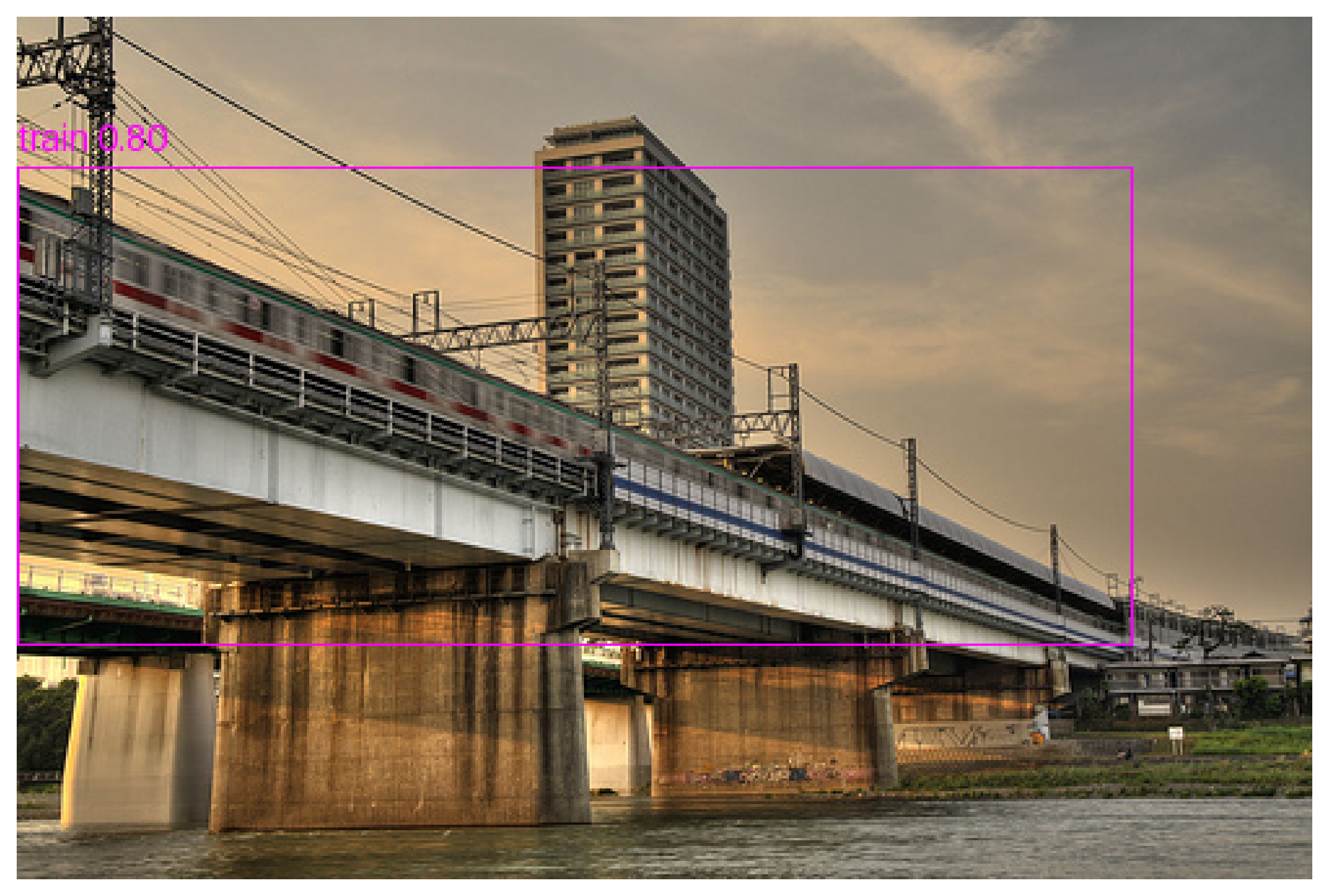}
    \caption{Blended \\ Outcome: added another train (on top)\\ LPIPS: 5.20e-06   $L_2$:  0.0006}
    \label{fig:combined_example_train:blended}
  \end{subfigure}
  \begin{subfigure}[ht]{0.29\linewidth}\centering
    \includegraphics[width=\linewidth]{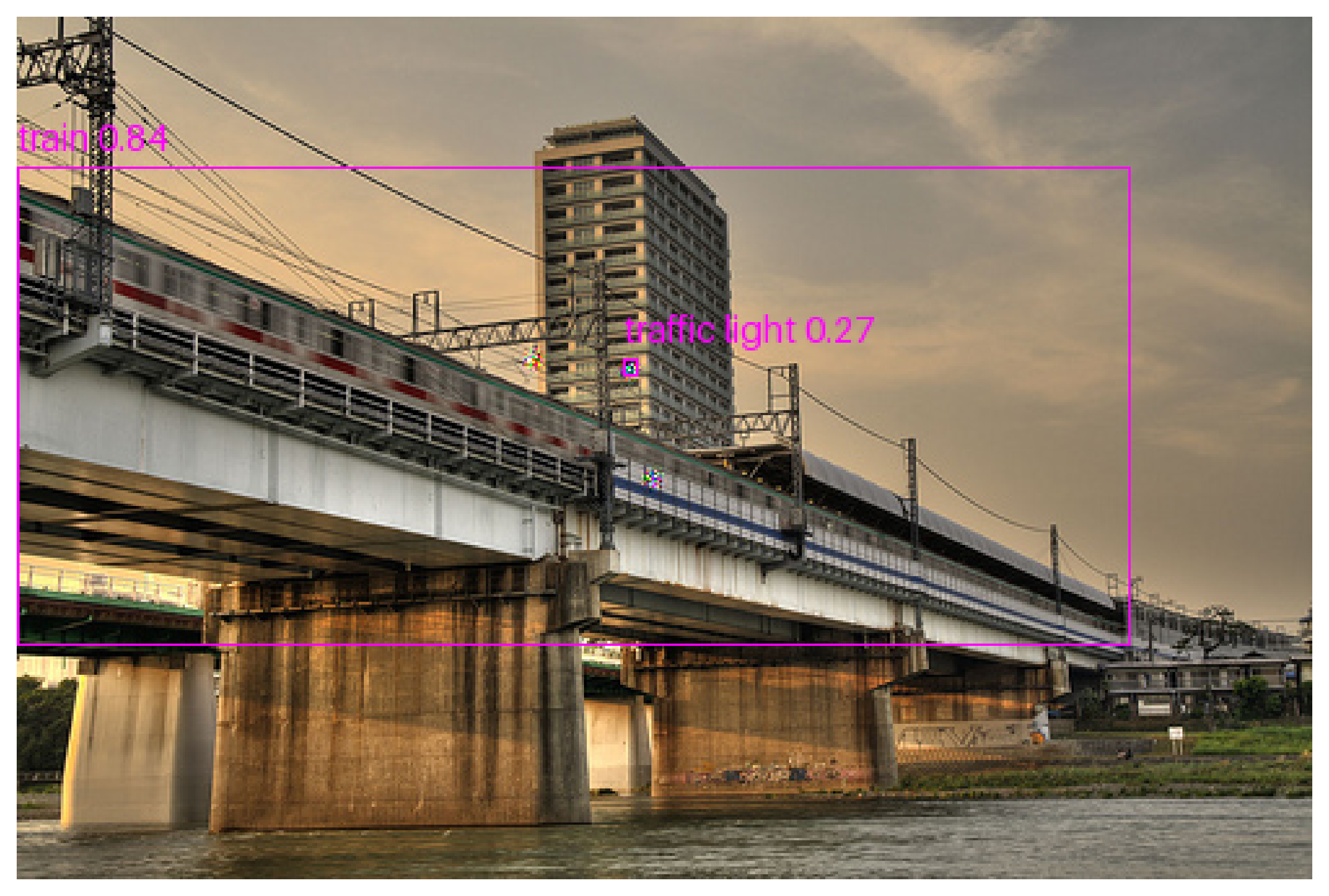}
    \caption{DRISE\_MoG \\ Outcome: added a traffic light\\ LPIPS: 0.001   $L_2$:  0.0084}
    \label{fig:combined_example_train:driseMoG}
  \end{subfigure}
  \begin{subfigure}[ht]{0.29\linewidth}\centering
    \includegraphics[width=\linewidth]{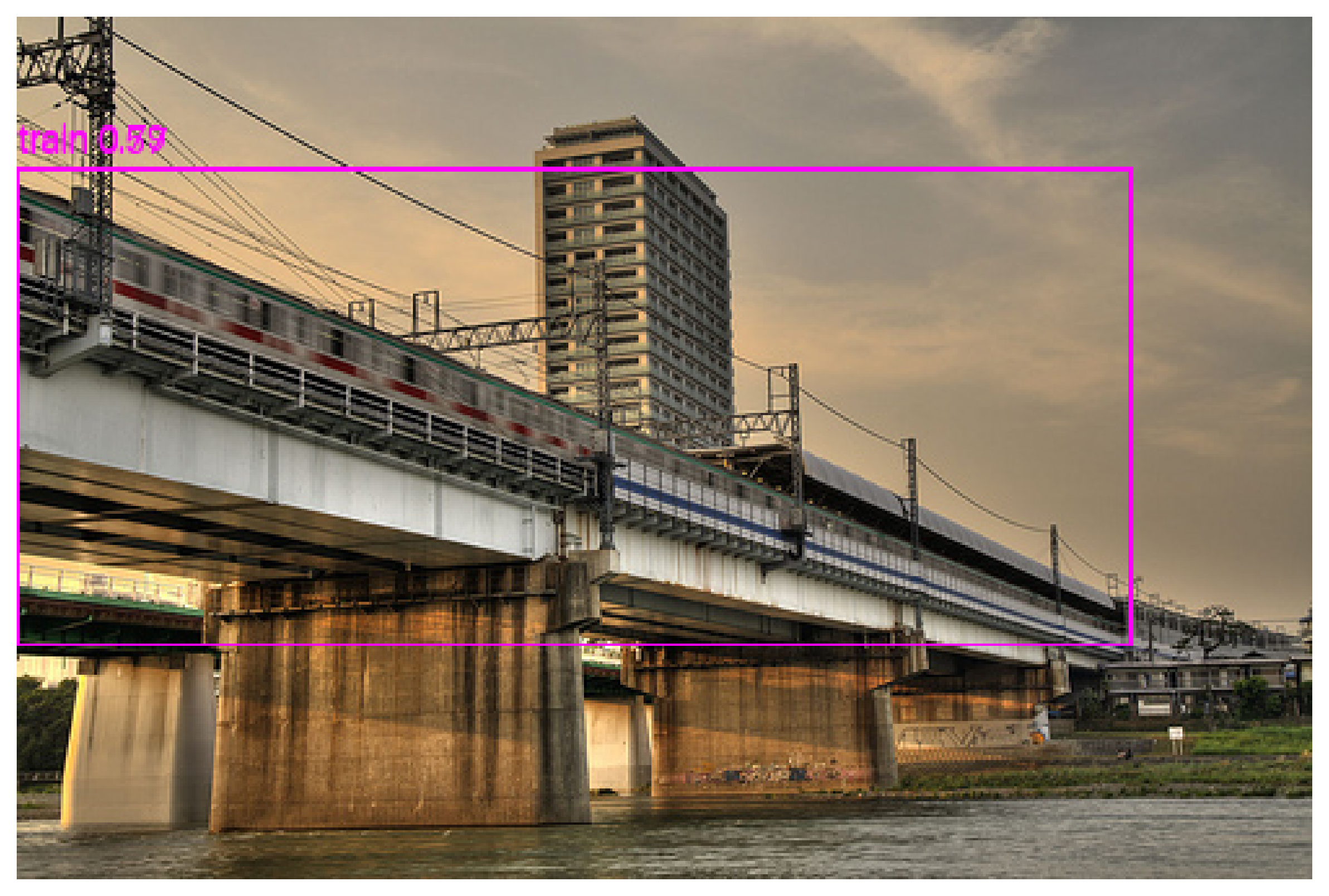}
    \caption{ReX\_MoG \\ Outcome: added another train (on top)\\ LPIPS: \textbf{\textit{7.18e-07}}   $L_2$:  \textbf{\textit{0.0004}}}
    \label{fig:combined_example_train:rexMoG}
  \end{subfigure}
    \caption{(a) Original image and detector bbox; (b–d) responsibility heatmaps (same min/max scale) used for \blackcattGreedy{} (e–g) minimal sufficient pixel sets (\msps) and saliency map for DRISE; (h–l) attack outcomes produced by different methods.}
    \label{fig:combined_example_train}
\end{figure*}

\begin{figure*}[t!]
    \centering
    \begin{subfigure}[t]{0.24\linewidth}\centering
    \includegraphics[width=\linewidth]{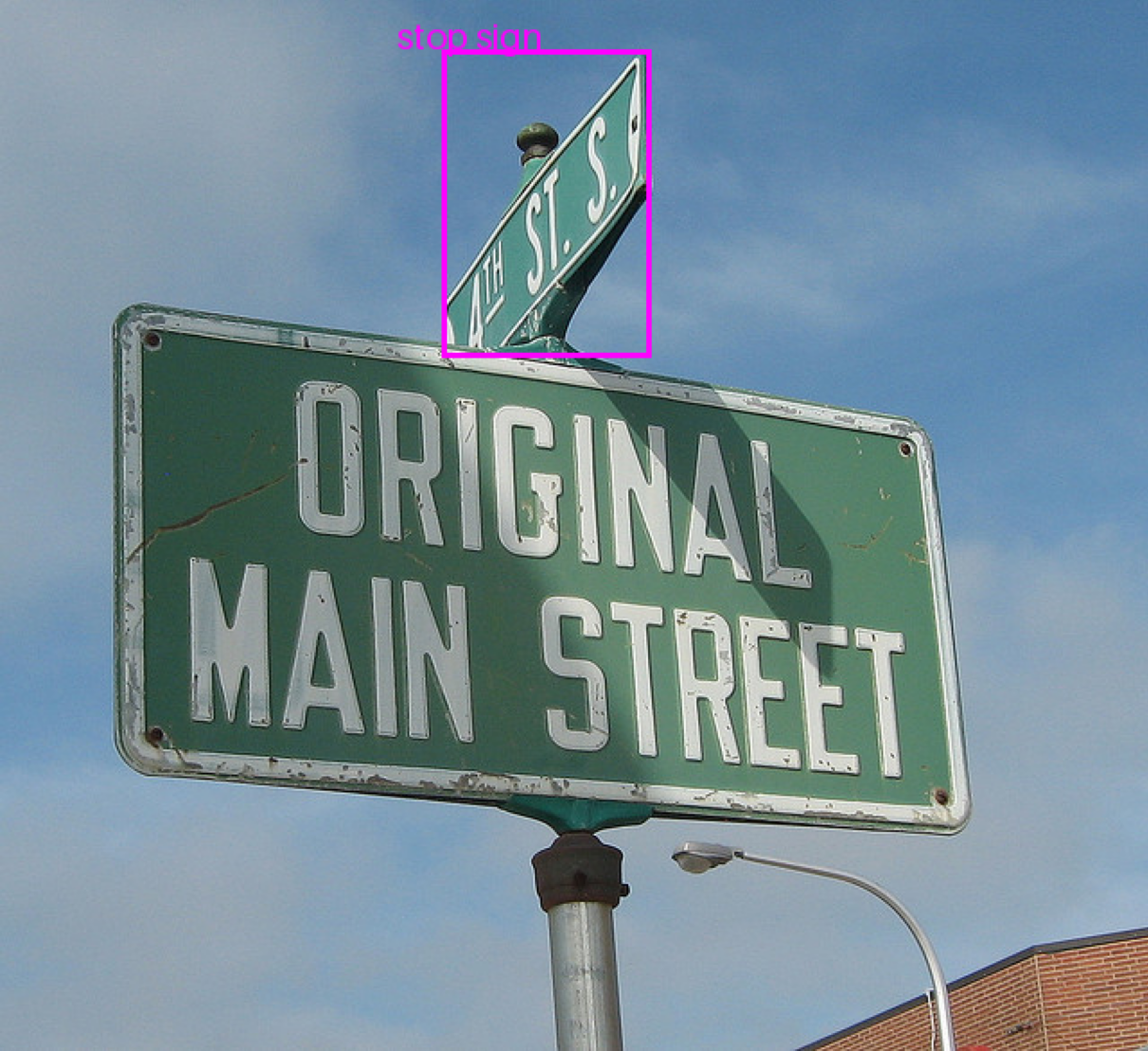}
    \caption{Original (bbox)}
     \end{subfigure}%
  \begin{subfigure}[t]{0.24\linewidth}\centering
    \includegraphics[width=\linewidth]{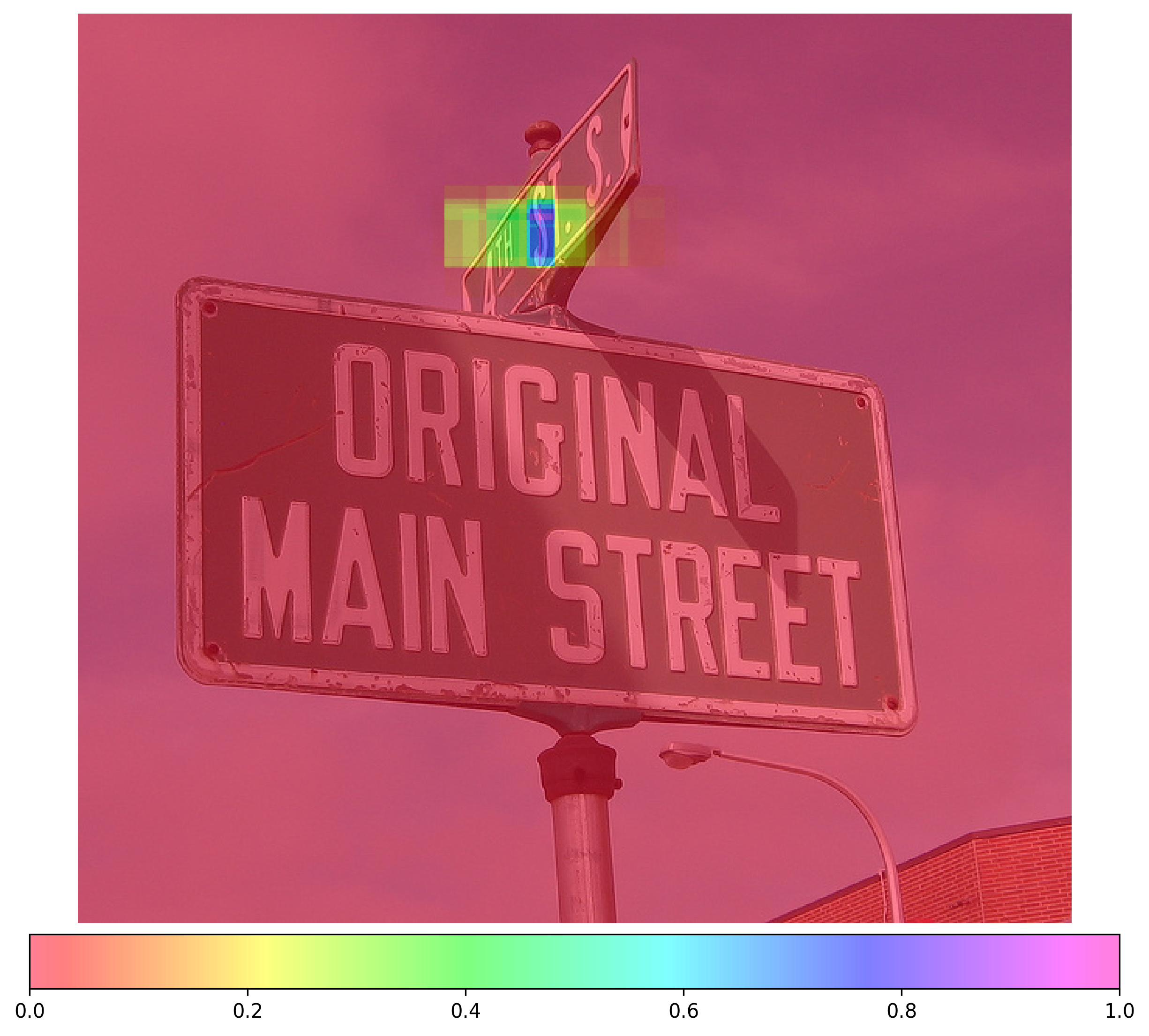}
    \caption{ReX Responsibility Map}
  \end{subfigure}%
  \begin{subfigure}[t]{0.24\linewidth}\centering
    \includegraphics[width=\linewidth]{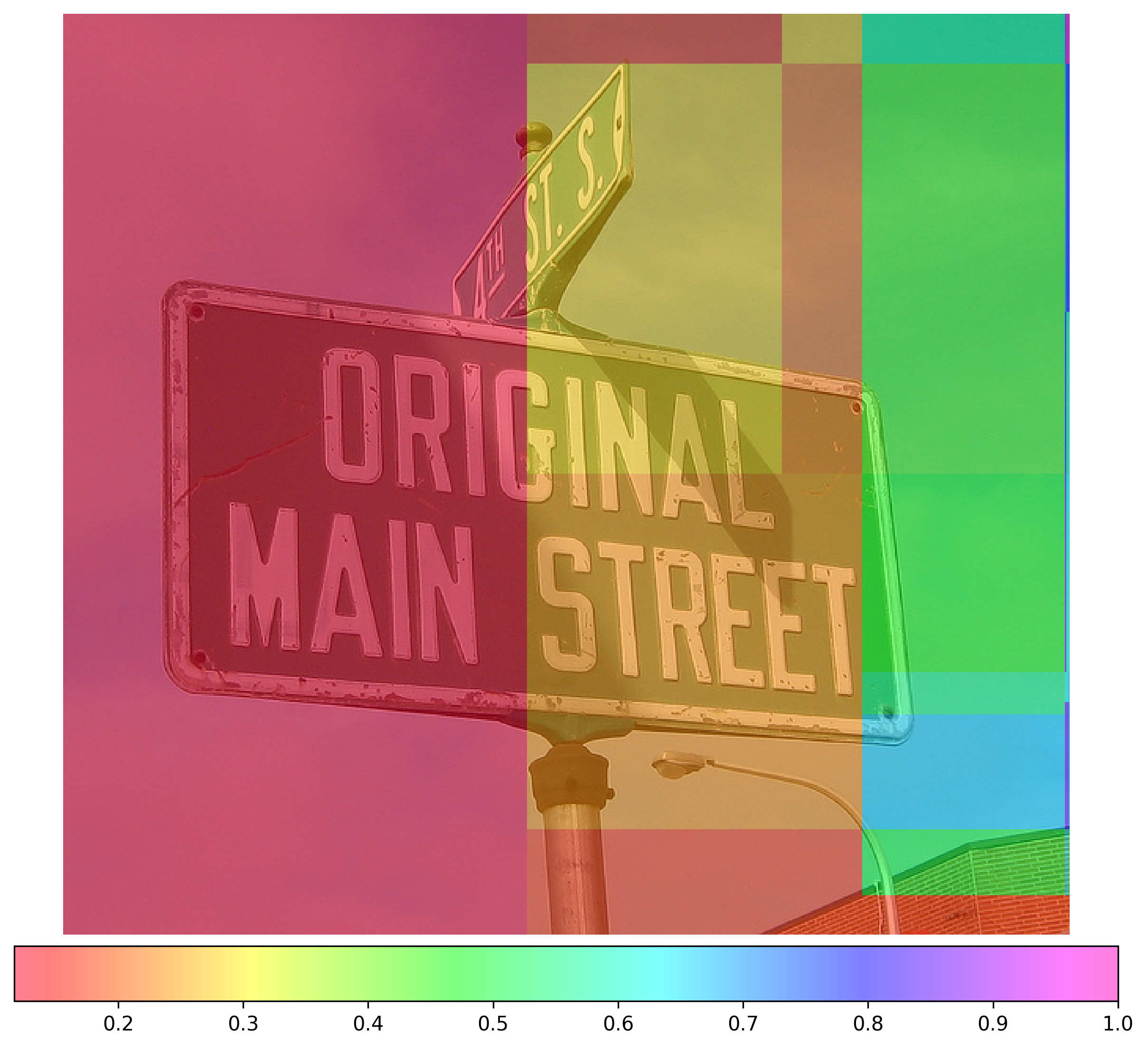}
    \caption{ReX Negative Responsibility Map}
  \end{subfigure}%
  \begin{subfigure}[t]{0.24\linewidth}\centering
    \includegraphics[width=\linewidth]{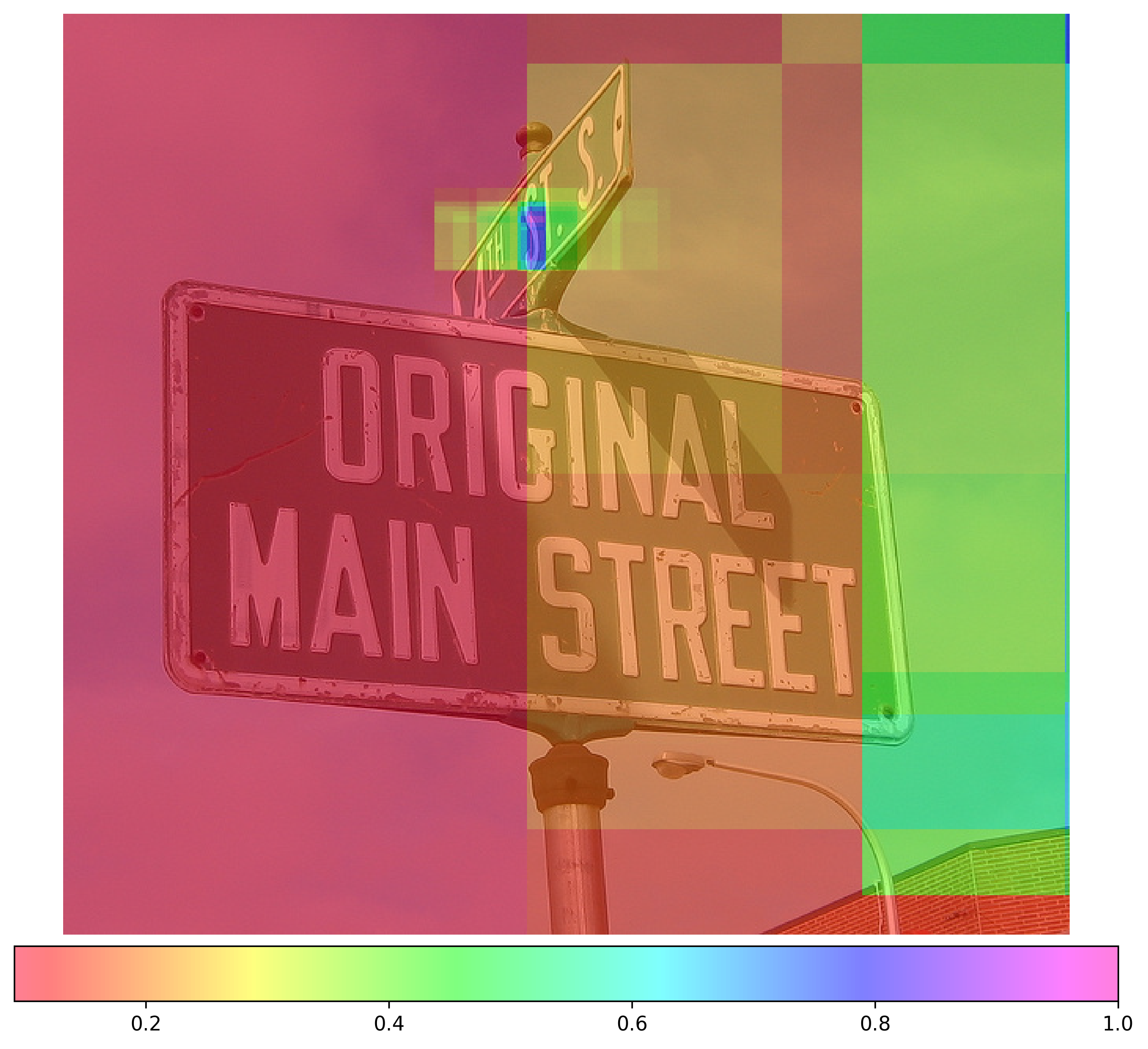}
    \caption{Combined responsibility}
  \end{subfigure}

  \begin{subfigure}[t]{0.24\linewidth}\centering
    \includegraphics[width=\linewidth]{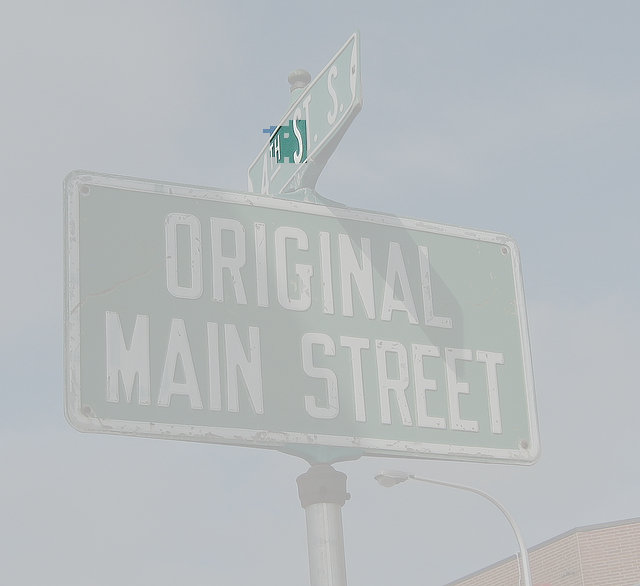}
    \caption{ReX \msps (overlay)}
  \end{subfigure}%
  \begin{subfigure}[t]{0.24\linewidth}\centering
    \includegraphics[width=\linewidth]{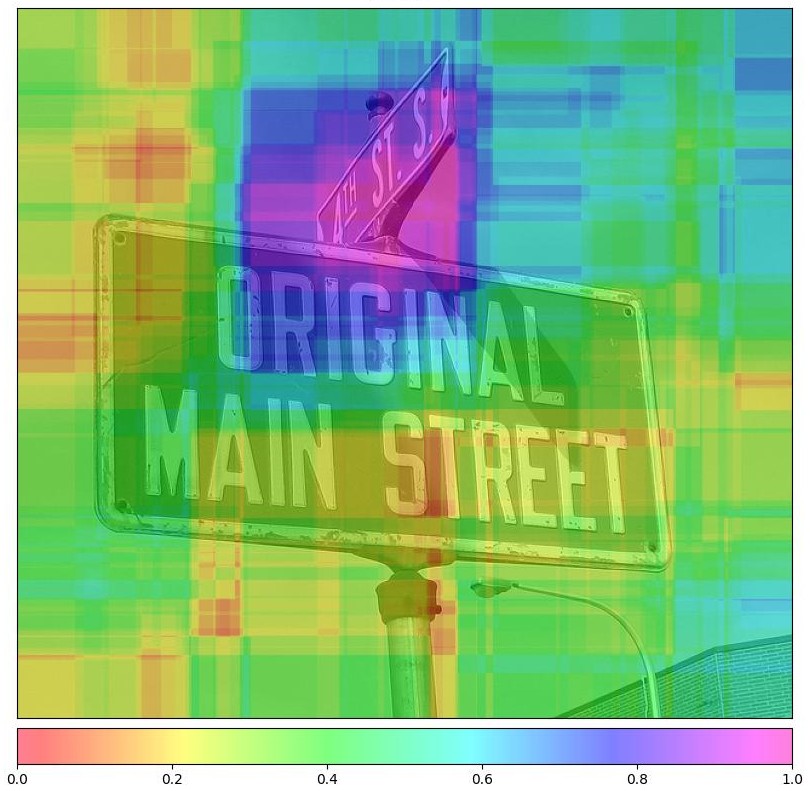}
    \caption{DRISE Saliency map}
  \end{subfigure}%
  \begin{subfigure}[t]{0.24\linewidth}\centering
    \includegraphics[width=\linewidth]{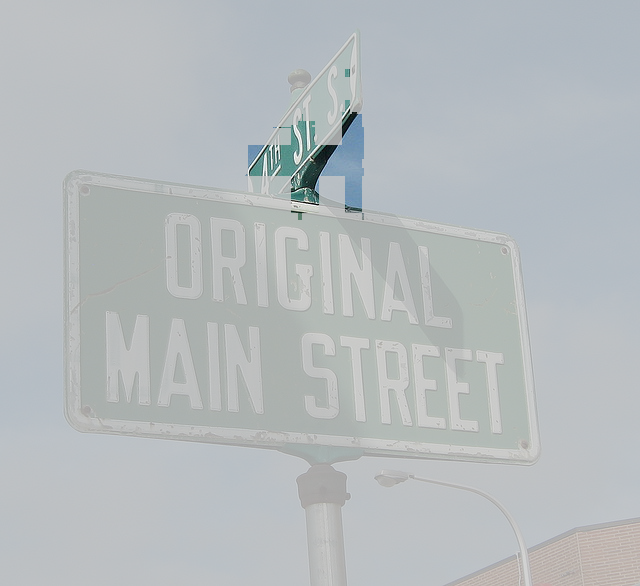}
    \caption{DRISE \msps (overlay)}
  \end{subfigure}

    \begin{subfigure}[ht]{0.29\linewidth}\centering
    \includegraphics[width=\linewidth]{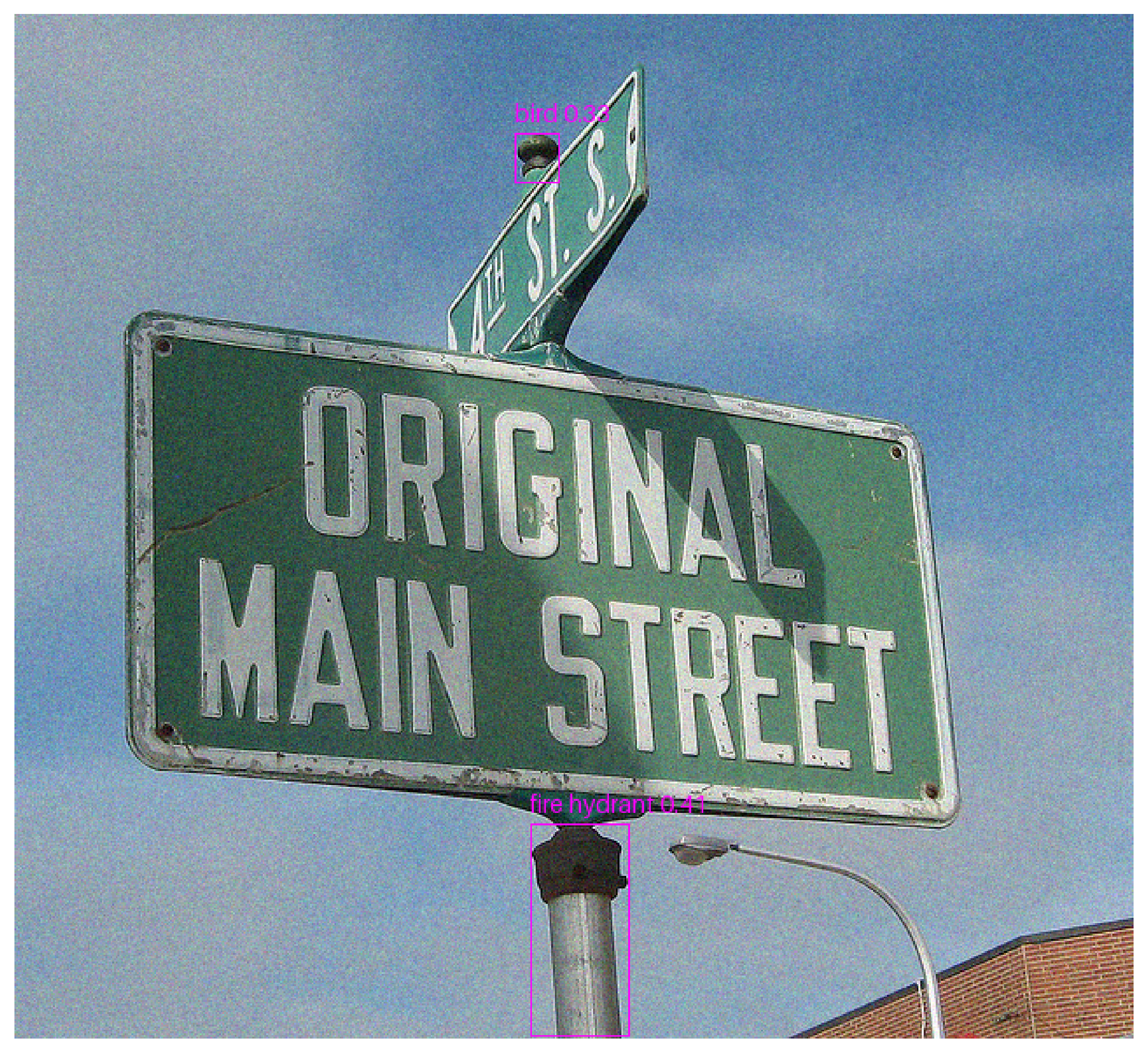}
    \caption{Noise \\ Outcome: change to bird and fire hydrant \\ LPIPS: 0.2201  $L_2:$ 0.0498}
    \end{subfigure}
  \begin{subfigure}[ht]{0.29\linewidth}\centering
    \includegraphics[width=\linewidth]{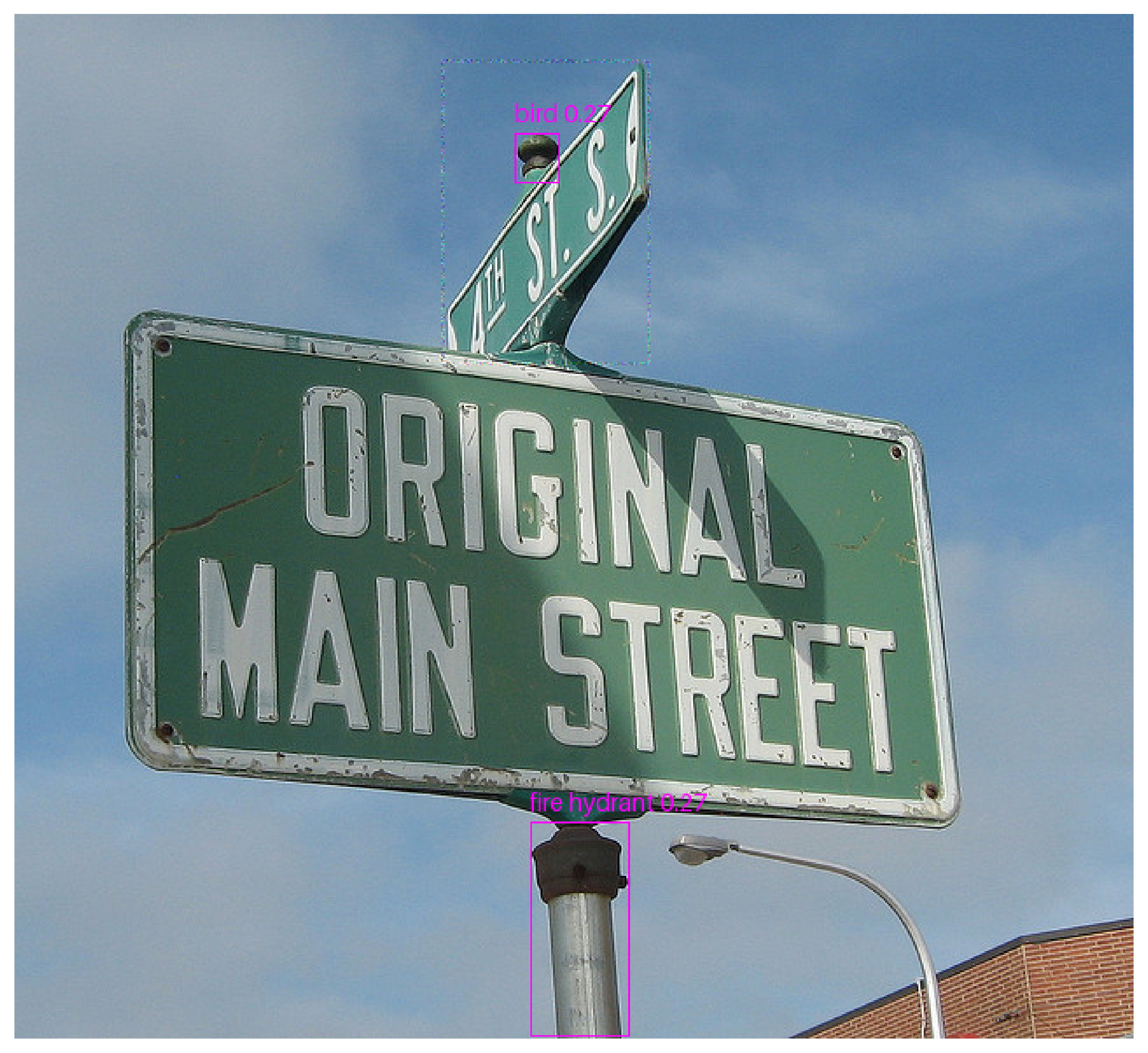}
    \caption{Targeted noise \\ Outcome: change to bird and fire hydrant\\ LPIPS: 0.0075  $L_2:$ 0.0057}
  \end{subfigure}
  \centering
  \begin{subfigure}[ht]{0.29\linewidth}\centering
    \includegraphics[width=\linewidth]{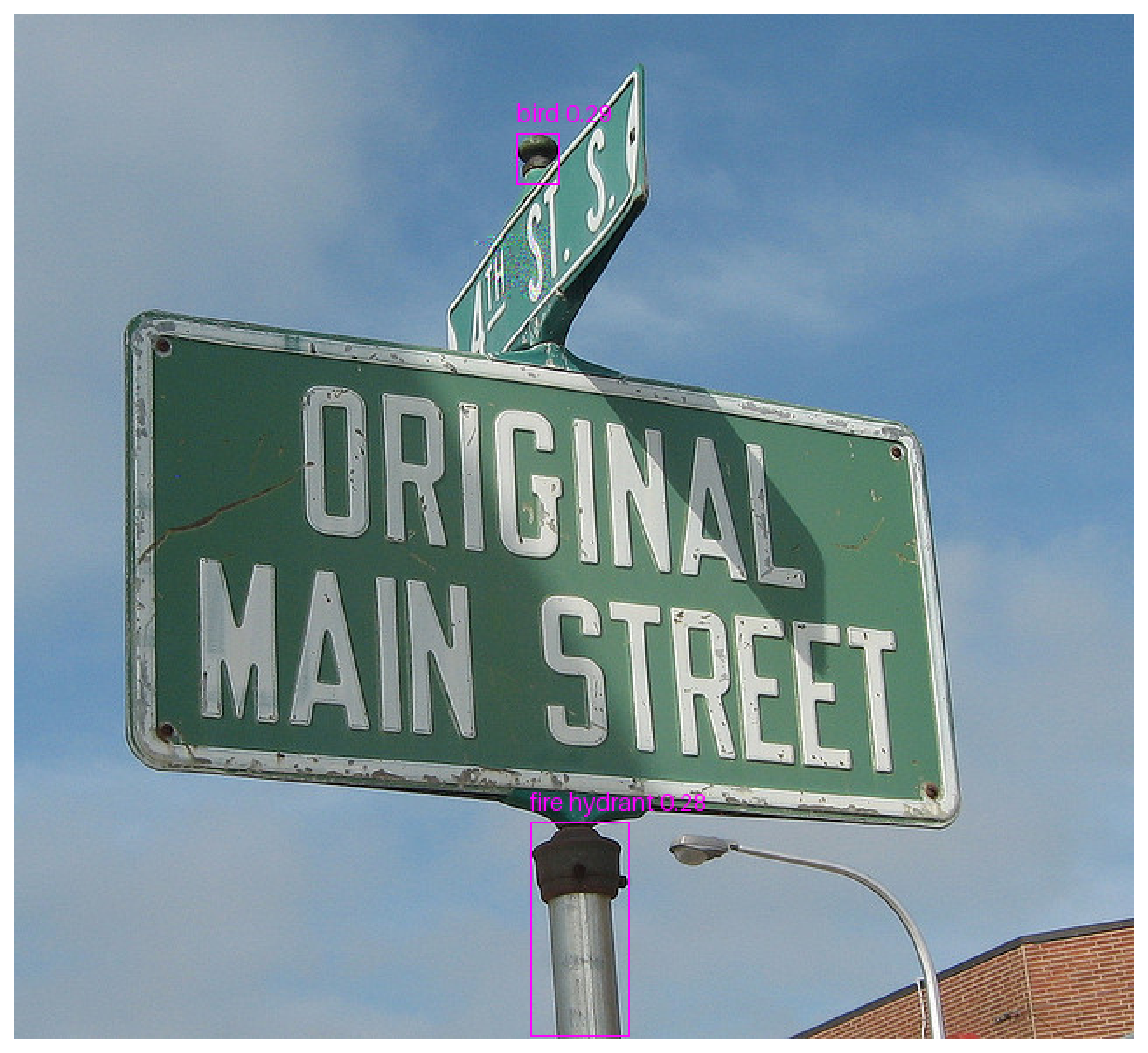}
    \caption{Blended \\ Outcome: change to bird and fire hydrant\\ LPIPS: 0.0004  $L_2:$  0.0059}
  \end{subfigure}
  \begin{subfigure}[ht]{0.29\linewidth}\centering
    \includegraphics[width=\linewidth]{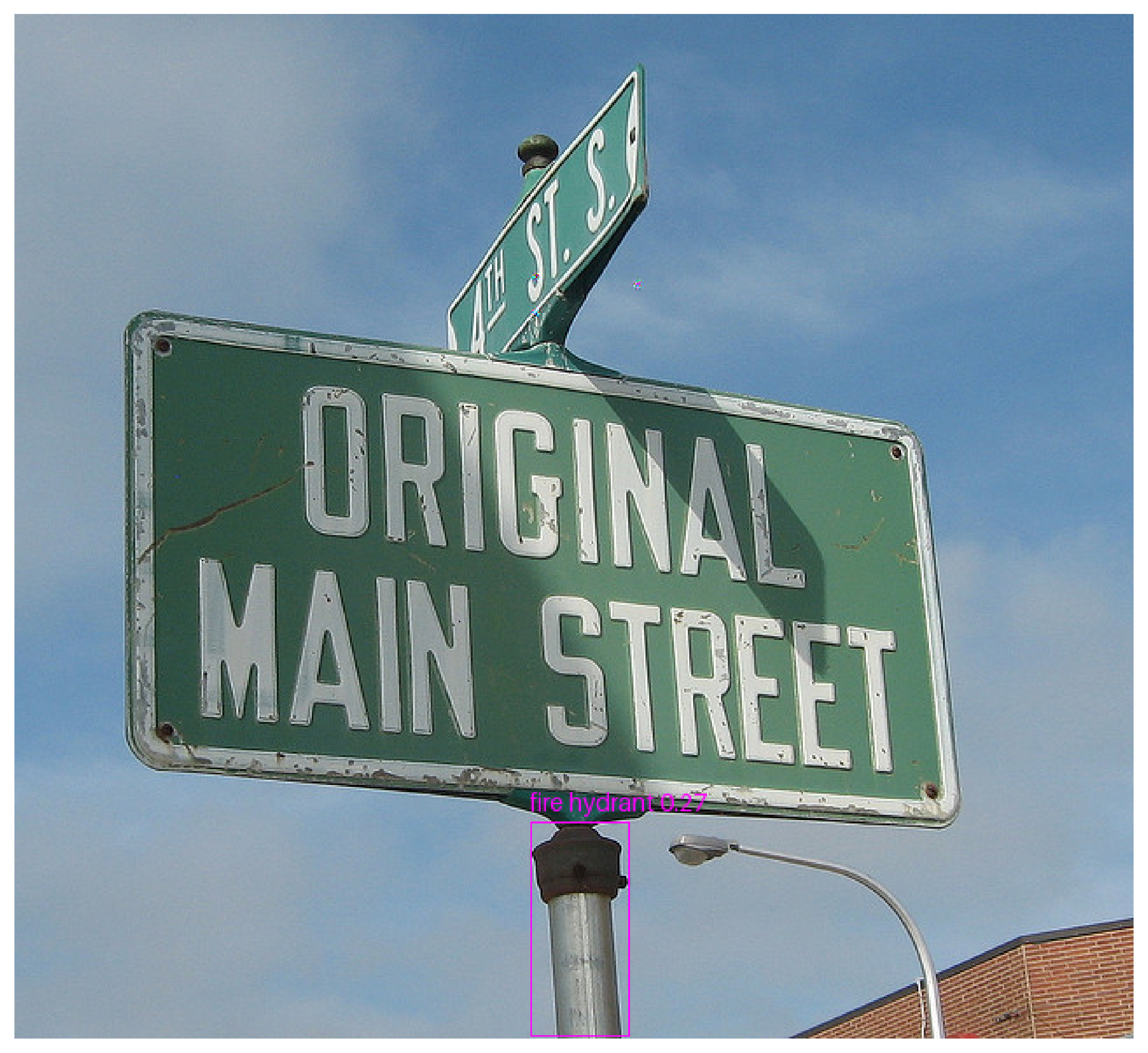}
    \caption{DRISE\_MoG \\ Outcome: change to fire hydrant\\ LPIPS: 0.0007  $L_2$: \textbf{\textit{0.0033}}} 
  \end{subfigure}
  \begin{subfigure}[ht]{0.29\linewidth}\centering
    \includegraphics[width=\linewidth]{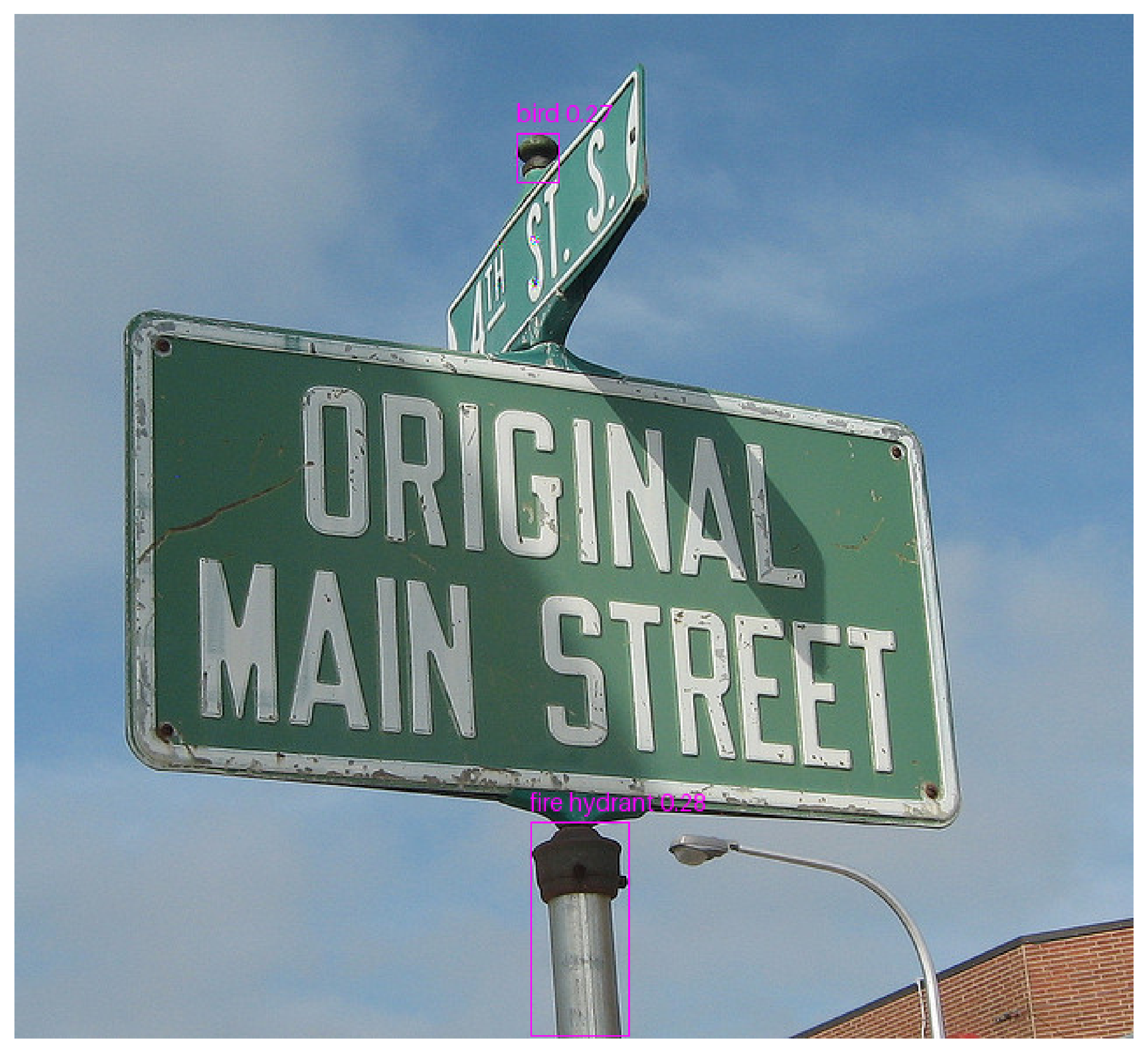}
    \caption{ReX\_MoG \\ Outcome: change to bird and fire hydrant \\ \textbf{\textit{LPIPS: 0.0001}} $L_2:$ 0.0038}
  \end{subfigure}
    \caption{(a) Original image and detector bbox; (b–d) responsibility heatmaps (same min/max scale) used for \blackcattGreedy{} (e–g) minimal sufficient pixel sets (\msps) and saliency map for DRISE; (h–l) attack outcomes produced by different methods. The original label, stop sign, doesn’t appear in any of the new detections in the new image.}
    \label{fig:combined_example_stop_sign}
\end{figure*}

\Cref{fig:combined_example_train} illustrate how the responsibility maps from \rex minimal sufficient pixel sets (\msps) isolate the causal regions that drive a detector’s decision, and how perturbing these regions in \Cref{fig:combined_example_train:blended,fig:combined_example_train:driseMoG,fig:combined_example_train:rexMoG} help yield lower LPIPS/$L_2$ distortion compared to noise baselines. As shown in \cref{fig:combined_example_stop_sign}, perturbing causally-relevant pixels enables efficient misclassification of the ‘stop sign,’ with \blackcatt{} variants requiring far smaller distortions than noise-based baselines.

\Cref{algo:refine} performs a fine-grained search around the best perturbation found so far, progressively shrinking the noise budget or concentrating it on responsibility-guided regions. This ensures that \blackcatt{} reaches the lowest-distortion adversarial outcome permitted by the causal prior.
\begin{algorithm}[t]
\caption{apply\_refinement$(\mathcal{X},\mathcal{N},\mathcal{R},\text{metrics})$ — (applies to Greedy / Blended refinement)}
\label{algo:refine}
\begin{flushleft}
  \textbf{INPUT:} original image $\mathcal{X}$, detector $\mathcal{N}$, map $\mathcal{R}$, current best values: metrics\\
    \textbf{OUTPUT: } updated best attack.
\end{flushleft}
\begin{algorithmic}[1]
        \STATE $\sigma_{\text{base}} \leftarrow$ \texttt{best.sigma} if defined, else default
      \STATE $\Sigma \leftarrow$ geometric $\cup$ linear schedule decreasing from $\sigma_{\text{base}}$ to $0.05 \times \sigma_{\text{base}}$
  \IF{\blackcattGreedy}
      \STATE $c \leftarrow \mathcal{R}.\text{flatten}()$;\;\; order $\leftarrow \text{argsort}(c)$ (descending)
      \STATE $K \leftarrow$ schedule of mask sizes (e.g.\ geometric from $1\%$ to $100\%$)
      \FOR{$k \in K$}
        \STATE mask\_flat $\leftarrow$ \textbf{False}($|c|$)
        \STATE mask\_flat[order$[:k]$] $\leftarrow$ \textbf{True}
        \STATE mask $\leftarrow$ reshape(mask\_flat)
    
        \FOR{$\sigma \in \Sigma$}
           \STATE $\mathcal{X}_p \leftarrow$ apply\_attack$(\mathcal{X}, mask, \sigma)$
           \STATE preds $\leftarrow \mathcal{N}(\mathcal{X}_p)$
           \IF{metrics\_improve(preds)}
                \STATE metrics $\leftarrow$ update\_metrics(preds)
            \ENDIF
        \ENDFOR
      \ENDFOR
  \ENDIF
  \IF{\blackcattMOG}
    \FOR{$\sigma \in \Sigma$}
           \STATE $\mathcal{X}_p \leftarrow$ apply\_attack$(\mathcal{X}, mask, \sigma)$
           \STATE preds $\leftarrow \mathcal{N}(\mathcal{X}_p)$
           \IF{metrics\_improve(preds)}
                \STATE metrics $\leftarrow$ update\_metrics(preds)
            \ENDIF
        \ENDFOR
  \ENDIF

  \STATE \textbf{return} metrics
\end{algorithmic}
\end{algorithm}

\begin{table*}[!]
    \centering
    \caption{The impact of using \blackcatt on RT-DETR. `Success' reports the attack success rate under the respective norm constraints. The \textbf{bold} values indicate the best value across method type; \underline{underlined} values indicate the best value across all method types within the constraint.}
    \label{tab:comparison_rt-detr}
    \resizebox{\textwidth}{!}{
    \begin{tabular}{ll cccccccc ||c}
    \toprule
    \textbf{Objective} & \textbf{Method} & \textbf{Constraint} &   
    \shortstack{\textbf{Avg.}\\\textbf{Queries}} &
    \shortstack{\textbf{Avg.}\\\bm{$L_0$}} & 
    \shortstack{\textbf{Std.}\\\bm{$L_0$}} & 
    \shortstack{\textbf{Avg.}\\\bm{$L_2$}} & 
    \shortstack{\textbf{Std.}\\\bm{$L_2$}} &
    \shortstack{\textbf{Avg.}\\\textbf{LPIPS}} & 
    \shortstack{\textbf{Std.}\\\textbf{LPIPS}} &
    \shortstack{\textbf{Success}\\(\bm{\%})} \\
    
    \midrule\midrule
    \multirow{6}{*}{\textbf{Remove Pred}}
    & SA & $L_2 \leq 4/255$         & 367.52 &0.988  &0.024  &0.015  &0.001  &0.021  &0.017& \underline{\textbf{8.1}} \\
    & SA$_{exp}$ & $L_2 \leq 4/255$ & \textbf{17.23} &\textbf{0.032}  &0.068  &\textbf{0.014}  &0.002  &\underline{\textbf{0.011}}  &0.011& 4.7 \\ 
    \cdashline{2-11}
    & PRFA & $L_2 \leq 4/255$ & \textbf{1.17} &0.990  &0.010  &0.016  &0.000  &0.093  &0.085& 2.3 \\
    & PRFA$_{exp}$ & $L_2 \leq 4/255$ & 333.93 &\textbf{0.097}  &0.124  &\underline{\textbf{0.005}}  &0.004  &\textbf{0.019}  &0.028  &  1.5\\
    \cdashline{2-11}
    & \blackcatt$_{\text{loss}}$ & $L_2 \leq 4/255$ &  274.45 &\underline{\textbf{0.027}}  &0.086  &0.013  &0.002  &0.013  &0.021& 5.3 \\
    \cmidrule{2-11}
    & SparseRS & $L_0 \leq 0.005$ & 431.72 &0.005  &0.000  &0.041  &0.002  &0.255  &0.152  &\underline{\textbf{18.0}} \\
    & SparseRS$_{exp}$ & $L_0 \leq 0.005$ & \textbf{285.91} &0.005  &0.000  &0.041  &0.002  &\underline{\textbf{0.051}}  &0.054  &17.1 \\ 
     \cdashline{2-11}
    & BlackCAtt$_{MoG_{\text{loss}}}$ & $L_0 \leq 0.005$ & \underline{\textbf{111.72}} &0.005  &0.000  &\underline{\textbf{0.031}}  &0.004  &0.065  &0.049  &17.4 \\ 
    \midrule
    
    \multirow{6}{*}{\shortstack{\textbf{Change}\\\textbf{Pred}}}
    & SA & $L_2 \leq 4/255$ & 601.04 &0.989  &0.015  &\textbf{0.014}  &0.001  &0.022  &0.018  &\underline{\textbf{2.5}} \\ 
    & SA$_{exp}$ & $L_2 \leq 4/255$ & \underline{\textbf{92.50}} &\textbf{0.197}  &0.260  &0.015  &0.001  &\textbf{0.018}  &0.017  &0.2 \\ 
     \cdashline{2-11}
    & PRFA & $L_2 \leq 4/255$ &  3500 &nan  &nan  &nan  &nan  &nan  &nan  &0.0 \\ 
    & PRFA$_{exp}$ & $L_2 \leq 4/255$ &  3500 &nan  &nan  &nan  &nan  &nan  &nan  &0.0 \\ 
     \cdashline{2-11}
    & BlackCAtt$_{\text{loss}}$ & $L_2 \leq 4/255$ & 645 &\underline{\textbf{0.125}}  &0.256  &\underline{\textbf{0.013}}  &0.003  &\underline{\textbf{0.011}}  &0.010  &0.5 \\ 
    \cmidrule{2-11}
    & SparseRS & $L_0 \leq 0.005$ & 1351.8 &0.005  &0.000  &0.041  &0.002  &0.317  &0.165  &\textbf{1.0} \\
    & SparseRS$_{exp}$ & $L_0 \leq 0.005$ &  \underline{\textbf{57.33}} &0.005  &0.000  &0.041  &0.003  &\textbf{0.077}  &0.078  &0.3 \\ 
     \cdashline{2-11}
    & BlackCAtt$_{MoG_{\text{loss}}}$  &  $L_0 \leq 0.005$ & 159.9 &0.005  &0.000  &\underline{\textbf{0.032}}  &0.004  &\underline{\textbf{0.070}}  &0.046  &\underline{\textbf{1.5}} \\ 
    \midrule
    
    \multirow{6}{*}{\shortstack{\textbf{Add}\\\textbf{New Pred}}}
    & SA & $L_2 \leq 4/255$ & 18.997 &0.988  &0.026  &0.015  &0.000  &0.032  &0.030  &\underline{\textbf{93.9}} \\ 
    & SA$_{exp}$ & $L_2 \leq 4/255$ & \underline{\textbf{8.517}} &\textbf{0.021}  &0.032  &\textbf{0.012}  &0.004  &\textbf{0.008}  &0.008  &88.1 \\ 
      \cdashline{2-11}
    & PRFA & $L_2 \leq 4/255$ & 35.76 &0.988  &0.027  &0.016  &0.000  &0.086  &0.074  &\textbf{88.1} \\
    & PRFA$_{exp}$ & $L_2 \leq 4/255$ & \underline{\textbf{26.06}} &\textbf{0.021}  &0.032  &\underline{\textbf{0.002}}  &0.001  &\underline{\textbf{0.001}}  &0.005  &86.7 \\ 
    \cdashline{2-11}
    & BlackCAtt$_{\text{loss}}$ & $L_2 \leq 4/255$ &  79.57 &\underline{\textbf{0.005}}  &0.011  &0.005  &0.005  &0.003  &0.007  &77.0 \\ 
    \cmidrule{2-11}
    & SparseRS & $L_0 \leq 0.005$ & \textbf{19.49}7 &0.005  &0.000  &0.040  &0.002  &0.224  &0.138  &\underline{\textbf{95.8}} \\
    & SparseRS$_{exp}$ & $L_0 \leq 0.005$ & 36.02 &0.005  &0.000  &0.040  &0.002  &0.045  &0.032  &91.2 \\
      \cdashline{2-11}
    & BlackCAtt$_{MoG_{\text{loss}}}$  &  $L_0 \leq 0.005$ & \underline{\textbf{5.96}} &0.005  &0.000  &\underline{\textbf{0.028}}  &0.003  &\underline{\textbf{0.032}}  &0.017  &90.2 \\ 
    \bottomrule
    \end{tabular}
    }
\end{table*}

\Cref{tab:comparison_rt-detr} shows how different attack methods and their \blackcatt{} variant performs against RT-DETR model across the different attack goals. Forcing the model to create new predictions is the easiest vulnerability to exploit, yielding success rates over 80\% across most methods. Conversely, changing an existing prediction is incredibly difficult, with almost all methods failing or achieving less than a 3\% success rate. This could be due to a loss function that doesn't capture the gradient for changing a prediction.

\blackcatt{} stands out in this experiment as it exhibits a balance of high efficiency and low perceptible changes. The reported standard deviations (\textit{Std.}) for $L_0$ and $L_2$ norms demonstrate the consistency of our attack. While PRFA occasionally fails to converge (denoted by \textit{nan} in the ``Change Pred'' task), \blackcatt shows stable performance across all three objective types.

\begin{table*}[t!]
    \centering
    \caption{The impact of using \blackcatt on \yolo for images with multi predictions only. `Success' reports the attack success rate under the respective norm constraints. The \textbf{bold} values indicate the best value across method type; \underline{underlined} values indicate the best value across all method types within the constraint.}
    \label{tab:comparison_multi}
    \resizebox{\textwidth}{!}{
    \begin{tabular}{ll cccccccc ||c}
    \toprule
    \textbf{Objective} & \textbf{Method} & \textbf{Constraint} &   
    \shortstack{\textbf{Avg.}\\\textbf{Queries}} &
    \shortstack{\textbf{Avg.}\\\bm{$L_0$}} & 
    \shortstack{\textbf{Std.}\\\bm{$L_0$}} & 
    \shortstack{\textbf{Avg.}\\\bm{$L_2$}} & 
    \shortstack{\textbf{Std.}\\\bm{$L_2$}} &
    \shortstack{\textbf{Avg.}\\\textbf{LPIPS}} & 
    \shortstack{\textbf{Std.}\\\textbf{LPIPS}} &
    \shortstack{\textbf{Success}\\(\bm{\%})} \\
    
    \midrule\midrule
    \multirow{6}{*}{\textbf{Remove Pred}}
    & SA & $L_2 \leq 4/255$ & 402.5 &0.988  &0.015  &0.015  &0.001  &0.023  &0.020  & \underline{\textbf{4.400}} \\
    & SA$_{exp}$ & $L_2 \leq 4/255$ & \textbf{109.5} &\textbf{0.163}  &0.132  &0.015  &0.001  &\textbf{0.020}  &0.013  &4.200 \\ 
    \cdashline{2-11}
    & PRFA & $L_2 \leq 4/255$ & \underline{\textbf{1.0}} &0.986  &0.028  &0.015  &0.000  &0.112  &0.070  &2.500 \\ 
    & PRFA$_{exp}$ & $L_2 \leq 4/255$ & 261.1 & \textbf{0.224}  &0.150  &\textbf{0.009 } &0.004  &\textbf{0.025}  &0.022  &1.900 \\ 
    \cdashline{2-11}
    & \blackcatt$_{\text{loss}}$ & $L_2 \leq 4/255$ &  164.4 &\textbf{0.060}  &0.061  &0.015  &0.001  &\textbf{0.019}  &0.014  &\underline{\textbf{4.400}} \\ 
    \cmidrule{2-11}
    & SparseRS & $L_0 \leq 0.005$ & \textbf{382.9} &0.005  &0.000  &0.041  &0.002  &0.244  &0.120  &13.000 \\ 
    & SparseRS$_{exp}$ & $L_0 \leq 0.005$ & 422.8 &0.005  &0.000  &0.041  &0.002  &\textbf{0.118}  &0.064  &\textbf{13.100} \\ 
     \cdashline{2-11}
    & BlackCAtt$_{MoG_{\text{loss}}}$ & $L_0 \leq 0.005$ &\textbf{189.4} &0.005  &0.000  &\textbf{0.033}  &0.004  &\textbf{0.111}  &0.062  &\textbf{16.000} \\
    \midrule
    \multirow{6}{*}{\shortstack{\textbf{Change}\\\textbf{Pred}}}
    & SA & $L_2 \leq 4/255$ & 328.5 &0.993  &0.010  &0.015  &0.001  &0.031  &0.016  &\textbf{0.600} \\ 
    & SA$_{exp}$ & $L_2 \leq 4/255$ & \textbf{154.3} &\textbf{0.176}  &0.091  &0.015  &0.000  &\textbf{0.024}  &0.007  &0.300 \\ 
     \cdashline{2-11}
    & PRFA & $L_2 \leq 4/255$ & 3500 &nan  &nan  &nan  &nan  &nan  &nan  &0.000 \\ 
    & PRFA$_{exp}$ & $L_2 \leq 4/255$ & \textbf{1543} &\textbf{0.101}  &0.0  &\textbf{0.009} &0.0  &\textbf{0.008}  &0.0  &\textbf{0.100} \\ 
     \cdashline{2-11}
    & BlackCAtt$_{\text{loss}}$ & $L_2 \leq 4/255$ & 341.5 &0.126  &0.096  &0.014  &0.002  &0.008  &0.003  &0.200 \\ 
    \cmidrule{2-11}
    & SparseRS & $L_0 \leq 0.005$ & \textbf{706} &0.005  &0.000  &\textbf{0.040}  &0.001  &0.191  &0.093  &\underline{\textbf{2.0}} \\ 
    & SparseRS$_{exp}$ & $L_0 \leq 0.005$ & 850 &0.005  &0.000  &0.041  &\textbf{0.002}  &0.103  &\textbf{0.055}  &1.6 \\ 
     \cdashline{2-11}
    & BlackCAtt$_{MoG_{\text{loss}}}$  &  $L_0 \leq 0.005$ & \textbf{160.6} &0.005  &0.000  &\underline{\textbf{0.031}}  &0.004  &0.067  &0.039  &1.1 \\ 
    \midrule
    
    \multirow{6}{*}{\shortstack{\textbf{Add}\\\textbf{New Pred}}}
    & SA & $L_2 \leq 4/255$ & 88.2 &0.986  &0.030  &0.015  &0.001  &0.028  &0.021  &\textbf{17.3} \\  
    & SA$_{exp}$ & $L_2 \leq 4/255$ & \textbf{24.7} &\textbf{0.154}  &0.128  &\textbf{0.014}  &0.001  &\textbf{0.015 } &0.012  &16.7 \\ 
      \cdashline{2-11}
    & PRFA & $L_2 \leq 4/255$& \textbf{2.1} &0.979  &0.049  &0.015  &0.001  &0.083  &0.057  &5.400 \\ 
    & PRFA$_{exp}$ & $L_2 \leq 4/255$ & 52.9 &\textbf{0.146}  &0.123  &\textbf{0.006}  &0.003  &\textbf{0.009}  &0.013  &\textbf{11.500} \\  
    \cdashline{2-11}
    & BlackCAtt$_{\text{loss}}$ & $L_2 \leq 4/255$ & 43.8 &\underline{\textbf{0.026}}  &0.050  &\underline{\textbf{0.005}}  &0.007  &\underline{\textbf{0.007}}  &0.014  &15.300 \\ 
    \cmidrule{2-11}
    & SparseRS & $L_0 \leq 0.005$ & \textbf{75.7} &0.005  &0.000  &0.041  &0.002  &0.222  &0.119  &\underline{\textbf{17.600}} \\ 
    & SparseRS$_{exp}$ & $L_0 \leq 0.005$& 124.3 &0.005  &0.000  &0.041  &0.002  &\textbf{0.108}  &0.060  &16.400 \\
      \cdashline{2-11}
    & BlackCAtt$_{MoG_{\text{loss}}}$  &  $L_0 \leq 0.005$ & \underline{\textbf{34.5}} &0.005  &0.000  & \underline{\textbf{0.029}}  &0.003  &\underline{\textbf{0.048}}  &0.028  &15.700 \\
    \bottomrule
    \end{tabular}
    }
\end{table*}

When dealing with more complex scenes containing multiple object predictions, the attack tasks become highly challenging across the board as shown in \Cref{tab:comparison_multi}. Under the $L_2$ constraint, success rates for ``Remove Pred'' task drops to 5\%. \blackcatt{} matches the highest success rate, while retaining lower $L_0$ and $LPIPS$ and requiring less than half the queries of the baseline Square Attack (SA). \blackcatt{} performs much better under the stricter L0 constraint: it achieves the highest overall success rate at 16\%, beating SparseRS, while slashing the required number of queries by more than half.

\end{document}